\documentclass{article}
\usepackage[utf8]{inputenc}
\usepackage{graphicx} % Required for inserting images
\usepackage{tikz} % Required for the figure of $G^\dagger_L$
\usetikzlibrary{decorations.pathmorphing,positioning}
\usepackage{amsmath,amsfonts,amssymb}
\usepackage{url}
\usepackage{amsthm}
\usepackage{mathrsfs}\usepackage{bbm}
\usepackage{dsfont}
\usepackage{color}
\usepackage{comment}
\usepackage[top=1in,bottom=1in,left=1in,right=1in]{geometry}
\usepackage{authblk}

\newtheorem{Condition}{Condition}
\newcommand{\kg}{k^*}
\newcommand{\cc}{\mathrm{cc}}

\newcommand{\bbE}{\mathbb{E}}
\newcommand{\bbP}{\mathbb{P}}

\usepackage[numbers,sort&compress]{natbib}
\usepackage{hyperref}
\usepackage{caption}

\newtheorem{theorem}{Theorem}

\newtheorem{proposition}{Proposition}
\newtheorem{lemma}{Lemma}

\newtheorem{remark}{Remark}

\def\cA{\mathcal{A}}
\def\cB{\mathcal{B}}

\def\cG{\mathcal{G}}

\def\cJ{\mathcal{J}}

\def\cN{\mathcal{N}}

\def\cW{\mathcal{W}}

\title{Low-degree polynomials lower bounds for clustering in small dimension}
\author[1]{Alexandra Carpentier}
\author[2]{Nicolas Verzelen}

\affil[1]{\textit{\scriptsize{Institut für Mathematik, Universität Potsdam, Germany.}}}
\affil[2]{\textit{\scriptsize{INRAE, MISTEA, Univ. Montpellier, Montpellier, France}} }

\begin{document}
\maketitle

\begin{abstract}
We study the computational limits of clustering $n$ observations drawn from a mixture of $K$ isotropic distributions in $\mathbb R^d$, in the regime where the number $K$ of groups is much larger than the dimension $d$. In this regime, tensor-based procedures recover the partition with a polylogarithmic separation between the centers, but they require a sample size that is a large and non-explicit polynomial of $dK$, whereas the existing low-degree lower bounds only show that $n$ must be larger than $dK^2$. We introduce a mixture of $2K$ product Bernoulli distributions on $\{0,1\}^d$, with $K=(d+1)^m$, whose centers are built from a dual BCH code. This construction ensures that the moment tensors of the observations coincide with those of the uniform distribution on $\{-1,1\}^d$ up to the order $2m+1$. For any fixed $m$, we establish a low-degree polynomial lower bound showing that clustering is hard as long as the separation $\Delta$ satisfies $\Delta^2\ll(K^2d^{3/2}/n)^{1/(2m+2)}\wedge\sqrt d$, up to polylogarithmic factors. In particular, with a polylogarithmic separation, low-degree polynomials require $n\gtrsim K^2d^{3/2}$ observations, up to polylogarithmic factors. Through a Gaussianization argument, this hardness carries over, in the sense of reductions, to isotropic Gaussian mixtures. Our proof relies on a family of polynomials of the observations, indexed by bipartite graphs, which is almost orthonormal. The moments of these polynomials are governed by the number of solutions of polynomial systems over the finite field $\mathbb F_{d+1}$ that are induced by the BCH code, and the crux of our analysis is to connect the combinatorial structure of the bipartite graphs with the structure of the solution sets of these systems. 

\end{abstract}

\section{Introduction}\label{sec:intro}

Clustering $n$ observations into $K$ groups is one of the most basic tasks of unsupervised learning. Its most emblematic statistical formulation is the isotropic Gaussian mixture model, which has attracted a lot of attention both in statistics and in theoretical computer science~\cite{Dasgupta99,VEMPALA2004,LuZhou2016,Regev2017,fei2018hidden,giraud2019partial,Kwon20,loffler2021optimality,LiuLi2022,Even24}. For binary data, the natural counterpart is a mixture of product Bernoulli distributions, also known as a latent class model~\cite{feldman2008learning,chen2019beyond}. In this paper, we are interested in the computational limits of clustering in the regime where the number $K$ of groups is much larger than the dimension $d$.

\subsection{Clustering Gaussian and Bernoulli mixtures}

\paragraph{Gaussian mixtures.} For some unknown centers $\theta_1,\ldots,\theta_K\in\mathbb R^d$ and an unknown assignment $\kg:[n]\to[K]$, we observe independent random vectors $Y_i\sim\mathcal N(\theta_{\kg(i)},I_d)$, $i=1,\ldots,n$, the groups $(\kg)^{-1}(\{k\})$ having comparable sizes of order $n/K$. The difficulty of the problem is driven by the separation $\Delta_*:=\min_{k\neq k'}\|\theta_k-\theta_{k'}\|_2$, and the goal is to recover the partition of $[n]$ induced by $\kg$ better than a random guess. Setting aside computational constraints, this is possible as soon as $\Delta_*^2$ is large compared to
$$\Delta_{\mathrm{stat}}^2:=\sqrt{\frac{dK\log(K)}{n}}\vee\log(K)\enspace ,$$
see~\cite{Regev2017,Kwon20,Even24}. This threshold is achieved by the exact $K$-means estimator~\cite{Even24}, which is however NP-hard to compute in the worst case~\cite{aloise2009np}, and even hard to approximate~\cite{awasthi2015hardness}. The central question is therefore whether a polynomial-time procedure can cluster at a separation close to $\Delta_{\mathrm{stat}}$, or whether a statistical-computational gap arises.

\paragraph{Bernoulli mixtures.} In the Bernoulli counterpart, we observe $Y^*_i\in\{0,1\}^d$ whose coordinates are, conditionally on $\kg$, independent with $\bbP[Y^*_{ij}=1]=(1+\theta_{\kg(i),j})/2$, where now $\theta_k\in[-1,1]^d$. The centered observations $Y_i:=2Y^*_i-1\in\{-1,1\}^d$ have mean $\theta_{\kg(i)}$ and independent sub-Gaussian coordinates with variance at most $1$, so that the Gaussian and Bernoulli models share the same signal-to-noise ratio, measured again by $\Delta_*$. In particular, the guarantees of distance-based, spectral and semidefinite procedures, which only rely on sub-Gaussian concentration, are the same in both models~\cite{LuZhou2016,fei2018hidden,giraud2019partial}. Conversely, as explained later in this section, the Bernoulli model can be mapped to a Gaussian mixture with a separation smaller by a factor of order $\sqrt{\log(n)}$, so that hardness results for the former carry over, in the sense of reductions, to the latter.

\subsection{Polynomial-time algorithms and low-degree lower bounds}

\paragraph{Low-degree polynomials.} Since clustering is an average-case problem, worst-case hardness results are not informative and computational barriers are rather assessed within restricted classes of algorithms. In the low-degree polynomial framework~\cite{Hopkins17,KuniskyWeinBandeira,SurveyWein2025}, one considers estimators that are multivariate polynomials of degree at most $D$ in the observations. For a large class of problems, polynomials of degree $D\asymp\log(n)$ are as powerful as the best known polynomial-time algorithms, so that the failure of such polynomials is considered strong evidence of computational hardness~\cite{KuniskyWeinBandeira,SurveyWein2025}. This heuristic has been formalized by Hopkins~\cite{hopkins2018statistical} as the low-degree conjecture, which predicts, for hypothesis testing problems with a product null distribution and a permutation-invariant planted distribution, that the failure of polynomials of degree slightly larger than $\log(n)$ implies the failure of all polynomial-time algorithms, once a small amount of independent noise is added. Following counterexamples to some of its variants~\cite{holmgren2021counterexamples,buhai2025quasi}, this conjecture has recently been disproved in its standard polynomial-time formulation~\cite{mao2026polynomial}. Nevertheless, the failure of low-degree polynomials, which match the best known polynomial-time algorithms for a wide range of problems, remains one of the strongest available arguments in favor of a computational barrier. Following~\cite{SchrammWein22}, low-degree lower bounds for estimation problems are formulated through the low-degree correlation between a functional $x$ of the hidden partition, for instance $x=\mathbf 1\{\kg(1)=\kg(2)\}$, and the polynomials of degree at most $D$ in the observations $Y=(Y_1,\ldots,Y_n)$, namely
\begin{equation}\label{eq:def:corr}
\mathrm{Corr}_{\leq D}:=\sup_{P\in\mathbb R_D[Y]:\ \bbE[P^2]\leq 1}\bbE[xP]\enspace ,
\end{equation}
where $\mathbb R_D[Y]$ stands for the space of polynomials with real coefficients and degree at most $D$ in the entries of $Y$. Low-degree polynomials fail to estimate $x$ when $\mathrm{Corr}_{\leq D}$ is not larger than the correlation $\bbE[x]=1/K$ achieved by constant polynomials.

\paragraph{High-dimensional regime.} When the dimension is large ($n\leq dK$), the difficulty of clustering is driven by dimension reduction. Lesieur et al.~\cite{lesieur2016phase} conjectured that, for a fixed number of groups, clustering is hard below the BBP threshold~\cite{BBP05} $\Delta_*^2\asymp\sqrt{dK^2/n}$. Even, Giraud and Verzelen~\cite{Even24,Even25} established that low-degree polynomials fail as long as $\Delta_*^2\ll_{\log}\Delta^2_{\mathrm{comp,HD}}$, where the subscript in $\ll_{\log}$ indicates that polylogarithmic factors are hidden (see Section~\ref{sec:notation}), and
$$\Delta^2_{\mathrm{comp,HD}}:=1\vee\Big(\sqrt d\wedge\sqrt{\frac{dK^2}{n}}\Big)\enspace .$$
This bound is matched, up to polylogarithmic factors and when $n\leq dK$, by a combination of hierarchical clustering --~which succeeds as soon as $\Delta_*^2\gtrsim\sqrt d\log(n)$, so that observations of the same group are closer than observations of distinct groups~-- and of a spectral projection onto a $(K\wedge d)$-dimensional subspace followed by hierarchical clustering~\cite{VEMPALA2004,Even25}. Overall, among distance-based and spectral procedures, including SDP relaxations and Lloyd's algorithm~\cite{LuZhou2016,fei2018hidden,giraud2019partial,loffler2021optimality,ndaoud2022sharp}, the best available guarantees require $\Delta_*^2\gtrsim_{\log}(\sqrt{dK^2/n}\vee\sqrt K)\wedge\sqrt d$~\cite{Even25}.

\paragraph{Moderate-dimensional regime and many groups.} When $n\geq dK$, the picture is much less clear. In this regime, the lower bound $\Delta^2_{\mathrm{comp,HD}}$ of~\cite{Even25} is still valid but it is no longer matched by spectral methods, whose guarantees reduce to $\Delta_*^2\gtrsim_{\log}\sqrt K\wedge\sqrt d$. A different stream of literature relies on high-order moment tensors, through the method of moments and tensor decompositions~\cite{hsu2013learning} or the sum-of-squares hierarchy~\cite{DBLP:journals/corr/abs-1711-07454,kothari2018robust,diakonikolas2018list}. The latter procedures cluster with a separation $\Delta_*\geq K^{\gamma}$, for any fixed $\gamma>0$, with a sample size and a running time that are polynomial in $d$ and $K$, but with an exponent that grows with $1/\gamma$. Liu and Li~\cite{LiuLi2022} proved that a polynomial-time procedure based on iterative tensor projections recovers the partition when $\Delta_*^2\gtrsim\log(K)^{1+\epsilon}$ for some $\epsilon>0$, almost matching $\Delta_{\mathrm{stat}}^2$, provided that $n\geq(dK)^{C/\epsilon}$ for some large and non-explicit numerical constant $C$. On the lower bound side, we recently established in~\cite{carpentier2026low} that, when $d\geq K$, low-degree polynomials fail as long as $\Delta_*^2\ll_{\log}\frac{K}{n^{1/4}}\wedge\sqrt K$, and that this non-parametric rate is achieved by a non-spectral procedure for the corresponding instances when $d=K$.

Statistical query lower bounds~\cite{SQclustering,pmlr-v195-diakonikolas23b} also exhibit statistical-computational gaps for clustering, but they rely on non-isotropic mixtures. For such mixtures, the computational barrier is much more severe when $K$ is large: if the covariance matrix is unknown and only bounded by $I_d$, any SQ procedure requires a complexity of order $d^{\Omega(1/\gamma)}$ for a separation $\Delta_*=K^{\gamma}$~\cite{pmlr-v195-diakonikolas23b}, so that, when $K$ grows polynomially with $d$, clustering with a polylogarithmic separation is out of reach for polynomial-time SQ procedures, whereas it is achievable in polynomial time for isotropic mixtures~\cite{LiuLi2022}.

The regime where the number $K$ of groups is much larger than the dimension $d$ remains largely open. There, $\sqrt K\geq\sqrt d$, so that distance-based and spectral procedures all require $\Delta_*^2\gtrsim_{\log}\sqrt d$, while, with a polylogarithmic separation, the only available procedures are tensor-based and require $n\geq(dK)^c$ for a large $c$. Conversely, the only low-degree lower bound in this regime is $\Delta^2_{\mathrm{comp,HD}}$ from~\cite{Even25}, which, for a polylogarithmic separation, only states that $n\gtrsim_{\log}dK^2$. This raises the following two questions:
\begin{itemize}
\item \emph{When the number $K$ of groups is large compared to the dimension $d$, what is the minimal separation $\Delta_*$, as a function of $n$, $K$ and $d$, for which clustering is possible in polynomial time?}
\item \emph{More specifically, in this regime, how large must the sample size $n$ be compared to $K$ and $d$ for clustering in polynomial time with only a polylogarithmic separation?}
\end{itemize}

\subsection{Our contribution}

Our results provide a partial answer to these questions on the lower bound side. For any fixed integer $m\geq 1$, we exhibit hard instances with $K=(d+1)^m$ groups, for which we establish a lower bound, as a function of $n$, $K$ and $d$, on the separation required by low-degree polynomials. Importantly, this separation is smaller than the $\sqrt{dK^2/n}$ bound that corresponds to the spectral threshold and its scaling with respect to $m$ reflects that higher-order tensors come into play.

\medskip

We construct a Bernoulli mixture with $2K$ groups, where $K=(d+1)^m$ for an integer $m\geq 1$ and $d=2^u-1$, whose centers $\pm X_k$ are, up to a random sign flip and a random permutation of the coordinates, the codewords of a dual BCH code~\cite[Chapter~9]{macwilliams1977theory}, rescaled by $\Delta/\sqrt d$ (Section~\ref{sec:main_results}). These centers are almost orthogonal, so that the separation $\Delta_*$ lies between $\Delta$ and $2\Delta$ as soon as $d\geq(4m)^{m+1}$ (Lemma~\ref{lem:BCH:separation}). The key property of this construction is that the moment tensors of the centers, and those of the observations, coincide with those of a Rademacher vector up to the order $2m+1$ (Lemma~\ref{lem:moments:XY}). In other words, the distribution of each observation cannot be distinguished from the uniform distribution on $\{-1,1\}^d$ through its moments of order at most $2m+1$. In particular, its covariance matrix is the identity, so that spectral methods based on the empirical covariance matrix are powerless, and the first moment tensor of the observations which carries information on the centers is of order $2m+2$, with entries of order $(\Delta/\sqrt d)^{2m+2}$.

Our main result, Theorem~\ref{thm:corr}, states that, for a fixed $m$, polynomials of degree $D$ polylogarithmic in $n$ do not estimate $x=\mathbf 1\{\kg(1)=\kg(2)\}$ better than the trivial estimator $1/K$, as long as
$$\Delta^2\ll_{\log}\Big(\frac{K^2d^{3/2}}{n}\Big)^{1/(2m+2)}\wedge\sqrt d\enspace ,$$
where $\ll_{\log}$ now also hides constants depending on $m$. Since $K=(d+1)^m$, the first term is of order $d/(n\sqrt d)^{1/(2m+2)}$, which is the form appearing in Condition~\eqref{eq:signalcond}. Besides, since $\Delta\leq\Delta_*\leq 2\Delta$, the same condition holds with $\Delta_*$ in place of $\Delta$. In particular, with a polylogarithmic separation, low-degree polynomials require at least $n\gtrsim_{\log}K^2d^{3/2}$ observations. This improves by a factor $\sqrt d$ over the bound $dK^2$ implied by~\cite{Even25}, and, more generally, Theorem~\ref{thm:corr} is stronger than the bound of~\cite{Even25} as soon as $\Delta^{4m}\ll_{\log}\sqrt d$. To the best of our knowledge, this is the first computational lower bound showing that, when $K\gg d$, clustering with a polylogarithmic separation requires a sample size that is polynomially larger than $dK^2$. For $m=1$, that is $K=d+1$, the number of groups is of the order of the dimension, a regime which is essentially covered by~\cite{carpentier2026low} with a stronger bound, of order $K^4/\Delta^8$ instead of $K^2d^{3/2}/\Delta^8$; the novelty of Theorem~\ref{thm:corr} therefore lies in the regime $m\geq 2$, that is $K\geq d^2$.

Let us compare this bound with the following heuristic. Since the moment tensors of order $2m+2$ of the observations carry the first non-trivial information, it is natural to estimate their entries, whose nonzero values are of order $(\Delta/\sqrt d)^{2m+2}$, by empirical averages over the $n$ observations. Detecting a single nonzero entry then requires $\sqrt n(\Delta/\sqrt d)^{2m+2}\gtrsim 1$, that is $n\gtrsim K^2d^2/\Delta^{4(m+1)}$, which is larger than our lower bound by a factor $\sqrt d$. Whether this remaining gap of order $\sqrt d$ can be closed, either by aggregating many entries of the empirical tensor or by improving the lower bound, is left open.

From a technical perspective, we follow the approach of~\cite{CGGV25,carpentier2025phase,carpentier2026low}, which consists in building a family of polynomials that is almost orthonormal under the distribution of the observations, and which is indexed here by bipartite graphs. As the polynomials only involve the algebraic structure of the BCH code through the number of solutions of polynomial systems over the finite field $\mathbb F_{d+1}$ (Lemma~\ref{lem:BCHeq}), the crux of our analysis is twofold. First, we relate the combinatorial properties of the graphs to the moment tensors of the centers. Since these tensors coincide with those of a Rademacher vector up to the order $2m+1$, the systems associated with many graphs have no solution (Lemmas~\ref{lem:scalprod} and~\ref{lem:number_solutions}). Second, we bound the number of solutions of the remaining systems through Bézout-type bounds and a lower bound on the rank of their Jacobian, in terms of the numbers of nodes, edges and connected components of the graph (Lemma~\ref{lem:eqcounting} and Section~\ref{sec:number_solution}).

\paragraph{From Bernoulli to Gaussian mixtures.} Although we consider a mixture of Bernoulli distributions, our setting can be turned into a Gaussian mixture model with almost the same separation, using the Gaussianization techniques of~\cite{ma2015computational,brennan2018reducibility,brennan2020reducibility}. This reduction only changes the separation threshold by a factor of order $\sqrt{\log(n)}$. %Write $\eta:=\Delta/\sqrt d\leq 1$. In the model of Section~\ref{sec:main_results}, conditionally on $\sigma$, $(a_j)$, $\kg$ and $(z_i)$, the entries $Y_{ij}\in\{-1,1\}$ are independent with mean $s_{ij}\eta$, where $s_{ij}:=z_ia_j\mu_{\kg(i),\sigma(j)}\in\{-1,1\}$. Given $T\geq 1$, let $\mu:=\eta/T$, let $\varphi_\pm$ be the density of $\mathcal N(\pm\mu,1)$, and let $\varepsilon:=\bbP[|\mathcal N(\mu,1)|>T]$. On $[-T,T]$, the functions $q_\pm:=\frac{\varphi_++\varphi_-}{2}\pm\frac{\varphi_+-\varphi_-}{2\eta}$ are nonnegative, because $\frac{\varphi_+-\varphi_-}{\varphi_++\varphi_-}(x)=\tanh(\mu x)$ and $|\tanh(\mu x)|\leq\mu T=\eta$. Each has mass $1-\varepsilon$ on $[-T,T]$, since $\varphi_+-\varphi_-$ is odd. Let $Q_\pm$ be the probability measure with density $q_\pm$ on $[-T,T]$ and an atom of mass $\varepsilon$ at $0$. Given $Y_{ij}=\pm1$, draw $\tilde Y_{ij}\sim Q_\pm$, independently over $(i,j)$. Since $\frac{1+s\eta}{2}q_++\frac{1-s\eta}{2}q_-=\varphi_s$ on $[-T,T]$ for $s\in\{-1,1\}$, the variable $\tilde Y_{ij}$ is distributed as $W\mathbf 1\{|W|\leq T\}$ with $W\sim\mathcal N(s_{ij}\mu,1)$, so that
%$$d_{TV}\big(\mathcal L(\tilde Y_{ij}\,|\,\sigma,(a_j),\kg,(z_i)),\ \mathcal N(s_{ij}\mu,1)\big)=\varepsilon\leq e^{-(T-1)^2/2}\leq n^{-3}\enspace ,$$
%when $T=\sqrt{c\log(n)}$ with $c=16$ and $n\geq 2$. Summing over the $nd\leq n^2$ entries, the joint distribution of $(x,\tilde Y)$ is at total variation distance at most $1/n$ from that of $(x,\bar Y)$. Here $\bar Y_i=z_iX'_{\kg(i)}+\xi_i$, with $\xi_i\sim\mathcal N(0,I_d)$ i.i.d., is a Gaussian mixture whose centers $X'_k:=X_k/T$ have norm $\Delta/\sqrt{c\log(n)}$, instead of $\Delta$ in our model. Hence, any procedure estimating $x$ in this Gaussian mixture model gives, after this randomized entrywise transformation, a procedure with essentially the same performance and running time in our Bernoulli model. Besides, since $Y_{ij}$ only takes two values, $\tilde Y_{ij}$ is an affine function of $Y_{ij}$ once the auxiliary randomness is fixed. Hence a degree-$D$ polynomial of $\tilde Y$ is a polynomial of degree at most $D$ of $Y$ whose coefficients are independent of $(x,Y)$, and Theorem~\ref{thm:corr} also bounds the low-degree correlation between $x$ and the entrywise truncated Gaussian observations $\tilde Y$. Since low-degree polynomials are not stable under small perturbations in total variation, this does not formally imply a low-degree lower bound for the untruncated observations $\bar Y$.

\subsection{Other related works}
\paragraph{Proof techniques for low-degree lower bounds.} For detection problems, low-degree lower bounds amount to bounding the norm of the low-degree projection of the likelihood ratio, which is explicit in many models~\cite{Hopkins17,KuniskyWeinBandeira}. No such closed form is available for the low-degree correlation~\eqref{eq:def:corr}. Schramm and Wein~\cite{SchrammWein22} bound $\mathrm{Corr}_{\leq D}$ in terms of joint cumulants of $x$ and of the entries of $Y$. This versatile approach has been applied, among others, to submatrix estimation~\cite{SchrammWein22}, graphon estimation~\cite{luo2023computational} and Gaussian mixture clustering~\cite{Even24}, but it may lead to involved computations and to thresholds that are only sharp up to polylogarithmic factors. It has been simplified and sharpened in latent variable models by conditioning on the latent variables~\cite{Even25}, and generalized by Sohn and Wein~\cite{SohnWein25}, who capture sharp thresholds such as the BBP transition in the spiked Wigner model and the Kesten--Stigum threshold in the stochastic block model. A different strategy, used by Montanari and Wein~\cite{montanari2024equivalence} in rank-one matrix estimation and systematized in~\cite{CGGV25}, consists in building a family of polynomials which is almost orthonormal under the distribution of the observations, so that $\mathrm{Corr}_{\leq D}^2$ is bounded, up to a factor close to one, by the sum of the squared correlations between $x$ and the elements of this family. It has been applied to stochastic block models~\cite{carpentier2025phase}, clustering~\cite{carpentier2026low} and ranking~\cite{carpentier2025rank}, and our proof follows this strategy. Finally, Li~\cite{li2026algorithmic} recently proposed a largely model-independent route, which turns any efficient recovery procedure into a test through a cross-validation argument, and then relies on low-degree bounds for this testing problem. The resulting lower bounds are however conditional on the low-degree conjecture for detection problems, whereas the above strategies, including ours, provide unconditional bounds on $\mathrm{Corr}_{\leq D}$.

\subsection{Organization of the manuscript}

In Section~\ref{sec:main_results}, we introduce the model, we state the two key properties of the construction, namely the matching of the moment tensors (Lemma~\ref{lem:moments:XY}) and the separation of the centers (Lemma~\ref{lem:BCH:separation}), and we state our main result. Section~\ref{sec:bch:code} describes the BCH construction and states its algebraic properties. All these properties are proved in Section~\ref{sec:proof:bch}. In Section~\ref{sec:discussion}, we discuss possible extensions. The proof of Theorem~\ref{thm:corr} is organized in Section~\ref{sec:graph}, and its main steps are proved in Section~\ref{sec:moments} and in the appendix.

\subsection{Notation}\label{sec:notation}

\paragraph{Sets and numbers.} For a positive integer $b$, we write $[b]:=\{1,\ldots,b\}$ and, for two integers $a\leq b$, $[a:b]:=\{a,a+1,\ldots,b\}$. The cardinality of a finite set $A$ is denoted by $|A|$. For an event or an assertion $\mathcal E$, $\mathbf 1\{\mathcal E\}$ is equal to $1$ if $\mathcal E$ holds and to $0$ otherwise. Given two sets $A$ and $B$, we write $A\oplus B:=(A\setminus B)\cup(B\setminus A)$ for their symmetric difference and, given finitely many sets $A_1,\ldots,A_r$, we write $\oplus_{i\leq r}A_i:=A_1\oplus\cdots\oplus A_r$ for the set of elements which belong to an odd number of the $A_i$'s. For two real numbers $a$ and $b$, we write $a\vee b:=\max(a,b)$ and $a\wedge b:=\min(a,b)$, and $\log$ stands for the natural logarithm.

\paragraph{Comparison of quantities.} For two positive quantities $a$ and $b$ depending on the parameters $n$, $d$, $K$ and $\Delta$ of the problem, we write $a\lesssim b$, or equivalently $b\gtrsim a$, when $a\leq Cb$ for some constant $C>0$ which does not depend on these parameters, and $a\asymp b$ when $a\lesssim b$ and $b\lesssim a$. The informal notation $a\ll b$ means that $a/b$ is small, and the subscript in $\lesssim_{\log}$, $\gtrsim_{\log}$ and $\ll_{\log}$ indicates that polylogarithmic factors and constants depending on $m$ are hidden as well. Finally, $O(\cdot)$ stands for the usual Landau notation.

\paragraph{Vectors and tensors.} For $x,y\in\mathbb R^d$, we write $\langle x,y\rangle:=\sum_{j\leq d}x_jy_j$ and $\|x\|_2:=\langle x,x\rangle^{1/2}$ for the Euclidean inner product and norm, and $I_d$ stands for the identity matrix of size $d$. For a positive integer $l$, the $l$-th tensor power of $x\in\mathbb R^d$ is the tensor $x^{\otimes l}$ of order $l$ whose entries are $(x^{\otimes l})_{j_1,\ldots,j_l}:=\prod_{s\leq l}x_{j_s}$, for $j_1,\ldots,j_l\in[d]$. For a random vector $Z\in\mathbb R^d$, the moment tensor $\bbE[Z^{\otimes l}]$ of order $l$ is defined entrywise.

\section{Model and main results}\label{sec:main_results}

Let $u$ and $m$ denote two positive integers. Define $d = 2^u-1$ and $K=(d+1)^m$. Our construction relies on a family $(\mu_k)_{k=1,\ldots, K}$ of vectors in $\{-1,1\}^{d}$ which, up to renormalization, correspond to the codewords of a dual BCH code of length $d$. As their precise definition requires some algebra over finite fields, we postpone it to Section~\ref{sec:bch:code}, see~\eqref{eq:definition_mu_k}. In this section, we introduce the mixture model built from the $\mu_k$'s and we only state, in Lemma~\ref{lem:moments:XY} below, the key properties of this construction that drive our lower bound.

Let $\sigma$ denote a uniformly random permutation on  $[d]$ and let $a_j$, $j=1,\ldots, d$, denote i.i.d.~Rademacher random variables, independent of $\sigma$.
For $\ell= 1,\ldots, K$, we define the vector $X_{\ell}\in \mathbb{R}^d$ by
\begin{equation}\label{eq:definition:Xlj}
X_{\ell,j} = \frac{\Delta}{\sqrt{d}} a_j \mu_{\ell,\sigma(j)}\enspace , \qquad \text{where } 0<\Delta\leq\sqrt d\ .
\end{equation}
This range of $\Delta$ is what makes $(1+z_iX_{\kg(i),j})/2$ a valid probability below, since $|\mu_{\ell,j}|=1$; it is in particular implied by Condition~\eqref{eq:signalcond}.

Let the random function $\kg: [n]\to [K]$ be such that $\kg(1),\ldots,\kg(n)$ are i.i.d.~$\mathrm{Unif}([K])$ and $z_i$, for $i=1,\ldots, n$, denote i.i.d.~Rademacher random variables, with $\sigma$, $(a_j)$, $\kg(\cdot)$, and $(z_i)$ mutually independent. Then, for $i=1,\ldots, n$, we observe the random vector $Y^*_i\in \{0,1\}^{d}$, which, conditionally on $\sigma$, $(a_j)$, $\kg(.)$, and $(z_i)$, follows a product distribution such that
\[
Y^*_{ij} \sim  \cB\Big(\frac{1+z_i X_{\kg(i),j}}{2}\Big)\ .
\]

Hence, given $\sigma$ and $(a_j)$, the $(Y^*_i)$ are independent observations from a mixture distribution with $2K$ groups with uniform weights. 
Our goal is to estimate the functional 
\begin{equation}\label{eq:definition_x}
x = \mathbf 1\{\kg(1) = \kg(2)\}\enspace . 
\end{equation}
For that we define
\begin{equation}
Y_{ij}= 2Y^*_{ij} -1
\end{equation}
and we write $Y_i=(Y_{i1},\ldots,Y_{id})\in\{-1,1\}^d$. Conditionally on $\sigma$, $(a_j)$, $\kg(i)$ and $z_i$, the coordinates of $Y_i$ are independent and $\bbE[Y_{ij}\,|\,\sigma,(a_j),\kg(i),z_i]=z_iX_{\kg(i),j}$, so that the $2K$ centers of the mixture are the vectors $\pm X_k$, $k\leq K$.
We now state the key properties of the BCH construction of the $\mu_k$'s: up to the order $2m+1$, the moment tensors of these centers, as well as those of the observations, are exactly those of a Rademacher random vector. This lemma is a variant of Lemma~\ref{lem:BCHdesign}, which is stated in Section~\ref{sec:bch:code}, where the $\mu_k$'s are defined; both lemmas are proved in Section~\ref{sec:proof:bch}.
\begin{lemma}\label{lem:moments:XY}
Let $U=(U_1,\ldots,U_d)$ be a vector of $d$ i.i.d.~Rademacher random variables. Recall that, for any positive integer $l$ and any $j_1,\ldots,j_l\in[d]$, $\bbE[U^{\otimes l}]_{j_1,\ldots,j_l}=\bbE[\prod_{s\leq l}U_{j_s}]$ is equal to $1$ if each value appears an even number of times in $(j_1,\ldots,j_l)$, and to $0$ otherwise. Then, for every $i\in[n]$ and every positive integer $l\leq 2m+1$, the following holds.
\begin{itemize}
\item[(i)] The order-$l$ tensor of the centers is that of a rescaled Rademacher vector:
\begin{equation}\label{eq:moments:X}
\bbE\Big[\big(z_iX_{\kg(i)}\big)^{\otimes l}\,\Big|\,\sigma,(a_j)\Big] = \frac{1}{2K}\sum_{k=1}^K\sum_{\varepsilon\in\{-1,1\}}\big(\varepsilon X_k\big)^{\otimes l} = \Big(\frac{\Delta}{\sqrt d}\Big)^l\,\bbE\big[U^{\otimes l}\big]\enspace .
\end{equation}
\item[(ii)] The order-$l$ tensor of the observation $Y_i$ is that of a Rademacher vector:
\begin{equation}\label{eq:moments:Y}
\bbE\big[Y_i^{\otimes l}\,\big|\,\sigma,(a_j)\big] = \bbE\big[U^{\otimes l}\big]\enspace .
\end{equation}
\end{itemize}
As the right-hand sides of~\eqref{eq:moments:X} and~\eqref{eq:moments:Y} do not depend on $\sigma$ and $(a_j)$, these two properties also hold unconditionally.
\end{lemma}

We now show that the vectors $\mu_k$ are almost orthogonal, so that the $2K$ centers $\pm X_k$ of the mixture are well separated, as long as $m$ is small compared to $\log(d)/\log\log(d)$. This lemma is also proved in Section~\ref{sec:proof:bch}.
\begin{lemma}\label{lem:BCH:separation}
Assume that $d\geq 2m$ and let
$$\eta_m:=\Big(\frac{4m^{m+1}}{d}\Big)^{1/(2m)}\enspace .$$
Then, for all $k\neq k'\in[K]$, we have $|\langle\mu_k,\mu_{k'}\rangle|\leq\eta_m d$. Consequently,
$$2(1-\eta_m)\,d\leq \|\mu_k-\mu_{k'}\|_2^2\leq 2(1+\eta_m)\,d\enspace ,\qquad\text{and}\qquad \|\mu_k+\mu_{k'}\|_2^2\geq 2(1-\eta_m)\,d\enspace .$$
In particular, if $d\geq(4m)^{m+1}$, then $\eta_m\leq 1/2$, so that
$\min_{k\neq k'}\|\mu_k-\mu_{k'}\|_2\geq\sqrt d\enspace$, 
and the $2K$ centers $\pm X_k$, $k\leq K$, are pairwise at Euclidean distance at least $\Delta$.
\end{lemma}
Since $K=(d+1)^m$, the condition $d\geq(4m)^{m+1}$ holds as soon as $d\geq e^{e^2}$ and $K\leq d^{\log(d)/(2\log\log(d))}$, so that the number $K$ of groups can be much larger than any fixed power of the dimension $d$.

Lemma~\ref{lem:BCH:separation} ensures that the mixture is well separated. Write $\Delta_*$ for its separation, that is the minimal Euclidean distance between two distinct centers $\varepsilon X_k$ and $\varepsilon' X_{k'}$, with $\varepsilon,\varepsilon'\in\{-1,1\}$ and $k,k'\in[K]$. Since $\|X_k\|_2=\Delta$ and $\|\varepsilon X_k-\varepsilon'X_{k'}\|_2^2=\frac{\Delta^2}{d}\big(2d-2\varepsilon\varepsilon'\langle\mu_k,\mu_{k'}\rangle\big)$ for $k\neq k'$, Lemma~\ref{lem:BCH:separation} gives
$$\sqrt{2(1-\eta_m)}\,\Delta\leq\Delta_*\leq\sqrt{2(1+\eta_m)}\,\Delta\enspace .$$
In particular, $\Delta\leq\Delta_*\leq 2\Delta$ as soon as $d\geq(4m)^{m+1}$, and $\Delta_*=\sqrt2\,\Delta\,(1+o(1))$ when $m$ is not too large, namely when $m=o\big(\log(d)/\log\log(d)\big)$. Hence, up to numerical constants, the separation of the mixture is of the order of $\Delta$, and Theorem~\ref{thm:corr} below can be read in terms of $\Delta_*$.

In the following, we bound the low-degree correlation~\eqref{eq:def:corr} between the functional $x$ defined in~\eqref{eq:definition_x} and the polynomials of degree at most $D$ in $Y=(Y_{ij})_{i\in[n],j\in[d]}$, namely
\[
\mathrm{Corr}_{\leq D}=\sup_{P\in\mathbb R_D[Y]:\ \bbE[P^2]\leq 1}\bbE[xP]\enspace .
\]
Since the constant polynomial $P=1$ is admissible, we always have $\mathrm{Corr}_{\leq D}\geq\bbE[x]=1/K$.

Assume that
\begin{align}\tag{$\mathcal{H}_{signal}$}\label{eq:signalcond}
\rho:= \Big[(32 D^3 d^{1/4} n^{1/2})^{1/(2m+2)} \Big(\frac{4m\Delta}{\sqrt{d}}\Big)\Big]\vee \Big[(32 D^3)\frac{4m\Delta}{d^{1/4}}\Big]  \leq D^{-16c_{\mathrm{s}}} \leq 1\enspace.
\end{align}

Under this condition, we bound the low-degree correlation as follows.
\begin{theorem}\label{thm:corr}
Let $D$ be a positive integer, and assume that $2^8D^4\leq d$, $n\geq d$, and that Condition~\eqref{eq:signalcond} holds for some $c_{\mathrm{s}} \geq 10$. Then, we have
    $$
    \mathrm{Corr}^2_{\leq D}\leq  \frac{1}{K^2}\left(1+ \frac{16}{D^{c_{\mathrm{s}}/2}} + \frac{16D^2}{d}\right)\enspace .$$
\end{theorem}
Since $\mathrm{Corr}_{\leq D}\geq 1/K$, Theorem~\ref{thm:corr} entails that $\mathrm{Corr}_{\leq D}=(1+o(1))/K$ as soon as $D\to\infty$. Hence, no polynomial of degree at most $D$ predicts whether the first two observations belong to the same group significantly better than the constant $1/K$, which ignores the data: low-degree polynomials do not improve over a random guess. Regarding the rates, for $D$ of order $\log(n)$ and since $K=(d+1)^m$, Condition~\eqref{eq:signalcond} amounts to
$$\Delta^2\ll_{\log}\frac{d}{(n\sqrt d)^{1/(2m+2)}}\asymp\Big(\frac{K^2d^{3/2}}{n}\Big)^{1/(2m+2)}\qquad\text{and}\qquad\Delta^2\ll_{\log}\sqrt d\enspace ,$$
where $\ll_{\log}$ hides polylogarithmic factors and constants depending on $m$. The second condition excludes the regime where distance-based procedures succeed. The first one entails that, with a polylogarithmic separation, low-degree polynomials require $n\gtrsim_{\log}K^2d^{3/2}$, which improves by a factor $\sqrt d$ over the bound $dK^2$ of~\cite{Even25}, and is smaller by a factor $\sqrt d$ than the sample size at which the empirical tensor of order $2m+2$ becomes informative (Section~\ref{sec:discussion:upper}).

\section{Definition and Properties of the $\mu_k$'s}\label{sec:bch:code}

We detail here the construction of the means $\mu_{\ell}$'s used in our model. Our construction corresponds, up to renormalization, to codewords of the dual of the binary Bose--Chaudhuri--Hocquenghem  (BCH) code of length $d$ and designed distance $2m+1$, see e.g.~\cite{macwilliams1977theory} for a textbook account.
For the sake of completeness, we nevertheless provide a self-contained definition of these vectors, together with self-contained proofs of the properties used in this work, without relying on coding theory.
Recall that $u$ and $m$ denote two positive integers. Write $d = 2^u-1$ and $K=(d+1)^m$.  
\subsection{Definition}

 First, we recall one construction of the finite field $\mathbb{F}_{2^{u}}$. We consider $\mathbb F_2[T]$, namely the set of polynomials with binary coefficients, and we then quotient this set by a primitive polynomial\footnote{A polynomial $f\in\mathbb F_2[T]$ of degree $u$ is primitive if it is irreducible and if the class of $T$ generates the multiplicative group of the field $\mathbb F_2[T]/f(T)$, that is, if the smallest positive integer $q$ such that $f(T)$ divides ${T}^{{q}}-1$ is $q$$=2^{u}-1$. Primitive polynomials exist for every degree $u\geq 1$, see e.g.~\cite[Chapter~4]{macwilliams1977theory}.} $f(T)$ of degree $u$. Since $f$ is irreducible, the quotient $\mathbb F_2[T]/f(T)$ is a field with $2^{u}$ elements, and we henceforth identify it with the finite field $\mathbb F_{2^{u}}$. The underlying set of $\mathbb F_{2^{u}}$ is furthermore in bijection with the set of integers $\{0,\ldots,2^{u}-1\}$ and with the product set $\mathbb F_2^{u}$: to a polynomial of $\mathbb F_2[T]/f(T)$ we associate, on the one hand, the vector of its ordered coefficients, which is an element of $\mathbb F_2^{u}$, and on the other hand its corresponding value when evaluated in base $2$, which is an element of $\{0,\ldots,2^{u}-1\}$. Throughout the paper, we use the resulting bijection to \emph{identify} the set of integers $\{0,\ldots,2^u-1\}$ with the field $\mathbb F_{2^u}$. This identification only concerns the coordinate labels $j\in[d]=\{1,\ldots,d\}$, which are thereby identified with the nonzero elements of $\mathbb F_{2^u}$, and, in Section~\ref{sec:number_solution}, the variables $x_w$: whenever we add or multiply such elements --~and, in particular, whenever we raise such an element to some power~-- the operations are those of the field $\mathbb F_{2^u}$, and not the operations of $\mathbb N$. All the other arithmetic operations, such as $k-1$, $d+1$ or $2a-1$, are the usual operations on integers.

 Let us emphasize that the set $\mathbb F_2^{u}$ plays a different role: it is only used as an \emph{additive} group. Indeed, the bijection $\mathbb F_{2^{u}}\to \mathbb F_2^{u}$ described above is an isomorphism of $\mathbb F_2$-vector spaces --~hence, in particular, an isomorphism of additive groups, the addition on $\mathbb F_2^{u}$ being the coordinatewise addition modulo $2$~-- but it is \emph{not} a field isomorphism, and the product set $\mathbb F_2^{u}$ endowed with the coordinatewise multiplication is not even a field as soon as $u\geq 2$, since e.g.~$(1,0,\ldots,0)$ and $(0,1,0,\ldots,0)$ are nonzero elements with product $0$. Accordingly, only the additive structure of $\mathbb F_2^{u}$ is ever used below, all multiplicative operations being performed in $\mathbb F_{2^{u}}$.
 
 In what follows, we write
 $$\psi:\{0, \ldots, 2^u-1\} \rightarrow \mathbb{F}_{2}^u,$$
 for the bijection that associates to any element in $\{0, \ldots, 2^u-1\}$ its representation in base $2$, as an element of $\mathbb{F}_{2}^u$.
 In other words, upon identifying $\{0,\ldots,2^u-1\}$ with $\mathbb F_{2^u}$ as explained above, the map $\psi$ is exactly the isomorphism of $\mathbb F_2$-vector spaces from $\mathbb F_{2^u}$ onto $\mathbb F_2^u$ described above. In particular, $\psi$ is additive: $\psi(v+v')=\psi(v)+\psi(v')$ for all $v,v'\in\mathbb F_{2^u}$, where the left-hand sum is taken in $\mathbb F_{2^u}$ and the right-hand one coordinatewise modulo $2$; and $\psi(v)=0$ if and only if $v=0$. We shall only use these additive properties of $\psi$.
 We also write
 $$\bar \psi:\{0, \ldots, 2^{um}-1\} \rightarrow \{0,1\}^{m\times u},$$
for the bijection that associates to any $k \in \{0, \ldots, 2^{um}-1\}$ its representation in base $2$, but transformed from its vector representation in dimension $um$ to the associated matrix representation of dimension $m\times u$. More precisely, we consider the base-$2$ representation as a vector, and we arrange its $um$ coordinates into $m$ rows of size $u$ (row by row).

Let $v\in \{0, \ldots, 2^u-1\}$, which we identify, as above, with an element of the finite field $\mathbb{F}_{2^u} = \mathbb{F}_{d+1}$. We write for any integer $l \geq 1$,
\begin{equation}\label{eq:definition_phi}
\phi^{(l)}(v) = \psi\left(v^{l}\right)\ \in \mathbb F_2^u\enspace ,
\end{equation}
where $v^l$ stands for the $l$-th power of $v$ \emph{in the field} $\mathbb F_{2^u}$, and where $\psi$ then transforms this field element into the associated binary code in $\mathbb F_2^u$. Let us stress that the power operation in~\eqref{eq:definition_phi} is the one of the field $\mathbb F_{2^u}$; it is \emph{not} a coordinatewise power of the binary vector $\psi(v)$ --~which would be meaningless, as $\psi(v)^l=\psi(v)$ for every $l\geq 1$ when powers are taken coordinatewise in $\{0,1\}$. Note also that $\phi^{(l)}(0)=\psi(0)=0$ for every $l\geq 1$.
%for the transformation of $v^{l}$ - where the power operation is here taken in $\mathbb{F}_{2^u}$, namely the regular power operation in $\mathbb N$, modulo $2^u-1$ - into the associated binary code in $\mathbb{F}_{2}^u$.% - namely, for the transformation of $\alpha^{l'}$ into its binary code.

Define for $v \in \{0, \ldots, 2^u-1\}$:
    \begin{equation}\label{eq:BCH}
        H_{v}(R) = \prod_{a=1}^m \prod_{b=1}^u (R_b^{(a)})^{[\phi^{(2a-1)}(v)]_b},
    \end{equation}
    where $R = (R_b^{(a)})_{b \leq u, a\leq m}$ is such that $R_b^{(a)} \in \{-1,1\}$.

%For such a matrix $R$ of size $u\times u$, we write $\bar \psi(R)$ for a vectorisation of $(R+1)/2$. This vector is an element of $\mathbb{F}_{2}^{um}$. 

%For any $j \in \mathbb{F}_{2^{um}}$ write 
%$$\bar \psi(j),$$
%for the transformation of the vector $j$ into the associated element in $\mathbb{F}_{2}^{um}$ - namely, for the transformation of $j$ into its binary code. We then transform

We write for any $k \in \{1,\ldots,2^{um}\} = \{1,\ldots,(d+1)^m\} = \{1,\ldots,K\}$ and any $j \in \{1,\ldots,2^u-1\} = \{1,\ldots,d\}$ -- so that $j$ ranges over the $d$ \emph{nonzero} elements of $\mathbb F_{2^u}$, excluding the element $0$ for which $\phi^{(l)}(0)=0$ would otherwise yield a trivial (constant) coordinate:
\begin{align}\label{eq:definition_mu_k}
\mu_{k,j} = H_{j}\Bigg(\Big(2(\bar \psi(k-1))_{a,b}-1\Big)_{a\leq m, b\leq u}\Bigg)\enspace .
\end{align}

%Here for any positive integer $p$, we identify any integer in resp.~$\{0,\ldots, 2^p-1\}$ with its associated element in $\mathbb F_{2^p}$.

\paragraph{Intuition.} Since $\bar\psi$ is a bijection, when $k$ is uniformly distributed on $[K]$, the matrix $R:=\big(2(\bar\psi(k-1))_{a,b}-1\big)_{a\leq m,b\leq u}$ has i.i.d.~Rademacher entries. By~\eqref{eq:BCH} and~\eqref{eq:definition_mu_k}, each coordinate $\mu_{k,j}=H_j(R)$ is the product of the entries of $R$ selected by the binary vector $h(j):=\big(\phi^{(1)}(j),\phi^{(3)}(j),\ldots,\phi^{(2m-1)}(j)\big)\in\mathbb F_2^{m\times u}$. Hence, given a positive integer $l$ and indices $j_1,\ldots,j_l\in[d]$, not necessarily distinct, the product $\prod_{i=1}^{l}\mu_{k,j_i}$ is again a product of entries of $R$, whose expectation with respect to $k\sim\mathrm{Unif}([K])$, namely $\frac{1}{K}\sum_{k\leq K}\prod_{i=1}^{l}\mu_{k,j_i}=\big(\frac1K\sum_{k\leq K}\mu_k^{\otimes l}\big)_{j_1,\ldots,j_l}$, is equal to $1$ if each entry appears an even number of times, that is if $\sum_{i=1}^{l}h(j_i)=0$ in $\mathbb F_2^{m\times u}$, and to $0$ otherwise: this is Lemma~\ref{lem:BCHeq} below. In comparison, for a Rademacher vector $U$, $\bbE[\prod_{i=1}^{l}U_{j_i}]=1$ if and only if each index appears an even number of times in $(j_1,\ldots,j_l)$. Therefore, matching the moment tensors of $U$ up to the order $2m$ amounts to requiring that any $2m$ of the vectors $h(1),\ldots,h(d)$ are linearly independent over $\mathbb F_2$. This is where the choice of the odd powers $v,v^3,\ldots,v^{2m-1}$ comes into play. Identifying the indices $j_1,\ldots,j_l\in[d]$ with nonzero elements $v_1,\ldots,v_l$ of $\mathbb F_{2^u}$, the condition $\sum_{i=1}^{l}h(j_i)=0$ means that the $m$ odd power sums $\sum_{i=1}^{l}v_i^{2a-1}$, $a\leq m$, vanish in $\mathbb F_{2^u}$. In characteristic $2$, we have\footnote{Observe that $(v_1+v_2)^2= v_1^2 + 2v_1v_2+ v_2^2 = v_1^2 + v_2^2$} $\sum_{i=1}^{l}v_i^{2k}=(\sum_{i=1}^{l}v_i^{k})^2$ for any $k\geq 1$, so that the first $2m$ power sums $\sum_{i=1}^{l}v_i^{k}$, $k\leq 2m$, then vanish as well, which is impossible when $1\leq l\leq 2m$ and $v_1,\ldots,v_l$ are pairwise distinct (Lemma~\ref{lem:fundBCH} in Section~\ref{sec:proof:bch}). This is why $m$ blocks of $u$ bits, that is $K=(d+1)^m$ vectors, suffice to match the moment tensors of $U$ up to the order $2m$, which is Lemma~\ref{lem:BCHdesign} below.

\subsection{Properties}

We state here the two properties of the BCH construction on which our analysis relies. Lemma~\ref{lem:BCHeq} expresses the moment tensors of the $\mu_k$'s, of any order, through the polynomial system $\sum_{i}\phi^{(2a-1)}(j_i)=0$, $a\leq m$, whose solutions are studied in Section~\ref{sec:number_solution}, and Lemma~\ref{lem:BCHdesign} states that these tensors coincide with those of a Rademacher vector up to the order $2m$. Apart from Lemma~\ref{lem:BCH:separation}, which ensures that the mixture is well separated, and from the fact that the $\mu_k$'s belong to $\{-1,1\}^d$, these are the only properties of the $\mu_k$'s that we use. Their proofs, as well as the proof of Lemma~\ref{lem:moments:XY}, are postponed to Section~\ref{sec:proof:bch}.

\begin{lemma}\label{lem:BCHeq}
    We have for any positive integer $l$
    \begin{align*}
       \Big(\frac{1}{K}\sum_{k \leq K} \mu_k^{\otimes l}\Big)_{j_1,\ldots, j_l}  = \mathbf 1\Big\{\forall a \leq m : \sum_{i=1}^l \phi^{(2a-1)}(j_i)=0\Big\},
    \end{align*} 
    where the sum in the equation in the indicator is taken in $\mathbb F_{2}^u$ - and therefore corresponds to a coordinate-wise sum of the $u$-dimensional vectors, modulo $2$.
\end{lemma}

\begin{lemma}\label{lem:BCHdesign}
    We have for any $l \leq 2m$ that
    $$\frac{1}{K}\sum_{k \leq K} \mu_k^{\otimes l} = \mathbb E_{U\sim \mathcal U\Big(\{-1,1\}^d\Big)} \Big[U^{\otimes l}\Big] =: \mathcal U^{(l)},$$
    where $\mathcal U(\{-1,1\}^d)$ is the uniform distribution over the set $\{-1,1\}^d$. Note that $ \mathcal U^{(l)}_{j_1,\ldots, j_l}$ is equal to $1$ if and only if there exists a pairing\footnote{By this, we mean that the entries in the vector $(j_1,\ldots, j_l)$ can be mapped bijectively into $l/2$ pairs $(j_{J_i},j_{J_i'})_{i\leq l/2}$ such that $j_{J_i} = j_{J_i'}$ for any $i$, where $\{J_i,J'_i,i\leq l/2\}$ is in bijection with $\{1,\ldots,l\}$. Note that this is only (sometimes) possible if $l$ is even.} of the vector $(j_1,\ldots, j_l)$, and otherwise it is equal to $0$. In this case, we say that $(j_1,\ldots, j_l)$ is paired.
\end{lemma}

\begin{remark}
Lemma~\ref{lem:BCHdesign} deals with the tensors $\frac{1}{K}\sum_{k\leq K}\mu_k^{\otimes l}$ up to the order $2m$ only. The additional order $2m+1$ in Lemma~\ref{lem:moments:XY} is due to the signs $z_i$: the set of centers $\{\pm X_k\}$ is symmetric, so that all its odd-order tensors vanish. 
%Without these signs, the order $2m+1$ is not matched in general: for $m=1$ and $u\geq 2$, if $v_1,v_2\in\mathbb F_{2^u}\setminus\{0\}$ are distinct and $v_3=v_1+v_2$, then $v_3\notin\{0,v_1,v_2\}$ and Lemma~\ref{lem:BCHeq} gives $\big(\frac1K\sum_{k\leq K}\mu_k^{\otimes 3}\big)_{v_1,v_2,v_3}=1$, whereas $\bbE[U^{\otimes 3}]=0$. Similarly, the order $2m+1$ in Lemma~\ref{lem:moments:XY} cannot be replaced by $2m+2$ in general: for $m=1$ and $u\geq 3$, choosing distinct $v_1,v_2,v_3\in\mathbb F_{2^u}\setminus\{0\}$ with $v_3\neq v_1+v_2$, and $v_4=v_1+v_2+v_3$, the four elements $v_1,\ldots,v_4$ are nonzero and pairwise distinct, and Lemma~\ref{lem:BCHeq} gives $\big(\frac1K\sum_{k\leq K}\mu_k^{\otimes 4}\big)_{v_1,\ldots,v_4}=1$. Hence, with $j_s=\sigma^{-1}(v_s)$, the entries at $(j_1,\ldots,j_4)$ of the left-hand sides of~\eqref{eq:moments:X} and~\eqref{eq:moments:Y} for $l=4$ are both equal to $(\Delta/\sqrt d)^4\prod_{s\leq 4}a_{j_s}\neq 0$, whereas $\bbE[U^{\otimes 4}]_{j_1,\ldots,j_4}=0$.
\end{remark}

\section{Discussion}\label{sec:discussion}

\paragraph{Upper bound for the BCH instances.}\label{sec:discussion:upper}
For the instances of Theorem~\ref{thm:corr}, the tensor of order $2m+2$ of the observations can be estimated exactly. Indeed, conditionally on $\sigma$ and $(a_j)$, Lemma~\ref{lem:BCHeq} and the conditional independence of the coordinates of $Y_i$ imply that, for pairwise distinct $j_1,\ldots,j_{2m+2}\in[d]$,
$$\bbE\Big[\prod_{s\leq 2m+2}Y_{ij_s}\,\Big|\,\sigma,(a_j)\Big]=\Big(\frac{\Delta}{\sqrt d}\Big)^{2m+2}\prod_{s\leq 2m+2}a_{j_s}\ \mathbf 1\Big\{\forall a\leq m:\ \sum_{s\leq 2m+2}\phi^{(2a-1)}(\sigma(j_s))=0\Big\}\enspace .$$
The empirical averages $\frac1n\sum_{i\leq n}\prod_{s}Y_{ij_s}$ fluctuate at the scale $n^{-1/2}$. By Hoeffding's inequality and a union bound over the at most $d^{2m+2}$ tuples, thresholding them recovers, with high probability, both the support of this tensor and the signs $\prod_s a_{j_s}$, as soon as
$$n\gtrsim m\log(d)\Big(\frac{d}{\Delta^2}\Big)^{2m+2}\asymp m\log(d)\,\frac{K^2d^2}{\Delta^{4(m+1)}}\enspace .$$
We conjecture that the centers $\pm X_k$ of the mixture can be recovered in polynomial time from this estimated tensor and that, in turn, the observations can be clustered, for instance by assigning each of them to the center $\pm X_k$ which maximizes its scalar product, as soon as $\Delta^2\gtrsim m\log(d)$. For these instances, clustering would then be possible in polynomial time as soon as $n\gtrsim_{\log}K^2d^2/\Delta^{4(m+1)}$ and $\Delta^2\gtrsim m\log(d)$, which differs from the lower bound of Theorem~\ref{thm:corr} by a factor $\sqrt d$ in $n$.

Even if this conjecture holds, such a procedure would heavily rely on the algebraic structure of the instances, namely on the knowledge that the centers form a permuted and sign-flipped dual BCH code. It is therefore not clear whether a sample size of order $K^2d^2$, or $K^2d^{3/2}$, characterizes the optimal complexity of clustering over all mixtures whose centers are $\Delta$-separated. On the one hand, a procedure that is agnostic to the structure of the centers may require more observations. On the other hand, other configurations of centers may lead to harder instances.

\paragraph{On Gaussian designs.}\label{sec:discussion:designs}
By Lemmas~\ref{lem:BCHdesign} and~\ref{lem:moments:XY}, the uniform distribution on $\{\mu_k,\ k\leq K\}$ (resp.\ on $\{\pm\mu_k,\ k\leq K\}$) has the same moment tensors as a vector $U$ of $d$ i.i.d.~Rademacher random variables, up to the order $2m$ (resp.\ $2m+1$). To introduce a tentative worst-case instance for Gaussian mixture models, we would rather use a design whose moment tensors match those of the standard normal distribution, that is a set $\{\xi_1,\ldots,\xi_K\}\subset\mathbb R^d$ such that
$$\frac{1}{K}\sum_{k\leq K}\xi_k^{\otimes l}=\bbE\big[G^{\otimes l}\big]\enspace ,\qquad\text{for all } l\leq 2m+1\enspace ,\qquad\text{where } G\sim\mathcal N(0,I_d)\enspace .$$
For general $m$, the minimal size of such Gaussian designs remains largely unknown~\cite{dillon2025fixed}, even when the uniform weights $1/K$ are replaced by positive weights, which amounts to a mixture with unequal proportions. Such a design has at least $\binom{d+m}{m}\geq d^m/m!$ points~\cite{stroud1971approximate}, which is of the same order as $K=(d+1)^m$, whereas the known constructions with positive weights have $O(d^{2m-1})$ points~\cite{kuperberg2006numerical}. This is the reason why we focused our interest in this manuscript on the Bernoulli case. 

\paragraph{Towards a direct low-degree proof in the Gaussian case.}\label{sec:discussion:gaussian}

The reduction described in the introduction transfers Theorem~\ref{thm:corr} to Gaussian mixtures at the price of a factor $\sqrt{\log(n)}$ in the separation, and the resulting low-degree bound only applies to entrywise truncated Gaussian observations. We believe that the strategy of this paper also gives a direct proof for the Gaussian mixture $\bar Y_i=z_iX_{\kg(i)}+\xi_i$, where the $\xi_i\sim\mathcal N(0,I_d)$ are independent and the centers $\pm X_k$ are those of Section~\ref{sec:main_results}. In the Bernoulli model, $Y_{ij}^2=1$, so that it suffices to consider multilinear polynomials, indexed by simple bipartite graphs. In the Gaussian model, an entry $\bar Y_{ij}$ may appear with any power, so that the polynomials should be indexed by bipartite multigraphs. An edge $e=(v,w)$ with multiplicity $\beta_e$ is then associated with the Hermite polynomial $\mathrm{He}_{\beta_e}(\bar Y_{\pi_1(v)\pi_2(w)})$ instead of the monomial $Y^{\beta_e}_{\pi_1(v)\pi_2(w)}$, similarly to what is done in~\cite{carpentier2026low} for Gaussian mixtures with orthogonal centers, where the polynomials are however indexed by multigraphs on the individuals.

%Since $\bbE[\mathrm{He}_\beta(s+\xi)]=s^\beta$ for $\xi\sim\mathcal N(0,1)$, and since $z_i^2=a_j^2=\mu_{k,j}^2=1$, the first moment computations of Section~\ref{sec:moments} carry over: for a multigraph whose degrees, counted with multiplicity, are all even, the first moment is equal to $(\Delta/\sqrt d)^{\sum_e\beta_e}$ times the probability that the BCH system~\eqref{eq:startsys} holds for the simple graph made of the edges of odd multiplicity, the edges of even multiplicity only contributing a deterministic factor. The main change lies in the second moments. A product of two Hermite polynomials of the same entry is not a Hermite polynomial, but a linear combination of Hermite polynomials of lower degrees, so that the merged, intersection and symmetric difference graphs of Section~\ref{sec:moments} have to be replaced by a sum over the possible ways of pairing the multi-edges, as in~\cite{carpentier2026low}. We conjecture that this Hermite version of our construction gives a direct proof that the conclusion of Theorem~\ref{thm:corr} holds for the Gaussian mixture, under Condition~\eqref{eq:signalcond} with possibly different numerical constants, and without losing a logarithmic factor in the separation.

\section{Proof organization for Theorem~\ref{thm:corr}}\label{sec:graph}

In this section, we introduce the main tools towards establishing Theorem~\ref{thm:corr}. Parametrizing polynomials of Boolean variables by graphs is now standard in the analysis of low-degree polynomials~\cite{Hopkins17,SchrammWein22,CGGV25}. First, we build upon the invariance of the distribution of $(x,Y)$ under permutations of the observations and of the coordinates, and under the symmetry of the random signs, to reduce the space of polynomials: we introduce a basis of permutation-invariant polynomials indexed by a family of bipartite graphs, and we show that graphs with a node of odd degree can be left out. Then, we establish that a suitable sub-family of these polynomials is almost orthogonal. 
In turn, this allows us to establish an explicit expression of the low-degree correlation $\mathrm{Corr}_{\leq D}$.

\subsection{Bipartite graph formalism and invariant polynomial basis}

Throughout this work, we restrict our attention to simple and undirected bipartite graphs. In the sequel, we write $G = (V,W,E)$ where $V$ and $W$ are two disjoint node sets and $E\subset V\times W$.  Given an edge $e\in V\times W$, we respectively write $(l(e),r(e))$ for the corresponding nodes in $V$ and $W$. Given a node $v$ of $G$, we write $\cN_G(v)$ for the set of neighbors of $v$ in $G$, and $\mathrm{deg}_G(v):=|\cN_G(v)|$ for the degree of $v$; the node $v$ is isolated when $\mathrm{deg}_G(v)=0$. Given two nodes $v$ and $w$, we write $v\sim_{G} w$ when $v$ and $w$ are adjacent in $G$. When there is no possibility of confusion, we simply write $\mathcal{N}(v)$, $\mathrm{deg}(v)$ and $v\sim w$.

Given a set $V=\{v_1,\ldots, v_r\}$ of size at least $2$, we denote by $\overline{\Pi}_{V}$  the collection of all injective maps $\pi_1$ from $V$ to $[n]$ such that $\pi_1(v_1)=1$ and $\pi_1(v_2)=2$. Given a set $W$ of the form $W=\{w_1,\ldots, w_s\}$ with $s\geq 1$, we denote by $\Pi_{W}$ the collection of all injective maps $\pi_2$ from $W$ to $[d]$.

Given a graph $G$, $\pi_1\in \overline{\Pi}_V$, and $\pi_2\in \Pi_W$, we define the multivariate polynomial $P_{G,\pi_1,\pi_2}$ by 
\begin{equation}\label{eq:P}
P_{G,\pi_1,\pi_2}= \prod_{e\in E} Y_{\pi_1(l(e)),\pi_2(r(e))}\enspace . 
\end{equation}
Two bipartite graphs $G=(V,W,E)$ and $G'=(V',W',E')$ are said to be equivalent if there exist bijections $\tau_1$ from $V$ to $V'$ such that $\tau_1$$(v_1)=v'_1$ and $\tau_1$$(v_2)=v'_2$, and $\tau_2$ from $W$ to $W'$ such that the pair $(\tau_1,\tau_2)$ maps $E$ to $E'$, that is $E'=\{(\tau_1(l(e)), \tau_2(r(e))): e\in E\}$.

Let $\mathcal{G}_{\leq D}$ denote a set of representatives of the equivalence classes of graphs $G=(V,W,E)$ with $v_1,v_2\in V$, with at least one and at most $D$ edges, and such that each node $v\in V\setminus\{v_1,v_2\}$ and each node $w\in W$ is not isolated.

Since $Y_{ij}^2=1$, we identify $\mathbb R_D[Y]$ with the space of multilinear polynomials of degree at most $D$ in the entries of $Y$. Then, $1$ and the distinct polynomials $P_{G,\pi_1,\pi_2}$, with $G\in \mathcal{G}_{\leq D}$, $\pi_1\in \overline{\Pi}_V$ and $\pi_2\in \Pi_W$, form a basis of $\mathbb R_D[Y]$. Define the polynomials $P_{G}$ by summing over all labelings $\pi_1$ and $\pi_2$
\[
P_{G} = \sum_{\pi_1 \in \overline{\Pi}_V} \sum_{\pi_2 \in \Pi_W}P_{G,\pi_1,\pi_2} \ .
\]
We readily check that $(1,(P_G)_{G\in \mathcal{G}_{\leq D}})$ is a basis of the subspace of polynomials in $\mathbb R_D[Y]$ that are invariant under permutations of the $d$ columns and of the rows $i\in [3:n]$. The following lemma is a consequence of the permutation invariance of the distribution of $Y$.
\begin{lemma}\label{lem:reduction:p}
We have 
\[
\mathrm{Corr}_{\leq D}= \sup_{P\in \mathrm{span}\big(1, (P_{G})_{G\in \mathcal{G}_{\leq D}}\big):\ \mathbb{E}[P^2]\leq 1} \mathbb{E}[xP]\enspace .
\]
\end{lemma}

\subsection{Correction and reduction to even-degree graphs}

Next, we establish that we can leave out graphs $G$ that contain at least one node whose degree is odd. First, the following lemma states that the corresponding polynomials are uncorrelated with the functional $x$.

\begin{lemma}\label{lem:first_moment_bch}
Consider any graph $G=(V,W,E)$ and any labeling $\pi=(\pi_1,\pi_2)$ in $\overline{\Pi}_V\times \Pi_W$. We have $\mathbb E[P_{G,\pi}] = 0$ and $\mathbb E[xP_{G,\pi}] = 0$  if at least one of the following two conditions holds: 
\begin{itemize}
    \item[(a)] The degree of a node $w\in W$ is odd.
    \item[(b)] The degree of a node $v\in V $ is odd.
\end{itemize}
\end{lemma}
Condition (a) is a consequence of the symmetry of the Rademacher variables $(a_j)$ in the definition~\eqref{eq:definition:Xlj} of $X_{l,j}$. Similarly, Condition (b) follows from the symmetry of the Rademacher random variables $(z_i)$ in the definition of $Y_i$.

\medskip 

A graph $G=(V,W,E)\in \cG_{\leq D}$ is said to be of even degree if the degree of all nodes $v\in V$ and $w\in W$ is even. We write $\cG^{\mathrm{even}}_{\leq D}\subset \cG_{\leq D}$ for the subcollection of $\cG_{\leq D}$ made of even-degree graphs. The following lemma states that polynomials $P_G$ associated to $G\in\cG_{\leq D}^{\mathrm{even}}$ are orthogonal to those with $G\in \cG_{\leq D}\setminus \cG_{\leq D}^{\mathrm{even}}$.

\begin{lemma}\label{lem:moment:ordre2:odd:degree}
Consider a graph $G^{(1)}=(V^{(1)},W^{(1)},E^{(1)})\in \cG_{\leq D}^{\mathrm{even}}$ and a graph $G^{(2)}=(V^{(2)},W^{(2)},E^{(2)})\in  \cG_{\leq D}\setminus \cG_{\leq D}^{\mathrm{even}}$. For any labeling $\pi^{(1)}= (\pi^{(1)}_1,\pi^{(1)}_2)$ and $\pi^{(2)}= (\pi^{(2)}_1,\pi^{(2)}_2)$ of those graphs, we have 
\[
\mathbb{E}\left[P_{G^{(1)},\pi^{(1)}}P_{G^{(2)},\pi^{(2)}}\right]=0\enspace . 
\]
\end{lemma}
Again, the property follows from the symmetry of the Rademacher random variables $(a_j)$ and $(z_i)$. As a consequence of the two previous lemmas, the low-degree correlation is achieved by the span of the $P_{G}$'s where $G$ is an even-degree graph.

Next, we shall transform the polynomials $P_{G,\pi_1,\pi_2}$ to ensure that they are almost-orthogonal. Since the distribution of $Y$ is invariant under permutations of $[n]$ and under permutations of $[d]$, we have that $\mathbb{E}[P_{G,\pi_1,\pi_2}]$ does not depend on the choice of $\pi_1$ and $\pi_2$.
Given any $G\in \mathcal{G}_{\leq D}$,  we write  $\cc(G)$ for the number of non-trivial connected components{ of $G$}\footnote{A connected component is said to be trivial if it is only made of a single isolated node.}.
Then, we partition $E$ into its $\cc(G)$ non-trivial connected components $E_1$, $E_2$,\ldots, $E_{\cc(G)}$, and we define accordingly the graphs $G_r=(V_r,W_r,E_r)$ for $r=1,\ldots, \cc(G)$, where $V_r\subset V$ and $W_r\subset W$ are the nodes induced by $E_r$ (so that $G_r$ has no isolated node). Then, we define $\Psi_{G,\pi_1,\pi_2}$ by centering by connected components.
\begin{equation}\label{eq:definition:PsiGpi}
\Psi_{G,\pi_1,\pi_2} = \prod_{r=1}^{\cc(G)} \left[P_{G_r,\pi_1,\pi_2} - \mathbb{E}\left(P_{G_r,\pi_1,\pi_2}\right)\right] \enspace .
\end{equation}
Centering the multivariate polynomials by connected components has already been done for the analysis of low-degree polynomials in rank-one matrix estimation~\cite{montanari2024equivalence}, in stochastic block models~\cite{CGGV25,carpentier2025phase}, and in clustering~\cite{carpentier2026low}.
This centering drastically reduces the cross-moments $\mathbb E[\Psi_{G^{(1)},\pi^{(1)}_1,\pi^{(1)}_2}\Psi_{G^{(2)},\pi^{(2)}_1,\pi^{(2)}_2}]$ when no non-isolated node of $G^{(1)}$ shares its label with a non-isolated node of $G^{(2)}$ --~where the label of a node $v\in V^{(i)}$ (resp.\ $w\in W^{(i)}$) is $\pi^{(i)}_1(v)$ (resp.\ $\pi^{(i)}_2(w)$)~--, and more generally when the labels of some connected component of one graph are disjoint from those of the other graph (see Proposition~\ref{lem:E_pi1pi2} and the remark below it).

Finally, we define the polynomials $\Psi_{G}$ and their normalized counterparts $\tilde{\Psi}_{G}$ by summing over all labelings $\pi_1$ and $\pi_2$
\begin{equation}\label{eq:definition:PsiG}
\Psi_{G} = \sum_{\pi_1 \in \overline{\Pi}_V} \sum_{\pi_2 \in \Pi_W}\Psi_{G,\pi_1,\pi_2} \ ; \quad \quad \quad \tilde{\Psi}_{G} = \frac{n^{-(|V|-2)/2}d^{-|W|/2}}{\sqrt{|\mathrm{Aut}(G)|}} \Psi_{G} \enspace,
\end{equation}
where $\mathrm{Aut}(G)$ is the set of all automorphisms of $G$ that preserve $v_1$ and $v_2$. Then, we deduce from Lemmas~\ref{lem:first_moment_bch} and~\ref{lem:moment:ordre2:odd:degree}, together with the parametrization of the $\Psi_G$'s, the following result. 

\begin{lemma}\label{lem:reduction:psi_even_bis}
We have 
\[
\mathrm{Corr}_{\leq D}= \sup_{P\in \mathrm{span}\big(1, (\Psi_{G})_{G\in \mathcal{G}^{\mathrm{even}}_{\leq D}}\big):\ \mathbb{E}[P^2]\leq 1} \mathbb{E}[xP]\enspace .
\]
\end{lemma}

\subsection{Almost-orthonormality of the $\tilde{\Psi}_G$'s.}

The main step of the proof is to establish that the polynomials $\tilde \Psi_G$ are almost-orthonormal. This is summarized in the following result. 
\begin{theorem}\label{thm:isorefo2}
Assume that Condition~\eqref{eq:signalcond} holds with a constant $c_{\mathrm{s}} \geq 10$, that $2^8D^4\leq d$, $D\geq 2$, and that $n\geq d$.
Then, for any $(\alpha_{\emptyset}, (\alpha_G)_{G \in \mathcal G^{\mathrm{even}}_{\leq D}})$, we have 
    $$(1-4D^{-c_{\mathrm{s}}/2}-  \frac{4D^2}{d} )\left[\alpha^2_{\emptyset}+ \sum_{G \in \mathcal G^{\mathrm{even}}_{\leq D}}\alpha^2_G\right] \leq \mathbb E \left[\left(\alpha_{\emptyset}+ \sum_{G \in \mathcal G^{\mathrm{even}}_{\leq D}} \alpha_G \tilde \Psi_G\right)^2\right] \leq (1+4D^{-c_{\mathrm{s}}/2}+  \frac{4D^2}{d} )\left[\alpha^2_{\emptyset}+ \sum_{G \in \mathcal G^{\mathrm{even}}_{\leq D}}\alpha^2_G\right]\enspace .$$
\end{theorem}
This theorem is proved in Section~\ref{sec:proof:thm:isorefo2}. The key ingredient for this proof is a tight control of the cross-moments $\mathbb{E}[\Psi_G\Psi_{G'}]$. This is done in Section~\ref{sec:moments} and this is where specific properties of our design come into play.

\medskip 

\noindent 
{\bf Remark}: We believe that Condition~\eqref{eq:signalcond} cannot be significantly weakened--say by some power of $d$--. Indeed, our bounds of the covariance between the $\tilde{\Psi}_G$ are order-wise tight --see the comment below Lemma~\ref{lem:eqcounting}.

\subsection{Proof of Theorem~\ref{thm:corr}}
    
Since $\kg(1)$ and $\kg(2)$ are i.i.d.~$\mathrm{Unif}([K])$, we have $\mathbb E[x]=\mathbb P[\kg(1)=\kg(2)]=1/K$. If $D=1$, then $\mathcal G^{\mathrm{even}}_{\leq D}$ is empty, since the two nodes of a graph with a single edge have degree $1$. By Lemma~\ref{lem:reduction:psi_even_bis}, the supremum defining $\mathrm{Corr}_{\leq D}$ then runs over constant polynomials $P=c$ with $c^2\leq 1$, so that $\mathrm{Corr}_{\leq D}=\mathbb E[x]=1/K$ and the bound of Theorem~\ref{thm:corr} trivially holds. Henceforth, we assume that $D\geq 2$, so that Theorem~\ref{thm:isorefo2} and Lemma~\ref{lem:scalprod} below apply.

Combining Theorem~\ref{thm:isorefo2} with Lemma~\ref{lem:reduction:psi_even_bis}, we readily get that the low-degree correlation satisfies
 \begin{equation}\label{eq:corrreint}
\mathrm{Corr}^2_{\leq D}\leq   \left[1-4D^{-c_{\mathrm{s}}/2}-  \frac{4D^2}{d} \right]^{-1}\left[\big(\mathbb E[x]\big)^2+\sum_{G \in \mathcal G^{\mathrm{even}}_{\leq D}} \big(\mathbb E[x \tilde \Psi_G]\big)^2\right]\enspace .      
 \end{equation}
Recall that $(\mathbb E[x])^2=1/K^2$. It only remains to bound  $\mathbb E[x \tilde \Psi_G]$. Recall that $\rho$ is defined in \eqref{eq:signalcond}.
\begin{lemma}\label{lem:scalprod}
Assume that Condition~\eqref{eq:signalcond} holds with $c_{\mathrm{s}} \geq 10$, that $2^8D^4\leq d$, $D\geq 2$, and that $n\geq d$. Consider a graph $G\in \cG_{\leq D}^{\mathrm{even}}$. If the degree of one node in $V\setminus \{v_1,v_2\}$  is smaller than $2m+2$, we have $\mathbb E[x \tilde \Psi_G]=0$. Otherwise, we have 
$$\left|\mathbb E[x \tilde \Psi_G]\right| \leq \frac{16}{K}\rho^{|E|}\enspace . $$
\end{lemma}

The proof of this lemma is postponed to Section~\ref{sec:proof:lem:scalprod}.
This implies that
$$\sum_{G \in \mathcal G^{\mathrm{even}}_{\leq D}} \left(\mathbb E[x \tilde \Psi_G]\right)^2 \leq  \frac{2^{8}}{K^2} \sum_{G \in \mathcal G^{\mathrm{even}}_{\leq D}: \mathrm{deg}(v)\geq 2m+2, \forall v\in V\setminus\{v_1,v_2\}} \rho^{2|E|}.$$
The number of graphs $G \in \mathcal G^{\mathrm{even}}_{\leq D}$ that have $e$ edges is smaller than $(D+2)^{2e}$ since such graphs contain at most $D+2$ nodes in $V$ and at most $D$ nodes in $W$ and since there are no isolated  nodes with the possible exception of $v_1$ and $v_2$. Since $D\geq 2$ and since $\rho\leq D^{-16c_{\mathrm{s}}}$ (Condition~\eqref{eq:signalcond}), we have $\rho^2(D+2)^2\leq D^{4-32c_{\mathrm{s}}}$, using that $(D+2)^2\leq D^4$ for $D\geq 2$. Since graphs $G \in \mathcal G^{\mathrm{even}}_{\leq D}$ have an even number of edges, it follows that
\begin{align*}
    \sum_{G \in \mathcal G^{\mathrm{even}}_{\leq D}} \left(\mathbb E[x \tilde \Psi_G]\right)^2 &\leq \frac{2^8}{K^2} \sum_{e\geq 2:\, e\ , \mathrm{ even}} \rho^{2e} (D+2)^{2e}
    = \frac{2^8}{K^2}\sum_{e'\geq1}\big[\rho^2(D+2)^2\big]^{2e'}
    \leq \frac{2^8}{K^2} \sum_{e'\geq 1} D^{(8-64c_{\mathrm{s}})e'}\\
    &\leq \frac{2^9}{K^2}D^{8-64c_{\mathrm{s}}}\leq \frac{4}{K^2}D^{-c_{\mathrm{s}}/2}\enspace ,
\end{align*}
where the second-to-last inequality bounds the geometric series (of ratio $D^{8-64c_{\mathrm{s}}}\leq D^{-632}$, since $c_{\mathrm{s}}\geq 10$) by twice its first term, and the last inequality uses $128D^{8-63.5c_{\mathrm{s}}}\leq 1$, which holds for $D\geq2$ and $c_{\mathrm{s}}\geq 10$.
Together with Equation~\eqref{eq:corrreint} and writing $a:= 4D^{-c_{\mathrm{s}}/2}+\frac{4D^2}{d}$, this implies that
$$\mathrm{Corr}^2_{\leq D}\leq\frac{1}{K^2}\cdot \frac{1+4D^{-c_{\mathrm{s}}/2}}{1-a} \leq \frac1{K^2}\cdot\frac{1+a}{1-a}\leq \frac{1}{K^2}\left(1+ \frac{16}{D^{c_{\mathrm{s}}/2}} + \frac{16D^2}{d}\right)\enspace ,$$
where we used $4D^{-c_{\mathrm{s}}/2}\leq a$ for the second inequality, and $\frac{1+a}{1-a}\leq 1+4a$ for the last one, which holds since $a\leq1/2$ (indeed, $D\geq2$, $c_{\mathrm{s}}\geq10$ and $2^8D^4\leq d$ give $a\leq 4\cdot2^{-5}+\frac{4D^2}{2^8D^4}\leq \frac18+\frac{1}{256}<1/2$).
This concludes the proof of the theorem.

\section{Moments of polynomials $P_G$ and $\Psi_G$}\label{sec:moments}

This section is where we build upon our construction of the design $\{\mu_k,\ k\leq K\}$.  Before controlling the cross-moments between the $\Psi_G$'s, we first evaluate the first moments of the canonical polynomials $P_{G,\pi_1,\pi_2}$.

\subsection{First moment}

Consider any $G\in \cG_{\leq D}^{\mathrm{even}}$. Conditional on $k^*$, $\sigma$, $z_i$, and $a_j$, the expectation of $Y_{ij}$ is equal to $\frac{\Delta}{\sqrt{d}} z_i a_j \mu_{k^*(i),\sigma(j)}$. Since the degree of all nodes is even, we deduce that  
\begin{align*}
\mathbb E[P_{G,\pi_1,\pi_2}(Y)]
&= \mathbb E_{\kg,\sigma}\Bigg[\prod_{e\in E} \frac{\Delta}{\sqrt{d}}  \mu_{\kg(\pi_1(l(e))),\sigma(\pi_2(r(e)))}\Bigg]\\
&= \Big(\frac{\Delta}{\sqrt{d}} \Big)^{|E|}\mathbb E_{\kg,\sigma}\Bigg[\prod_{v\in V} \prod_{w\in W: w\sim v}  \mu_{\kg(\pi_1(v)),\sigma(\pi_2(w))}\Bigg]\\
&= \Big(\frac{\Delta}{\sqrt{d}} \Big)^{|E|}\mathbb E_{\sigma}\Bigg[\prod_{v\in V} \mathbb E_{\kg(\pi_1(v))}\Big[\prod_{w\in W: w\sim v}  \mu_{\kg(\pi_1(v)),\sigma(\pi_2(w))}\Big]\Bigg].
\end{align*}
The expectation $\mathbb E_{\kg(\pi_1(v))}\Big[\prod_{w\in W: w\sim v}  \mu_{\kg(\pi_1(v)),\sigma(\pi_2(w))}\Big]$ is equal to the evaluation of the order $\mathrm{deg}_G(v)$ tensor $\frac{1}{K}\sum_{k \leq K} \mu_k^{\otimes \mathrm{deg}_G(v)}$ at the coordinate $(\sigma(\pi_2(w)))$ for the neighbors $w$ of $v$ --since this tensor is symmetric, this does not depend on the ordering of the neighbors $w$ of $v$. In other words, we have
$$\mathbb E_{\kg(\pi_1(v))}\Big[\prod_{w\in W: w\sim v}  \mu_{\kg(\pi_1(v)),\sigma(\pi_2(w))}\Big] = \Big(\frac{1}{K}\sum_{k \leq K} \mu_k^{\otimes \mathrm{deg}_G(v)}\Big)_{(\sigma(\pi_2(w)))_{w\sim v}}\enspace .$$
Lemma~\ref{lem:BCHeq} states that this expectation takes values in $\{0,1\}$ and that it is equal to $1$ if and only if, for all $1\leq s \leq m$, we have 
\begin{equation}\label{eq:startsysune}
\sum_{w: w\sim_G v} \phi^{(2s-1)}(\sigma(\pi_2(w))) = 0\enspace .
\end{equation}
Introduce the system $(\mathcal{S}_{G})$ of equations
\begin{equation}\tag{$\mathcal{S}_{G}$}\label{eq:startsys}
    \sum_{w: w\sim_G v} \phi^{(2s-1)}(\sigma(\pi_2(w))) = 0\enspace , \quad \quad  \forall  1\leq s\leq m, \forall v\in V\ ,
\end{equation}
and define $\mathcal{A}_{G}$ as the event that the system of equations~\eqref{eq:startsys} is satisfied{ --~we also write $\mathcal{A}[G]$ for $\mathcal{A}_{G}$ when $G$ is given by a lengthy expression~--}\footnote{The event $\mathcal{A}_{G}$ actually depends on $\pi_2$, but we keep this dependency implicit as the probability of this event does not depend on $\pi_2$.}. Then, it follows that 
\begin{align}\label{eq:sysdeg1}
\mathbb E[P_{G,\pi_1,\pi_2}(Y)]
&= \Big(\frac{\Delta}{\sqrt{d}} \Big)^{|E|}\mathbb P_{\sigma}[\cA_{G}]\enspace .
\end{align}
Then, we control the probability $P_{\sigma}[\cA_{G}]$ that the random system of equations $(\mathcal{S}_{G})$ is satisfied. The proofs of the following lemmas are postponed to Section~\ref{sec:number_solution}.

\begin{lemma}\label{lem:number_solutions}
Consider a bipartite graph $G=(V,W,E)$ such that the degree of all nodes is even.   If there exists $V_0\subset V$ such that $|\oplus_{v\in V_0} \cN(v)|\in [1:2m+1]$, then $\mathbb{P}_{\sigma}[\cA_{G}]=0$.
\end{lemma}

\noindent{\it Proof sketch.} Assume that~\eqref{eq:startsys} holds and sum its equations over $v\in V_0$. Since the sums are taken modulo $2$, the coordinates $w$ that are adjacent to an even number of nodes of $V_0$ cancel out, so that $\sum_{w\in\oplus_{v\in V_0}\cN(v)}\phi^{(2s-1)}(\sigma(\pi_2(w)))=0$ for all $s\leq m$. Besides, $|\oplus_{v\in V_0}\cN(v)|$ has the same parity as $\sum_{v\in V_0}\mathrm{deg}(v)$, which is even, so that it lies in $[1:2m]$. As the $\sigma(\pi_2(w))$ are pairwise distinct, this contradicts Lemma~\ref{lem:fundBCH}, which states that such a vanishing sum of at most $2m$ terms is necessarily paired. The full proof is given in Section~\ref{sec:number_solution}.

In particular, the lemma implies that $\mathbb{E}_{\sigma}[\mathbf{1}\{\cA_{G}\}]=0$ as soon as there exists a node $v\in V$ whose degree belongs to $[1:2m+1]$. In light of this lemma, we introduce the following condition on the graph $G$.

\begin{Condition}[$\cW_G$]\label{condition_W_G}
A graph $G=(V,W,E)$ satisfies Condition $(\mathcal{W}_{G})$ if (i) the degree of each node $w\in W$ is even and (ii) for any $V_0\subset V$, we have 
\begin{equation}\label{eq:cond:graph}
|\oplus_{v\in V_0} \cN(v)|\notin [1:2m+1] \ .
\end{equation}

\end{Condition}

For any graph $G=(V,W,E)$, we define
\begin{align}\label{eq:definition:b(G)}
b(G):=  \frac{|W|}{2} - \frac{|E|}{4} + \frac{|V|}{2} (m+1/2) - \frac{\cc(G)}{2}\enspace .
\end{align}
The following lemma  bounds $\mathbb{P}_{\sigma}[\cA_{G}]$ by evaluating the cardinality of the variety defined by~\eqref{eq:startsys} --see Section~\ref{sec:number_solution}.
\begin{lemma}\label{lem:eqcounting}
Consider any graph $G=(V,W,E)$ that satisfies Condition $(\cW_G)$, that does not contain any isolated node, and such that $|E|\leq 2D$ (this covers in particular any graph $G_{\oplus}$ obtained as the symmetric difference of two graphs in $\mathcal G_{\leq D}$). Assume that $D\geq 2$ and that $d\geq 2^8D^4$. We have
\begin{align}\label{eq:bound:proba:solutation}
\mathbb{P}_{\sigma}[\cA_{G}]\leq 2 (2m)^{|E|/2}\left(\frac{1}{d}\right)^{\lceil b(G)\rceil}\enspace .
\end{align}
\end{lemma}

\noindent
{\bf Remark}. Consider a graph $G$ such that the degree of all nodes is even. If $G$ does not satisfy Condition~$(\cW_G)$, the bound~\eqref{eq:bound:proba:solutation}
still obviously holds by Lemma~\ref{lem:number_solutions}.

\medskip
\noindent
{\bf Remark (optimality of the exponent $\lceil b(G)\rceil$).} The exponent $\lceil b(G)\rceil$ in~\eqref{eq:bound:proba:solutation} cannot be improved in general. Given an integer $r\geq 1$, consider the graph $G^{\circ}=(V^{\circ},W^{\circ},E^{\circ})$ made of a cycle of $2r$ pairs of twin nodes, defined as follows. For $t\in\mathbb Z/2r\mathbb Z$, the twin nodes $v_t,v'_t\in V^{\circ}$ have the same neighborhood, made of $2m+2$ nodes of $W^{\circ}$: $2m$ nodes $w_{t,1},\ldots,w_{t,2m}$ of degree $2$, which are only adjacent to $v_t$ and $v'_t$, and two nodes $u_{t-1}$ and $u_t$ of degree $4$, where $u_t$ is adjacent to $v_t$, $v'_t$, $v_{t+1}$ and $v'_{t+1}$. Hence, the nodes of degree $4$ connect each pair of twin nodes to the rest of the cycle. The graph $G^{\circ}$ is connected, with $|V^{\circ}|=4r$, $|W^{\circ}|=2r(2m+1)$ and $|E^{\circ}|=8r(m+1)$, so that $b(G^{\circ})=2rm-1/2$ and $\lceil b(G^{\circ})\rceil=2rm$. Besides, $G^{\circ}$ satisfies Condition~$(\cW_{G})$: since twin nodes have the same neighborhood, $\oplus_{v\in V_0}\cN(v)$ only depends on the set $T$ of indices $t$ such that $V_0$ contains exactly one of the nodes $v_t,v'_t$, and its cardinality is equal to $2m+2$ if $|T|=1$ and is at least $2m|T|\geq 4m$ if $|T|\geq 2$. We show in Section~\ref{sec:optimality:b} that, as $d\to\infty$ with $m$ and $r$ fixed,
$$(1-o(1))\Big(\frac{1}{d^{m}}\Big)^{2r}\leq \mathbb{P}_{\sigma}[\cA_{G^{\circ}}]\leq 2(2m)^{4r(m+1)}\Big(\frac{1}{d^{m}}\Big)^{2r}\enspace ,$$
so that $\mathbb{P}_{\sigma}[\cA_{G^{\circ}}]$ is of the order of $(1/d)^{\lceil b(G^{\circ})\rceil}$. In other words, each pair of twin nodes contributes a factor $d^{-m}$, which corresponds to its $m$ equations in~\eqref{eq:startsys}, the equations associated with $v'_t$ being identical to those associated with $v_t$. As noticed in Section~\ref{sec:optimality:b}, the rank of the Jacobian matrix of the system~\eqref{eq:startsys} is then exactly equal to $\lceil b(G^{\circ})\rceil$.

\subsection{Second Moments}

In order to control the second moments of the form $\mathbb{E}[P_{G^{(1)},\pi^{(1)}}P_{G^{(2)},\pi^{(2)}}]$, we need to introduce further graph notation.

\noindent
\paragraph{Merged graph $G_\cup$, intersection graph $G_{\cap}$, and symmetric difference graph $G_{\oplus}$.} Given a graph $G$ and labelings $\pi_1$ and $\pi_2$, we write $(\pi_1,\pi_2)[G]$ for the corresponding labeled graph. 
For such labeled bipartite graphs $(\pi^{(1)}_1,\pi^{(1)}_2)[G^{(1)}]$ and $(\pi^{(2)}_1,\pi^{(2)}_2)[G^{(2)}]$, we define $G_{\cup}$ as the merged graph, with the convention that identical edges are merged into a single edge.\\
Similarly,  we define the intersection graph 
$G_{\cap}=(V_{\cap},W_{\cap},E_{\cap})$ and the symmetric difference graph $G_{\oplus}=(V_{\oplus},W_{\oplus},E_{\oplus})$ so that $E_{\oplus}=E_{\cup}\setminus E_{\cap}$. Here, $V_{\cap}$ and $W_{\cap}$ (resp. $V_{\oplus}$ and $W_{\oplus}$) are the sets of nodes induced by the edges $E_{\cap}$ (resp. $E_{\oplus}$)
so that $G_{\cap}$ (resp. $G_{\oplus}$) does not contain any isolated node.   
A connected component of $(\pi^{(1)}_1,\pi^{(1)}_2)[G^{(1)}]$ with at least one edge is said to be unmatched if none of its nodes is incident to an edge of $(\pi^{(2)}_1,\pi^{(2)}_2)[G^{(2)}]$, and symmetrically for the connected components of $(\pi^{(2)}_1,\pi^{(2)}_2)[G^{(2)}]$. In other words, the labels of an unmatched component are disjoint from those of the non-isolated nodes of the other labeled graph. We write $\cc_{\mathrm{u}}$ for the total number of unmatched connected components of the two labeled graphs. This notion should not be confused with the sets of unmatched nodes introduced in Section~\ref{sec:additional:graph:notations}, which also account for isolated nodes.

\medskip

Consider any two graphs $G^{(1)}$ and  $G^{(2)}$ in $\mathcal{G}^{\mathrm{even}}_{\leq D}$ and any labelings $\pi^{(1)}$ and $\pi^{(2)}$. Since $Y^2_{ij}=1$ for all $i,j$, we obtain as in  \eqref{eq:sysdeg1} that 
\begin{equation}\label{eq:cross:PP}
\mathbb E[P_{G^{(1)},\pi^{(1)}}P_{G^{(2)},\pi^{(2)}}]= \Big(\frac{\Delta}{\sqrt{d}} \Big)^{|E_{\oplus}|}\mathbb{P}_{\sigma}[\cA_{G_{\oplus}}]\enspace . 
\end{equation}
Then, we can control the probability $\mathbb{P}_{\sigma}[\cA_{G_{\oplus}}]$ thanks to Lemmas~\ref{lem:number_solutions} and \ref{lem:eqcounting}.
Let us emphasize that neither the above identity nor the derivation of~\eqref{eq:sysdeg1}  relies on the membership of $G^{(1)}$ and $G^{(2)}$ in $\mathcal{G}^{\mathrm{even}}_{\leq D}$: the derivation of~\eqref{eq:sysdeg1} only requires that all the degrees of the (simple) bipartite graph $G$ are even  and that the labelings $\pi_1$ and $\pi_2$ are injective. Likewise, the above identity only requires that all the degrees of $G^{(1)}$ and $G^{(2)}$ are even. Accordingly, we shall freely apply the above identity to any two labeled bipartite graphs with even degrees and injective labelings, even when they do not belong to $\mathcal{G}^{\mathrm{even}}_{\leq D}$ nor carry distinguished nodes $v_1,v_2$.

Thanks to the correction in the definition~\eqref{eq:definition:PsiGpi} of the polynomials $\Psi_{G,\pi}$, the cross-moment $\mathbb E[\Psi_{G^{(1)},\pi^{(1)}}\Psi_{G^{(2)},\pi^{(2)}}]$ can be much smaller than $\mathbb E[P_{G^{(1)},\pi^{(1)}}P_{G^{(2)},\pi^{(2)}}]$. The following proposition is one of the key steps of the proof.

\begin{proposition}\label{lem:E_pi1pi2}
Consider any two graphs $G^{(1)}=(V^{(1)},W^{(1)},E^{(1)})$ and $G^{(2)}=(V^{(2)},W^{(2)},E^{(2)})$, each with at most $D$ edges and such that every node has even degree, and any injective labelings $\pi^{(1)}=(\pi_1^{(1)},\pi_2^{(1)})$ and $\pi^{(2)}=(\pi_1^{(2)},\pi_2^{(2)})$, with $\pi_1^{(i)}:V^{(i)}\to[n]$ and $\pi_2^{(i)}:W^{(i)}\to[d]$ (not necessarily lying in $\overline{\Pi}_V\times \Pi_W$; in particular $G^{(1)},G^{(2)}$ need not belong to $\mathcal{G}^{\mathrm{even}}_{\leq D}$ nor carry distinguished nodes $v_1,v_2$). Assume that $D\geq2$, $2^8D^4\leq d$, $n\geq d$, and that Condition~\eqref{eq:signalcond} holds for some $c_{\mathrm{s}}\geq10$. We have the following
\begin{itemize}
    \item[(a)] If at least one node $v\in V^{(1)}$ (resp. $v\in V^{(2)}$) satisfies $\mathrm{deg}_{G^{(1)}}(v)\in[1:2m+1]$ and $\pi^{(1)}_1(v)\notin \pi^{(2)}_1(V^{(2)})$ (resp. $\mathrm{deg}_{G^{(2)}}(v)\in[1:2m+1]$ and $\pi^{(2)}_1(v)\notin \pi^{(1)}_1(V^{(1)})$),
    then
    \[
    \mathbb{E}\left[\Psi_{G^{(1)},\pi^{(1)}}\Psi_{G^{(2)},\pi^{(2)}}\right] =0 \enspace ; \quad \quad  \mathbb{E}\left[P_{G^{(1)},\pi^{(1)}}\Psi_{G^{(2)},\pi^{(2)}}\right] =0\enspace . 
    \]
    \item[(b)] In any case, we have
    \[
\left|\mathbb{E}\left[\Psi_{G^{(1)},\pi^{(1)}}\Psi_{G^{(2)},\pi^{(2)}}\right]\right|\leq 2 (32 D^3)^{\cc_{\mathrm{u}}}  \left(4m\frac{\Delta}{d^{1/4}}\right)^{|E_{\oplus}|}\left(\frac{1}{d}\right)^{(2m+1)|V_{\oplus}|/4 + |W_{\oplus}|/2 - \cc(G_{\oplus})/2 +\cc_{\mathrm{u}}/2}\enspace . 
\]
    \item[(c)] If $E_{\oplus}= \emptyset$, then 
    \[
    \left|\mathbb{E}\left[\Psi_{G^{(1)},\pi^{(1)}}\Psi_{G^{(2)},\pi^{(2)}}\right]-1 \right|\leq  \frac{8m D \Delta }{d^{1/4}}\enspace . 
    \]
    \item[(d)] If no non-isolated node of $\pi^{(1)}[G^{(1)}]$ shares its label with a non-isolated node of $\pi^{(2)}[G^{(2)}]$, that is, if $\cc_{\mathrm{u}}=\cc(G^{(1)})+\cc(G^{(2)})$, then 
    \[
\left|\mathbb{E}\left[P_{G^{(1)},\pi^{(1)}}\Psi_{G^{(2)},\pi^{(2)}}\right]\right|\leq  (32 D^3)^{\cc(G^{(2)})}  \left(4m\frac{\Delta}{d^{1/4}}\right)^{|E_{\oplus}|}\left(\frac{1}{d}\right)^{(2m+1)|V_{\oplus}|/4 + |W_{\oplus}|/2 - \cc(G^{(1)})/2}\enspace .     
    \]
\end{itemize}

\end{proposition}

\noindent
{\bf Remark:} {\bf Benefit of centered polynomials}. By~\eqref{eq:cross:PP} and Lemma~\ref{lem:eqcounting}, the non-centered cross-moment satisfies
$$\left|\mathbb E\left[P_{G^{(1)},\pi^{(1)}}P_{G^{(2)},\pi^{(2)}}\right]\right|\leq 2\left(\sqrt{2m}\,\frac{\Delta}{d^{1/4}}\right)^{|E_{\oplus}|}\left(\frac{1}{d}\right)^{(2m+1)|V_{\oplus}|/4 + |W_{\oplus}|/2 - \cc(G_{\oplus})/2}\enspace .$$
Up to the factor $(32D^3)^{\cc_{\mathrm{u}}}$ and to the constants, Bound~(b) improves upon this bound by the factor $(1/d)^{\cc_{\mathrm{u}}/2}$, that is by a factor $d^{-1/2}$ per unmatched connected component. This gain is due to the centering by connected components in~\eqref{eq:definition:PsiGpi}. Indeed, consider an unmatched connected component of, say, $\pi^{(1)}[G^{(1)}]$. Its rows and columns are distinct from all the other rows and columns involved in the cross-moment, so that, conditionally on $\sigma$, it is independent of the remaining factors of the product. Since it is centered, the cross-moment would vanish if the coordinates were not coupled through the random permutation $\sigma$. As $\sigma$ only induces a weak dependence between the images of disjoint sets of columns, each unmatched component only contributes a factor of order $d^{-1/2}$. In particular, when the two labeled graphs do not intersect, we have $\cc_{\mathrm{u}}=\cc(G^{(1)})+\cc(G^{(2)})$, and the gain is of order $d^{-(\cc(G^{(1)})+\cc(G^{(2)}))/2}$.

\subsubsection*{AI Disclosure}

In an early phase, the authors used the AI assistant Gemini (Google) for bibliographical research on BCH codes, spherical design  and on algebraic geometry (algebraic arguments of Lemma~\ref{lem:eqcounting}).
In the final phase, the authors used the AI assistant Claude (Anthropic, Opus and Sonnet models 5.0 and 5.5), through the Claude Code interface, to help write and edit the main sections of this manuscript (Sections~\ref{sec:intro} to~\ref{sec:discussion}), and to audit the proofs. This assistance included drafting and rephrasing paragraphs of the introduction and of the discussion, reorganizing the presentation of the model and of the BCH construction, and harmonizing the notation. All the text produced with this assistance was reviewed and rephrased by the authors, who take full responsibility for the content of the manuscript, including all its mathematical statements and proofs.

\subsubsection*{Acknowledgements}

The work of A. Carpentier is partially supported by the Deutsche Forschungsgemeinschaft (DFG)- Project-ID 318763901 - SFB1294 "Data Assimilation", Project A03, by the DFG on the Forschungsgruppe FOR5381 "Mathematical Statistics in the Information Age - Statistical Efficiency and Computational Tractability", Project TP 02 (Project-ID 460867398), by the Universite franco-allemande (UFA) through the college doctoral franco-allemand CDFA-02-25 "Statistisches Lernen für komplexe stochastische Prozesse", by the European Research council (ERC) through the ERC Consolidator Grant SOCE (Nr: 101229569). Funded by the European Union (ERC, project SOCE, 101229569). Views and opinions expressed are however those of the author(s) only and do not necessarily reflect those of the European Union or the European Research Council. Neither the European Union nor the granting authority can be held responsible for them. The work of A.~Carpentier and N.~Verzelen is also supported by the Universite franco-allemande (UFA) through the college doctoral franco-allemand CDFA-02-25 "Statistisches Lernen für komplexe stochastische Prozesse".

\appendix

\section{Proof of Proposition~\ref{lem:E_pi1pi2}}\label{sec:proof:lem:E_pi1pi2}

In this section, our main purpose is to control $\mathbb{E}\left[\Psi_{G^{(1)},\pi^{(1)}}\Psi_{G^{(2)},\pi^{(2)}}\right]$.
Recall the definition 
\[
\Psi_{G^{(1)},\pi^{(1)}}= \prod_{i\in [\cc^{(1)}]} \left(P_{G^{(1)}_i,\pi^{(1)}} - \mathbb{E}\left[P_{G^{(1)}_i,\pi^{(1)}}\right]\right)\enspace .
\]
First, we expand the products inside the definitions of $\Psi_{G^{(1)},\pi^{(1)}}$ and $\Psi_{G^{(2)},\pi^{(2)}}$. Then, we integrate with respect to $k^*$ and $Y$.
Given $T_1\subset [\cc^{(1)}]$, and $T_2\subset [\cc^{(2)}]$, we define $G_{\oplus}^{T_1;T_2}$ as the graph associated to the symmetric difference of the labeled graphs
$(G^{(1)}, \pi^{(1)})$ and $(G^{(2)}, \pi^{(2)})$ from which we have removed the connected components in $[\cc^{(1)}]\setminus T_1$ and $[\cc^{(2)}]\setminus T_2$. Then, we have 
\begin{align*}
\mathbb{E}\left[\Psi_{G^{(1)},\pi^{(1)}}\Psi_{G^{(2)},\pi^{(2)}}\right]&= \sum_{T_1\subset [\cc^{(1)}]} \sum_{T_2\subset [\cc^{(2)}]} (-1)^{\cc^{(1)}+\cc^{(2)}-|T_1|-|T_2|} \left(\frac{\Delta}{\sqrt{d}}\right)^{|E_{\oplus}^{(T_1,T_2)}|+\sum_{i\notin T_1}|E_i^{(1)}|+ \sum_{i\notin T_2}|E_i^{(2)}| }\\
& \quad \cdot \mathbb{P}_{\sigma}\left[\mathcal{A}[G^{T_1;T_2}_{\oplus}]\right] \prod_{i\in [\cc^{(1)}]\setminus T_1}   \mathbb{P}_{\sigma}\left[\mathcal{A}[G^{(1)}_i]\right]\prod_{i\in [\cc^{(2)}]\setminus T_2}\mathbb{P}_{\sigma}\left[\mathcal{A}[G^{(2)}_i]\right] 
\end{align*}

Denote by $R^{(1)}_0\subset [\cc^{(1)}]$ (resp. $R^{(2)}_0\subset [\cc^{(2)}]$) the set of connected components of $G^{(1)}$ (resp. $G^{(2)}$) such that at least one of its nodes is matched to a node of $G^{(2)}$ (resp. $G^{(1)}$). 
Then, we decompose  $\mathbb{E}\left[\Psi_{G^{(1)},\pi^{(1)}}\Psi_{G^{(2)},\pi^{(2)}}\right]$ as follows. 
\begin{align}\label{eq:formula_psi_2}
\mathbb{E}\left[\Psi_{G^{(1)},\pi^{(1)}}\Psi_{G^{(2)},\pi^{(2)}}\right]&= \sum_{R^{(1)}\subset R^{(1)}_0}
\sum_{R^{(2)}\subset R^{(2)}_0} \left(\frac{\Delta}{\sqrt{d}}\right)^{u[R^{(1)}; R^{(2)}]}  \omega[R^{(1)};R^{(2)}]\enspace ; \\
u[R^{(1)};R^{(2)}]& = |E_{\oplus}^{R^{(1)}; R^{(2)}}|+ \sum_{i\in [\cc^{(1)}]\setminus R^{(1)}} |E^{(1)}_i| +  \sum_{i\in [\cc^{(2)}]\setminus R^{(2)}} |E^{(2)}_i|\enspace ; \label{eq:formula_u} \\ \nonumber 
\omega[R^{(1)};R^{(2)}]&:= \sum_{\substack{T_1\subset [\cc^{(1)}]: T_1\cap R^{(1)}_0= R^{(1)}\\ T_2\subset [\cc^{(2)}]: T_2\cap R^{(2)}_0= R^{(2)}} }(-1)^{\cc^{(1)}+\cc^{(2)}-|T_1|-|T_2|}  \mathbb{P}_{\sigma}\left[\mathcal{A}[G^{T_1;T_2}_{\oplus}]\right] \prod_{i\in [\cc^{(1)}]\setminus T_1}   \mathbb{P}_{\sigma}\left[\mathcal{A}[G^{(1)}_i]\right]\\ &\quad\quad \quad\quad\quad \cdot \prod_{i\in [\cc^{(2)}]\setminus T_2}\mathbb{P}_{\sigma}\left[\mathcal{A}[G^{(2)}_i]\right] \enspace . 
\end{align}

The main challenge is to control $|\omega[R^{(1)};R^{(2)}]|$. 

\bigskip 

First,  we prove Case (a) in Proposition~\ref{lem:E_pi1pi2}. By symmetry, we may assume that $v\in V^{(1)}$ satisfies $\mathrm{deg}_{G^{(1)}}(v)\in[1:2m+1]$ and $\pi_1^{(1)}(v)\notin \pi_1^{(2)}(V^{(2)})$. Since $v$ is non-isolated, $\pi_1^{(1)}(v)$ belongs to the (edge-induced) node set of at least one of $G_{\oplus}^{T_1;T_2}$, $G_i^{(1)}$, $G_i^{(2)}$ for any choice of $T_1$ and $T_2$, and its degree in that graph is at most $2m+1$. Then, Lemma~\ref{lem:number_solutions} implies that $\omega[R^{(1)};R^{(2)}]=0$. We conclude that 
$\mathbb{E}\left[\Psi_{G^{(1)},\pi^{(1)}}\Psi_{G^{(2)},\pi^{(2)}}\right]= 0$. Arguing similarly, we obtain $\mathbb{E}\left[P_{G^{(1)},\pi^{(1)}}\Psi_{G^{(2)},\pi^{(2)}}\right]= 0$.

\bigskip

Let us turn to Case (b), which is the most challenging part of the proof. We have 
\begin{align}\label{eq:upper:omega}
|\omega[R^{(1)};R^{(2)}]|\leq \left|\mathbb{E}_{\sigma}\left[L[R^{(1)};R^{(2)}]\right]\right| \prod_{i\in R_0^{(1)}\setminus R^{(1)}}\mathbb{P}_{\sigma}\left[\mathcal{A}[G^{(1)}_i]\right]\prod_{i\in R_0^{(2)}\setminus R^{(2)}}\mathbb{P}_{\sigma}\left[\mathcal{A}[G^{(2)}_i]\right]\ , 
\end{align}
where
\begin{align*}
L[R^{(1)};R^{(2)}]:= \mathbf{1}\left\{\mathcal{A}[G_{\oplus}^{R^{(1)}; R^{(2)}}]\right\}&\prod_{i\in [\cc^{(1)}]\setminus R_0^{(1)}}\left(\mathbf{1}\left\{\mathcal{A}[G^{(1)}_i]\right\}- \mathbb{P}_{\sigma}\left[\mathcal{A}[G^{(1)}_i]\right]\right)\\
\cdot&\prod_{i\in [\cc^{(2)}]\setminus R_0^{(2)}}\left(\mathbf{1}\left\{\mathcal{A}[G^{(2)}_i]\right\}- \mathbb{P}_{\sigma}\left[\mathcal{A}[G^{(2)}_i]\right]\right)
\end{align*}
\begin{lemma}\label{lem:L:R1R2}
Under the assumptions of Proposition~\ref{lem:E_pi1pi2}, we have 
\begin{align*}
\left|\mathbb{E}_{\sigma}\left[L[R^{(1)};R^{(2)}]\right]\right|
\leq     (32 D^3)^{\cc_{\mathrm{u}}} (4m)^{\frac{|E_{\oplus}^{R^{(1)};R^{(2)}}|+\sum_{i\in [\cc^{(1)}]\setminus R_0^{(1)}} |E_i^{(1)}|+ \sum_{i\in [\cc^{(2)}]\setminus R_0^{(2)}}|E_i^{(2)}|}{2}}
\left(\frac{1}{d}\right)^{ b_{R^{(1)},R^{(2)}} } 
\enspace ,
\end{align*}    
where 
\begin{align*}
b_{R^{(1)},R^{(2)}} & =  -\frac{|E_{\oplus}^{R^{(1)};R^{(2)}}|+\sum_{i\in [\cc^{(1)}]\setminus R_0^{(1)}} |E_i^{(1)}|+ \sum_{i\in [\cc^{(2)}]\setminus R_0^{(2)}}|E_i^{(2)}|}{4} \\ & \quad + (2m+1)\frac{ |V_{\oplus}^{R^{(1)};R^{(2)}}|+ \sum_{i\in [\cc^{(1)}]\setminus R_0^{(1)}} |V_i^{(1)}|+ \sum_{i\in [\cc^{(2)}]\setminus R_0^{(2)}}|V_i^{(2)}|  }{4} \\ &\quad\quad - \frac{\cc(G^{R^{(1)};R^{(2)}}_{\oplus})}{2} + \frac{|W_{\oplus}^{R^{(1)};R^{(2)}}|+ \sum_{i\in [\cc^{(1)}]\setminus R_0^{(1)}} |W_i^{(1)}|+ \sum_{i\in [\cc^{(2)}]\setminus R_0^{(2)}} |W_i^{(2)}|}  {2}
\enspace .
\end{align*}
\end{lemma}

Coming back to~\eqref{eq:upper:omega}, we control the terms of the form $\mathbb{P}_{\sigma}[\mathcal{A}[G^{(1)}_i]]$ and $\mathbb{P}_{\sigma}[\mathcal{A}[G^{(2)}_i]]$ using Lemmas~\ref{lem:number_solutions} and \ref{lem:eqcounting}.
This leads us to 
\begin{align}\label{eq:upper:omega2}
|\omega[R^{(1)};R^{(2)}]|\leq (32 D^3)^{\cc_{\mathrm{u}}} (4m)^{u[R^{(1)};R^{(2)}]/2}\left(\frac{1}{d}\right)^{b'_{R^{(1)},R^{(2)}}}
\end{align}
where 
\begin{align*}
b'_{R^{(1)},R^{(2)}} & =  -\frac{|E_{\oplus}^{R^{(1)};R^{(2)}}|+\sum_{i\in [\cc^{(1)}]\setminus R^{(1)}} |E_i^{(1)}|+ \sum_{i\in [\cc^{(2)}]\setminus R^{(2)}}|E_i^{(2)}|}{4} \\ & \quad +(2m+1)\frac{ |V_{\oplus}^{R^{(1)};R^{(2)}}|+ \sum_{i\in [\cc^{(1)}]\setminus R^{(1)}} |V_i^{(1)}|+ \sum_{i\in [\cc^{(2)}]\setminus R^{(2)}}|V_i^{(2)}|  }{4} \\ &\quad - \frac{\cc(G^{R^{(1)};R^{(2)}}_{\oplus})+ |R_0^{(1)}\setminus R^{(1)}| + |R_0^{(2)}\setminus R^{(2)}| }{2} \\ &\quad + \frac{|W_{\oplus}^{R^{(1)};R^{(2)}}|+ \sum_{i\in [\cc^{(1)}]\setminus R^{(1)}} |W_i^{(1)}|+ \sum_{i\in [\cc^{(2)}]\setminus R^{(2)}} |W_i^{(2)}|}  {2}
\enspace .    
\end{align*}
Then, coming back to~\eqref{eq:formula_psi_2} and~\eqref{eq:formula_u}, we get 
\begin{align*}
\left|\mathbb{E}\left[\Psi_{G^{(1)},\pi^{(1)}}\Psi_{G^{(2)},\pi^{(2)}}\right]\right|&\leq (32 D^3)^{\cc_{\mathrm{u}}} \left(\frac{\Delta2\sqrt{m}}{\sqrt{d}}\right)^{u[R_0^{(1)}; R_0^{(2)}]}  \left(\frac{1}{d}\right)^{b'_{R_0^{(1)},R_0^{(2)}}} ~\\ & \quad \quad 
\sum_{R^{(1)}\subset R^{(1)}_0}
\sum_{R^{(2)}\subset R^{(2)}_0} 
\left(\frac{\Delta 2\sqrt{m}}{\sqrt{d}}\right)^{u[R^{(1)}; R^{(2)}]- u[R_0^{(1)}; R_0^{(2)}]}  \left(\frac{1}{d}\right)^{b'_{R^{(1)},R^{(2)}} - b'_{R_0^{(1)},R_0^{(2)}}}
\enspace ;
\end{align*}
When we move from $R^{(1)}$ to $R^{(1)}_0$, we reduce the number of connected components from $\cc(G^{R^{(1)};R^{(2)}}_{\oplus})+ |R_0^{(1)}\setminus R^{(1)}| + |R_0^{(2)}\setminus R^{(2)}|$ to $\cc(G^{R_0^{(1)};R_0^{(2)}}_{\oplus})$ by merging some nodes. Since merging two nodes allows one to reduce the number of connected components by at most one and since $2m+1\geq 2$, we deduce that
\[
b'_{R^{(1)},R^{(2)}} - b'_{R_0^{(1)},R_0^{(2)}} \geq \frac{u[R_0^{(1)}; R_0^{(2)}] - u[R^{(1)}; R^{(2)}]}{4}\ .
\]
Let us make this argument precise, as the passage from the $(R^{(1)},R^{(2)})$-configuration to the $(R_0^{(1)},R_0^{(2)})$-configuration not only merges nodes, but also deletes the edges that cancel out in the symmetric difference. Write $H$ for the disjoint union of $G^{R^{(1)};R^{(2)}}_{\oplus}$, of the labeled components $\pi^{(1)}[G^{(1)}_i]$ with $i\in R_0^{(1)}\setminus R^{(1)}$ and of the labeled components $\pi^{(2)}[G^{(2)}_i]$ with $i\in R_0^{(2)}\setminus R^{(2)}$. Since the components indexed by $[\cc^{(1)}]\setminus R_0^{(1)}$ and $[\cc^{(2)}]\setminus R_0^{(2)}$ contribute in the same way to $b'_{R^{(1)},R^{(2)}}$ and to $b'_{R_0^{(1)},R_0^{(2)}}$, and since each excluded component $\pi^{(j)}[G_i^{(j)}]$, $i\in R_0^{(j)}\setminus R^{(j)}$, is connected --so that the quantity $\cc(G^{R^{(1)};R^{(2)}}_{\oplus})+ |R_0^{(1)}\setminus R^{(1)}| + |R_0^{(2)}\setminus R^{(2)}|$ arising in $b'_{R^{(1)},R^{(2)}}$ is nothing but $\cc(H)$--, we have
\[
b'_{R^{(1)},R^{(2)}} - b'_{R_0^{(1)},R_0^{(2)}}= -\frac{\delta_E}{4} + (2m+1)\frac{\delta_V}{4} + \frac{\delta_W}{2} - \frac{\delta_{\cc}}{2}\enspace ,
\]
where $\delta_E$, $\delta_V$, $\delta_W$, and $\delta_{\cc}$ respectively denote the numbers of edges, of nodes in $V$, of nodes in $W$, and of connected components of $H$ minus the corresponding quantities for $G^{R_0^{(1)};R_0^{(2)}}_{\oplus}$. As $\delta_E= u[R^{(1)}; R^{(2)}]-u[R_0^{(1)}; R_0^{(2)}]$, the announced inequality is equivalent to
\begin{equation}\label{eq:merge:target}
(2m+1)\delta_V + 2\delta_W\geq 2\delta_{\cc}\enspace .
\end{equation}
Now, $G^{R_0^{(1)};R_0^{(2)}}_{\oplus}$ is obtained from $H$ in two steps: (i) we identify the nodes of $H$ that carry the same label; (ii) we delete the pairs of edges that have become parallel in this identification --namely the labeled edges belonging to both $\pi^{(1)}[G^{(1)}]$ and $\pi^{(2)}[G^{(2)}]$, which cancel out in the symmetric difference-- as well as the nodes that have become isolated. Write $p_V$ and $p_W$ for the numbers of identifications performed at step (i) among the nodes of $V$ and of $W$, and $r_V$ and $r_W$ for the numbers of nodes of $V$ and of $W$ that are deleted at step (ii), so that $\delta_V= p_V+r_V$ and $\delta_W= p_W+r_W$. Step (i) reduces the number of connected components by at most $p_V+p_W$, since each identification of two nodes merges at most two connected components. As for step (ii), each connected component of the graph obtained at step (i) that keeps at least one edge gives rise to at least one connected component of $G^{R_0^{(1)};R_0^{(2)}}_{\oplus}$, so that step (ii) reduces the number of connected components by at most the number $k$ of connected components of the graph obtained at step (i) that lose all their edges. Any of these $k$ components contains at least one edge --the connected components of the graph obtained at step (i) being unions of connected components of $H$, which all contain at least one edge--, hence at least one node of $V$ and one node of $W$, all of which become isolated and are therefore deleted; consequently, $r_V\geq k$ and $r_W\geq k$. Gathering these bounds and using $2m+1\geq 2$, we get
\[
(2m+1)\delta_V + 2\delta_W \geq 2(p_V+r_V) + 2(p_W+r_W) \geq 2(p_V+p_W+k)\geq 2\delta_{\cc}\enspace ,
\]
which is~\eqref{eq:merge:target}.
We have that $u[R_0^{(1)}; R_0^{(2)}]=  |E_{\oplus}|$ and $b'_{R_0^{(1)},R_0^{(2)}}= -u[R_0^{(1)}; R_0^{(2)}]/4 + (2m+1)|V_{\oplus}|/4 + |W_{\oplus}|/2 - \cc(G_{\oplus})/2 +\cc_{\mathrm{u}}/2$.
We conclude that  
\begin{align*}
\left|\mathbb{E}\left[\Psi_{G^{(1)},\pi^{(1)}}\Psi_{G^{(2)},\pi^{(2)}}\right]\right|&\leq   (32 D^3)^{\cc_{\mathrm{u}}} \left( 2\sqrt{m}\frac{\Delta}{d^{1/4}}\right)^{|E_{\oplus}|}\left(\frac{1}{d}\right)^{(2m+1)|V_{\oplus}|/4 + |W_{\oplus}|/2 - \cc(G_{\oplus})/2 +\cc_{\mathrm{u}}/2}\\ &\quad \quad 
\sum_{R^{(1)}\subset R^{(1)}_0}
\sum_{R^{(2)}\subset R^{(2)}_0} \left( 2\sqrt{m} \frac{\Delta}{d^{1/4}}\right)^{u[R^{(1)}; R^{(2)}] - u[R_0^{(1)}; R_0^{(2)}]}
\end{align*}
Let $T_0^{(1)}\subset R_0^{(1)}$ (resp. $T_0^{(2)}\subset R_0^{(2)}$) be the indices of the connected components such that all their edges are pruned in the construction of $E_{\oplus}$. As a consequence, we have 
\[
u[R^{(1)}; R^{(2)}] - u[R_0^{(1)}; R_0^{(2)}]\geq |T_0^{(1)}\setminus R^{(1)} | + |T_0^{(2)}\setminus R^{(2)}|\enspace . 
\]
Since we assume that $8mD \Delta\leq d^{1/4}$ --see Hypothesis~\eqref{eq:signalcond}--, we get 
\begin{align*}
&\sum_{R^{(1)}\subset R^{(1)}_0}
\sum_{R^{(2)}\subset R^{(2)}_0}  \left( 2\sqrt{m}\frac{\Delta}{d^{1/4}}\right)^{u[R^{(1)}; R^{(2)}] - u[R_0^{(1)}; R_0^{(2)}]}  \\ 
&\quad \quad \leq \sum_{\substack{T^{(1)}\subset T^{(1)}_0\\ T^{(2)}\subset T^{(2)}_0
}}\sum_{\substack{R^{(1)}\subset R^{(1)}_0\setminus T^{(1)}_0 
\\
R^{(2)}\subset R^{(2)}_0\setminus T^{(2)}_0}} \left(2\sqrt{m}\frac{\Delta}{d^{1/4}}\right)^{|T^{(1)}|+ |T^{(2)}|} \\
&\quad \quad \leq 2^{|R^{(1)}_0\setminus T^{(1)}_0|+ |R^{(2)}_0\setminus T^{(2)}_0|}\sum_{t=0}^{\infty} (2D)^{t}\left(2\sqrt{m}\frac{\Delta}{d^{1/4}}\right)^{t}\\
&\quad \quad \leq 2 \cdot 2^{|E_{\oplus}|}\enspace .
\end{align*}
We conclude that 
\[
\left|\mathbb{E}\left[\Psi_{G^{(1)},\pi^{(1)}}\Psi_{G^{(2)},\pi^{(2)}}\right]\right|\leq 2 (32 D^3)^{\cc_{\mathrm{u}}}  \left(4m\frac{\Delta}{d^{1/4}}\right)^{|E_{\oplus}|}\left(\frac{1}{d}\right)^{(2m+1)|V_{\oplus}|/4 + |W_{\oplus}|/2 - \cc(G_{\oplus})/2 +\cc_{\mathrm{u}}/2}
\]
We have established Case (b) in Proposition~\ref{lem:E_pi1pi2}.

\bigskip 

It remains to deal with Case (c). We build on the previous arguments, up to minor modifications. Since $E_{\oplus}=\emptyset$, we have $u[R_0^{(1)}; R_0^{(2)}]=0$. Starting back from~\eqref{eq:formula_psi_2} and arguing as in Case (b),  we get
\begin{align*}
\left|\mathbb{E}\left[\Psi_{G^{(1)},\pi^{(1)}}\Psi_{G^{(2)},\pi^{(2)}}-1\right]\right|&= \Big|\sum_{\substack{R^{(1)}\subset R^{(1)}_0\\
R^{(2)}\subset R^{(2)}_0\\ \text{s.t.} R^{(1)}\neq R^{(1)}_0 \text{ or } R^{(2)}\neq R^{(2)}_0}} \left(\frac{\Delta}{\sqrt{d}}\right)^{u[R^{(1)}; R^{(2)}]}  \omega[R^{(1)};R^{(2)}]\Big|\\
&\leq  \Big|\sum_{\substack{R^{(1)}\subset R^{(1)}_0\\
R^{(2)}\subset R^{(2)}_0\\ \text{s.t.} R^{(1)}\neq R^{(1)}_0 \text{ or } R^{(2)}\neq R^{(2)}_0}} \left( 2m^{1/2}\frac{\Delta}{d^{1/4}}\right)^{u[R^{(1)}; R^{(2)}]}  \Big| \enspace . 
\end{align*}
Since $E_{\oplus}=\emptyset$, we have  
\[
u[R^{(1)}; R^{(2)}]\geq |R_0^{(1)}\setminus R^{(1)}|+ |R_0^{(2)}\setminus R^{(2)}|\ .
\]
Since we assume that $8 mD \Delta \leq d^{1/4}$, we conclude 
\[
\left|\mathbb{E}\left[\Psi_{G^{(1)},\pi^{(1)}}\Psi_{G^{(2)},\pi^{(2)}}-1\right]\right|
\leq \sum_{r=1}^{\infty} \left( \frac{ 4mD \Delta }{d^{1/4}}\right)^{r}\leq  \frac{8m D \Delta }{d^{1/4}}\enspace .
\]

\medskip 

It remains to establish the result (d). The proof is very close to that of (b) and we only sketch the differences. 
Define $G_{\oplus}^{T_2}$ as the graph associated to the symmetric difference of the labeled graphs $(G^{(1)}, \pi^{(1)})$ and $(G^{(2)}, \pi^{(2)})$ from which we have removed the connected components in $[\cc^{(2)}]\setminus T_2$. Observe that $G_{\oplus}^{T_2}$ is made of a collection of $\cc^{(1)}+|T_2|$ non-trivial connected components.
Similarly to~\eqref{eq:formula_psi_2}, we have 
\begin{align}\label{eq:formula_psi_2_bis}
\mathbb{E}\left[P_{G^{(1)},\pi^{(1)}}\Psi_{G^{(2)},\pi^{(2)}}\right]&=  \left(\frac{\Delta}{\sqrt{d}}\right)^{|E^{(1)}|+|E^{(2)}|}  \mathbb{E}_{\sigma}[L]\enspace ; \\
\nonumber 
L&:= \mathbf{1}\left\{\mathcal{A}[G^{(1)}]\right\} \prod_{i\in [\cc^{(2)}]}\left(\mathbf{1}\left\{\mathcal{A}[G^{(2)}_i]\right\}- \mathbb{P}_{\sigma}\left[\mathcal{A}[G^{(2)}_i]\right]\right)  \enspace .
\end{align}
Since $\mathcal{A}[G^{(1)}]$ and $\mathcal{A}[G^{(2)}_i]$ depend only on the respective edge sets, we may replace $G^{(1)}$ and each $G^{(2)}_i$ by their edge-induced subgraphs (removing any isolated node) before applying Lemma~\ref{lem:eqcounting}, exactly as for the pieces $G^*_0,\ldots,G^*_l$ in the proof of Lemma~\ref{lem:L:R1R2} below; this is consistent with $|V_{\oplus}|,|W_{\oplus}|$ below already denoting the node counts induced by $E^{(1)}\cup E^{(2)}$. Analogously to Lemma~\ref{lem:L:R1R2}, we get
\begin{align*}
\left|\mathbb{E}_{\sigma}\left[L\right]\right|
\leq     (32 D^3)^{\cc(G^{(2)})} (4m)^{\frac{|E_\oplus|}{2}}
\left(\frac{1}{d}\right)^{ -|E_{\oplus}|/4+(2m+1)|V_{\oplus}|/4+ |W_{\oplus}|/2- \cc(G^{(1)})/2 } 
\enspace .
\end{align*}    
Together with~\eqref{eq:formula_psi_2_bis}, we arrive at the desired result.

\begin{proof}[Proof of Lemma~\ref{lem:L:R1R2}]
Note that the labeled graphs $G_{\oplus}^{R^{(1)};R^{(2)}}$, $\pi^{(1)}[G^{(1)}_i]$ with $i\in [\cc^{(1)}]\setminus R_0^{(1)}$ and $\pi^{(2)}[G^{(2)}_i]$ with $i\in [\cc^{(2)}]\setminus R_0^{(2)}$ do not have any node in common. We define  $(G^*,\pi^*_1,\pi^*_2)$ as the labeled graph that contains $G_{\oplus}^{R^{(1)};R^{(2)}}$ as well as the graphs $\pi^{(1)}[G^{(1)}_i]$ with $i\in [\cc^{(1)}]\setminus R_0^{(1)}$ and $\pi^{(2)}[G^{(2)}_i]$ with $i\in [\cc^{(2)}]\setminus R_0^{(2)}$. Write $l= |[\cc^{(1)}]\setminus R_0^{(1)}| + |[\cc^{(2)}]\setminus R_0^{(2)}|$. For simplicity, we write $G^*_i$, $i=1,\ldots, l$ for those $l$ connected components. We also write $G^*_0$ for the subgraph corresponding to $G_{\oplus}^{R^{(1)};R^{(2)}}$. Note that $G^*_0$ is possibly empty especially if $|R^{(1)}|+|R^{(2)}|=0$. With this notation, we have

\begin{equation}\label{eq:definition:L:R1:R2}
L[R^{(1)};R^{(2)}] = \mathbf{1}\left\{\mathcal{A}[G^*_0]\right\}\prod_{i=1}^l \left[\mathbf{1}\left\{\mathcal{A}[G^*_i]\right\}- \mathbb{E}_{\sigma}\left[\mathbf{1}\left\{\mathcal{A}[G^*_i]\right\}\right]\right]
\end{equation}

\noindent
\emph{The case $l=0$.} Assume first that $l=0$, that is $R_0^{(1)}=[\cc^{(1)}]$ and $R_0^{(2)}=[\cc^{(2)}]$ (equivalently $\cc_{\mathrm{u}}=0$). Then the product in~\eqref{eq:definition:L:R1:R2} is empty, so that $L[R^{(1)};R^{(2)}]= \mathbf{1}\{\mathcal{A}[G^*_0]\}$ and
\[
\left|\mathbb{E}_{\sigma}\left[L[R^{(1)};R^{(2)}]\right]\right|= \mathbb{P}_{\sigma}\left[\mathcal{A}[G^*_0]\right]\enspace ,
\]
with $G^*=G^*_0= G^{R^{(1)};R^{(2)}}_{\oplus}$. Writing $|E^*_0|$, $|V^*_0|$, $|W^*_0|$ for the numbers of edges and of (non-isolated) nodes of $G^*_0$, the exponent in the statement of the lemma reduces, when $l=0$, to
\[
b_{R^{(1)},R^{(2)}}= -\frac{|E^*_0|}{4}+ (2m+1)\frac{|V^*_0|}{4} - \frac{\cc(G^*_0)}{2}+ \frac{|W^*_0|}{2}= b(G^*_0)\enspace ,
\]
by the definition~\eqref{eq:definition:b(G)} of $b(\cdot)$, while the prefactor $(32D^3)^{\cc_{\mathrm{u}}}$ equals one. Hence, it suffices to prove that $\mathbb{P}_{\sigma}[\mathcal{A}[G^*_0]]\leq (4m)^{|E^*_0|/2}(1/d)^{b(G^*_0)}$.
If $G^*_0$ is empty, the system~\eqref{eq:startsys} is empty, so that $\mathbb{P}_{\sigma}[\mathcal{A}[G^*_0]]=1$, whereas the above right-hand side is also equal to one since $|E^*_0|=|V^*_0|=|W^*_0|=\cc(G^*_0)=0$. If $G^*_0$ is non-empty, then it contains no isolated node (being induced by its edge set), all its degrees are even (a symmetric difference of graphs with even degrees has even degrees), and $|E^*_0|\leq |E^{(1)}|+|E^{(2)}|\leq 2D$. If $G^*_0$ satisfies Condition~$(\cW_G)$, Lemma~\ref{lem:eqcounting} gives
\[
\mathbb{P}_{\sigma}\left[\mathcal{A}[G^*_0]\right]\leq 2 (2m)^{|E^*_0|/2}\left(\frac{1}{d}\right)^{b(G^*_0)}\leq (4m)^{|E^*_0|/2}\left(\frac{1}{d}\right)^{b(G^*_0)}\enspace ,
\]
where we used $2\leq 2^{|E^*_0|/2}$, since $|E^*_0|\geq 4$ (any non-empty connected component of a graph with even degrees contains at least $4$ edges, see the end of this proof). If $G^*_0$ does not satisfy Condition~$(\cW_G)$, then $\mathbb{P}_{\sigma}[\mathcal{A}[G^*_0]]=0$ by Lemma~\ref{lem:number_solutions} --see the Remark following Lemma~\ref{lem:eqcounting}-- and the bound is trivial. This settles the case $l=0$.

\medskip
\noindent
\emph{In the remainder of this proof, we therefore assume that $l\geq 1$.}

\medskip

Write $W^*_0,W^*_1,\ldots, W^*_l$ for the sets of nodes corresponding to $G^*_0,\ldots, G^*_l$. Note that these sets are disjoint. 
Define $\tilde{\mathbb{P}}_{\sigma}$ as the distribution such that for $i=0,\ldots, l$, $\sigma(\pi_2(w))$ with $w\in W^*_i$ are sampled uniformly without replacement on $[d]$, \emph{but} the $\sigma(\pi_2(W^*_i))'s$ are sampled independently. We write $\tilde{\mathbb{E}}_{\sigma}$ for the corresponding expectation. 
This distribution is chosen in such a way that $\mathbb{P}_{\sigma}\left[\mathcal{A}[G^*_i]\right]= \tilde{\mathbb{P}}_{\sigma}\left[\mathcal{A}[G^*_i]\right]$ and the events $\mathcal{A}[G^*_i]$'s are independent under $\tilde{P}$.

Define the event $\mathcal{E}$ on $\sigma$ such that,  on this event,  the sets $\sigma(\pi_2(W^*_i))$'s are disjoint for $i=0,\ldots, l$. We have 
\[
\mathbb{E}_{\sigma}\left[\mathbf{1}_{\mathcal{E}}L[R^{(1)};R^{(2)}]\right]= \tilde{\mathbb{P}}^{-1}_{\sigma}[\mathcal{E}] \tilde{\mathbb{E}}_{\sigma}\left[\mathbf{1}_{\mathcal{E}}L[R^{(1)};R^{(2)}]\right]\enspace . 
\]
By a union bound over the $\binom{l+1}2$ pairs $i\neq j$, each contributing at most $|W^*_i||W^*_j|/d$ to $\tilde{\mathbb{P}}_{\sigma}[\mathcal{E}^c]$ ($1/d$ being the probability that two elements of $[d]$, drawn independently and uniformly for $W^*_i$ and $W^*_j$, coincide), we get $\tilde{\mathbb{P}}_{\sigma}[\mathcal{E}^c]\leq |W^*|^2/(2d)\leq 2D^2/d$, using $|W^*|\leq 2D$. Since $2^8D^4\leq d$ and $D\geq 2$, we have $2D^2/d\leq 2^{-7}D^{-2}\leq 1/2$, which gives $ \tilde{\mathbb{P}}_{\sigma}[\mathcal{E}] \geq 1/2$. Since, under $\tilde{\mathbb{P}}_{\sigma}$, the expectation of $L^{R^{(1)};R^{(2)}}$ is zero, we obtain
\begin{align*}
\tilde{\mathbb{E}}_{\sigma}\left[\mathbf{1}_{\mathcal{E}}L[R^{(1)};R^{(2)}]\right]= - \tilde{\mathbb{E}}_{\sigma}\left[\mathbf{1}_{\mathcal{E}^c}L[R^{(1)};R^{(2)}]\right]
\end{align*}

In the sequel, $\mathcal{L}$ stands for a partition of $[0:l]$. Given any element $L\in \mathcal{L}$, we define the event $\mathcal{B}_{L}$ such that for any $a_0$ and $a'$ in $L$, there exist $a_1,\ldots, a_s$ in $L$ with  $\sigma(\pi_2(W^*_{a_i}))\cap  \sigma(\pi_2(W^*_{a_{i+1}}))\neq \emptyset$ for $i=0,\ldots, s-1$ and $\sigma(\pi_2(W^*_{a_s}))\cap  \sigma(\pi_2(W^*_{a'}))\neq \emptyset$. If $L$ is a singleton, this event is of probability one. Since the connecting chain is required to remain inside $L$, the event $\mathcal{B}_L$ only involves the variables $\sigma(\pi_2(W^*_a))$ with $a\in L$; in particular, under $\tilde{\mathbb{P}}_{\sigma}$, $\mathbf{1}_{\mathcal{B}_L}$ is independent of the randomness attached to the blocks $W^*_a$ with $a\notin L$. Also, we define  the event $\mathcal{E}_{\mathcal{L}}$ such that for any $a$ and $a'$ in two groups $L$ and $L'$ of $\mathcal{L}$, we have $\sigma(\pi_2(W^*_a))\cap\sigma(\pi_2(W^*_{a'}))=\emptyset$. Equipped with these definitions, we deduce that 
\[
\tilde{\mathbb{E}}_{\sigma}\left[\mathbf{1}_{\mathcal{E}}L[R^{(1)};R^{(2)}]\right]= - \sum_{\mathcal{L}} \tilde{\mathbb{E}}_{\sigma}\left[\mathbf{1}_{\mathcal{E}_{\mathcal{L}}}\prod_{L\in \mathcal{L}}\mathbf{1}_{\mathcal{B}_L}L[R^{(1)};R^{(2)}]\right]\ ,
\]
where the sum runs over all partitions $\mathcal{L}$ of $[0:l]$ that are distinct from $\{\{0\},\{1\},\ldots, \{l\}\}$.  If the partition $\mathcal{L}$ contains a singleton distinct from $\{0\}$, then, we deduce from the definition of $L[R^{(1)};R^{(2)}]$ that  
\[
\tilde{\mathbb{E}}_{\sigma}\left[\prod_{L\in \mathcal{L}}\mathbf{1}_{\mathcal{B}_L}L[R^{(1)};R^{(2)}]\right]= 0 
\]
so that 
\[
\tilde{\mathbb{E}}_{\sigma}\left[\mathbf{1}_{\mathcal{E}_{\mathcal{L}}}\prod_{L\in \mathcal{L}}\mathbf{1}_{\mathcal{B}_L}L[R^{(1)};R^{(2)}]\right]= - \tilde{\mathbb{E}}_{\sigma}\left[\mathbf{1}_{\mathcal{E}^c_{\mathcal{L}}}\prod_{L\in \mathcal{L}}\mathbf{1}_{\mathcal{B}_L}L[R^{(1)};R^{(2)}]\right]\enspace . 
\]
Hence, by a standard recursion, we arrive at 
\begin{align*}
\left|\tilde{\mathbb{E}}_{\sigma}\left[\mathbf{1}_{\mathcal{E}}L[R^{(1)};R^{(2)}]\right]\right|\leq \sum_{o=1}^{l+1}\sum_{\mathcal{L}^{(1)},\mathcal{L}^{(2)},\ldots, \mathcal{L}^{(o)}}\tilde{\mathbb{E}}\left[\mathbf{1}_{\mathcal{E}_{\mathcal{L}^{(o)}}}\prod_{L\in \mathcal{L}^{(o)}}\mathbf{1}_{\mathcal{B}_L}\left|L[R^{(1)};R^{(2)}]\right|\right]\ , 
\end{align*}
where  $\mathcal{L}^{(a+1)}$ is strictly coarser than $\mathcal{L}^{(a)}$ and where $\mathcal{L}^{(o)}$ is the only partition in the sequence that does not contain any singleton, with the possible exception of $\{0\}$.
Write $\boldsymbol{\mathcal{L}}^*$ for the collection of all partitions $\mathcal{L}$ that do not contain any singleton distinct from $\{0\}$. 

\begin{lemma}\label{lem:Ll}
Given $\mathcal{L}$ in $\boldsymbol{\mathcal{L}}^*$, there exist no more than $l^{2l}2^l$ sequences $\mathcal{L}^{(1)},\ldots, \mathcal{L}^{(o)}$ such that $\mathcal{L}^{(o)}=\mathcal{L}$.    
\end{lemma}

We conclude that
\begin{align}\label{eq:upper:L1}
\left|\tilde{\mathbb{E}}_{\sigma}\left[\mathbf{1}_{\mathcal{E}}L[R^{(1)};R^{(2)}]\right]\right|\leq l^{2l}2^l\sum_{\mathcal{L}\in \boldsymbol{\mathcal{L}}^*}\tilde{\mathbb{E}}_{\sigma}\left[\mathbf{1}_{\mathcal{E}_{\mathcal{L}}}\prod_{L\in \mathcal{L}}\mathbf{1}_{\mathcal{B}_L}\left|L[R^{(1)};R^{(2)}]\right|\right]\enspace . 
\end{align}

In the following, we write $\mathcal{S}$ for a partition of $W^*$ such that, for $i=0,\ldots, l$, elements of $W^*_i$ are not in the same group. 
Let us define the event $\mathcal{C}_{\mathcal{S}}$ such that, for any $w$ and $w'\in W^*$, we have  $\sigma(\pi_2(w))= \sigma(\pi_2(w'))$ if and only if $w$  and $w'$ are in the same group in $\mathcal{S}$. Given $\mathcal{L}\in \boldsymbol{\mathcal{L}}^*$, define $\boldsymbol{\mathcal{S}}[\mathcal{L}]$ as the collection of partitions $\mathcal{S}$ that are compatible with $\mathcal{E}_{\mathcal{L}}\cap_{L\in \mathcal{L}} \mathcal{B}_L$. Finally, we define $\boldsymbol{\mathcal{S}}^*=\cup_{\mathcal{L} \in \boldsymbol{\mathcal{L}}^*}\boldsymbol{\mathcal{S}}[\mathcal{L}]$. 
We then deduce from~\eqref{eq:definition:L:R1:R2} and \eqref{eq:upper:L1}  that 
\begin{align*}
\left|\tilde{\mathbb{E}}_{\sigma}\left[\mathbf{1}_{\mathcal{E}}L[R^{(1)};R^{(2)}]\right]\right|& \leq l^{2l}2^l\sum_{\mathcal{S}\in \boldsymbol{\mathcal{S}}^*}\tilde{\mathbb{E}}_{\sigma}\left[\mathbf{1}_{\mathcal{C}_{\mathcal{S}}}\left|L[R^{(1)};R^{(2)}]\right|\right] \\
& \leq  l^{2l}2^l\sum_{\mathcal{S}\in \boldsymbol{\mathcal{S}}^*} \sum_{0\in T\subset [0:l]}\tilde{\mathbb{E}}_{\sigma}\left[\mathbf{1}_{\mathcal{C}_{\mathcal{S}}}\prod_{i\in T}\mathbf{1}\left\{\mathcal{A}[G^*_{i}]\right\}\right]\prod_{i\in [0:l]\setminus T}\mathbb{E}_{\sigma}\left[\mathbf{1}\left\{\mathcal{A}[G^*_i]\right\}\right]
\enspace .\end{align*}
Define $r(\mathcal{S})= \sum_{S\in \mathcal{S}}[|S|-1]$.
Since $|W^*|\leq 2D$ and $d\geq 4D$ (in particular, $d\geq 2|W^*|$), we have
\[
\tilde{\mathbb{P}}_{\sigma}\left[\mathcal{C}_{\mathcal{S}}\right]\leq  \left(\frac{2}{d}\right)^{r(\mathcal{S})}\enspace .
\]
Also, given $T\subset[0:l]$ and a partition $\mathcal{S}$, we define the graph $G^*_{T,\mathcal{S}}$ as the union of the graphs $G^*_i$ for $i\in T$ where we also identify nodes $w$ of those graphs that belong to the same group of the partition $\mathcal{S}$. This leads us to 
\[
\tilde{\mathbb{E}}_{\sigma}\left[\mathbf{1}_{\mathcal{C}_{\mathcal{S}}}\prod_{i\in T}\mathbf{1}\left\{\mathcal{A}[G^*_{i}]\right\}\right]= \tilde{\mathbb{P}}_{\sigma}\left[\mathcal{C}_{\mathcal{S}}\right] \mathbb{E_{\sigma}} \left[\mathbf{1}\left\{\mathcal{A}[G^*_{T,\mathcal{S}}]\right\} \right]
\]
Putting everything together, and coming back from $\tilde{\mathbb{E}}_{\sigma}[\mathbf{1}_{\mathcal{E}}L[R^{(1)};R^{(2)}]]$ to $\mathbb{E}_{\sigma}[L[R^{(1)};R^{(2)}]]$ thanks to the identity $\mathbb{E}_{\sigma}[L[R^{(1)};R^{(2)}]]= \mathbb{E}_{\sigma}[\mathbf{1}_{\mathcal{E}}L[R^{(1)};R^{(2)}]]= \tilde{\mathbb{P}}^{-1}_{\sigma}[\mathcal{E}]\tilde{\mathbb{E}}_{\sigma}[\mathbf{1}_{\mathcal{E}}L[R^{(1)};R^{(2)}]]$ --the event $\mathcal{E}$ occurring $\mathbb{P}_{\sigma}$-almost surely, since under $\mathbb{P}_{\sigma}$ the map $w\mapsto \sigma(\pi_2(w))$ is injective on $W^*$-- together with $\tilde{\mathbb{P}}^{-1}_{\sigma}[\mathcal{E}]\leq 2$, which accounts for the extra factor $2$ in the constant below, we arrive at
\begin{align*}
\left|\mathbb{E}_{\sigma}\left[L[R^{(1)};R^{(2)}]\right]\right|&
 \leq  l^{2l}2^{2l+1}\sum_{\mathcal{S}\in \boldsymbol{\mathcal{S}}^*} \left(\frac{2}{d}\right)^{r(\mathcal{S})}\max_{0\in T\subset [0:l]}
\mathbb{E}_{\sigma}\left[\mathbf{1}\left\{\mathcal{A}[G^*_{T,\mathcal{S}}]\right\}\right]\prod_{i\in [0:l]\setminus T}\mathbb{E}_{\sigma}\left[\mathbf{1}\left\{\mathcal{A}[G^*_i]\right\}\right]\enspace . 
\end{align*}

We apply Lemmas~\ref{lem:number_solutions} and~\ref{lem:eqcounting} to the graph $G^*_{T,\mathcal{S}}$ and to each of the pieces $G^*_i$ with $i\in [0:l]\setminus T$; since $0\in T$, these pieces are connected components of $G^{(1)}$ or of $G^{(2)}$ and are in particular non-empty. All these graphs have even degrees (a symmetric difference of graphs with even degrees has even degrees, connected components of $G^{(1)}$ and $G^{(2)}$ have even degrees, and identifying nodes of $W$ adds up their degrees), they do not contain any isolated node, and their edge sets are disjoint with total cardinality $|E^*|\leq |E^{(1)}|+|E^{(2)}|\leq 2D$. If one of them does not satisfy Condition~$(\cW_G)$, then Lemma~\ref{lem:number_solutions} yields that the corresponding probability, hence the whole term indexed by $(T,\mathcal{S})$, vanishes --~see the Remark following Lemma~\ref{lem:eqcounting}~-- so that we may assume that all of them satisfy Condition~$(\cW_G)$. Each of these graphs is non-empty, except possibly $G^*_{T,\mathcal{S}}$ when $T=\{0\}$ and $G^*_0$ is empty; a non-empty one has at least $4$ edges (any non-empty connected component of a graph with even degrees does, see the end of this proof; and $G^*_{T,\mathcal{S}}$ contains a non-empty piece as soon as it is non-empty). Consequently, each application of Lemma~\ref{lem:eqcounting} to a non-empty such graph $H$ contributes a factor $2(2m)^{|E(H)|/2}\leq (4m)^{|E(H)|/2}$, since $2\leq 2^{|E(H)|/2}$, whereas an empty $G^*_{T,\mathcal{S}}$ simply contributes the factor $\mathbb{P}_{\sigma}[\mathcal{A}[G^*_{T,\mathcal{S}}]]=1$ (the System~\eqref{eq:startsys} being then empty), in accordance with $b(\emptyset)=0$. Multiplying these bounds and using that $\cc(G^*_i)=1$ for $i\in [0:l]\setminus T$ yields the prefactor $(4m)^{|E^*|/2}$ and the exponent $\kappa_{T,\mathcal{S}}$ below, where $|E^*|=\sum_{i=0}^l|E^*_i|$.

We obtain
\begin{align*}
\left|\mathbb{E}_{\sigma}\left[L[R^{(1)};R^{(2)}]\right]\right|
&\leq   l^{2l}2^{2l+1} (4m)^{|E^*|/2}\sum_{\mathcal{S}\in \boldsymbol{\mathcal{S}}^*} \left(\frac{2}{d}\right)^{r(\mathcal{S})}\max_{0\in T\subset [0:l]}
\left(\frac{1}{d}\right)^{\kappa_{T,\mathcal{S}}}
\enspace ,
\end{align*}
where we define
\begin{align}\label{eq:definition:kappa}
  \kappa_{T,\mathcal{S}}:=   \frac{|W^*_{T,\mathcal{S}}|+ \sum_{i\notin T}|W^*_i|}{2} - \frac{|E^*|-(2m+1)|V^*|}{4}- \frac{\cc(G^*_{T,\mathcal{S}})+ (l+1)-|T|}{2}\enspace . 
\end{align}

\begin{lemma}\label{lem:kappa_T_S}
For any $\mathcal{S}\in \boldsymbol{\mathcal{S}}^*$ and any $T\subset [0:l]$ with $0\in T$,
we have
\[
\kappa_{T,\mathcal{S}}+  r(\mathcal{S})  \geq  -\frac{[|E^*|-(2m+1)|V^*|]}{4} - \frac{\cc(G^*_{0})}{2} + \frac{|W^*|}{2} + \frac{(r(\mathcal{S})- l/2)_+} {2}  \enspace . 
\]
\end{lemma}

Summing over all possible partitions $\mathcal{S}$, we conclude that 
\begin{align*}
\left|\mathbb{E}_{\sigma}\left[L[R^{(1)};R^{(2)}]\right]\right|
&\leq    l^{2l}2^{2l+1} (4m)^{|E^*|/2}
\left(\frac{1}{d}\right)^{ -\frac{[|E^*|-(2m+1)|V^*|]}{4} - \frac{\cc(G^*_0)}{2} + \frac{|W^*|}{2} }   \sum_{r=\lceil l/2\rceil}^{\infty} 2^{r} \left(\frac{1}{d}\right)^{(r-l/2)/2} |W^*|^{2r}
\\
&\leq   l^{2l}2^{3l+1}(2D)^{l} (4m)^{|E^*|/2} \left(\frac{1}{d}\right)^{ -\frac{[|E^*|-(2m+1)|V^*|]}{4} - \frac{\cc(G^*_0)}{2} + \frac{|W^*|}{2} } \sum_{r=\lceil l/2\rceil}^{\infty}  \left(\frac{2^6D^4}{d}\right)^{(r-l/2)/2} \\
&\leq  2(32 l^2 D)^{l} (4m)^{|E^*|/2}\left(\frac{1}{d}\right)^{ -\frac{[|E^*|-(2m+1)|V^*|]}{4} - \frac{\cc(G^*_0)}{2} + \frac{|W^*|}{2} } \enspace ,
\end{align*}
where we used that $|W^*|\leq 2D$ in the second line, and that $2^8D^4 \leq d$ in the last line{ --~indeed, writing $q:=2^6D^4/d\leq 1/4$, the geometric-type sum satisfies $\sum_{r\geq \lceil l/2\rceil} q^{(r-l/2)/2}= q^{\epsilon}/(1-\sqrt q)\leq 1/(1-1/2)=2$, where $\epsilon=0$ if $l$ is even and $\epsilon=1/4$ if $l$ is odd~--}.

To conclude, we note that $l\leq D/2$ as any non-empty connected component of a graph with even degrees contains at least $4$ edges. Hence,
\[
2\,(32l^2D)^l\leq 2\,(8D^3)^l\leq 4^l(8D^3)^l= (32D^3)^l= (32D^3)^{\cc_{\mathrm{u}}}\enspace ,
\]
where we used that $l= |[\cc^{(1)}]\setminus R_0^{(1)}| + |[\cc^{(2)}]\setminus R_0^{(2)}|= \cc_{\mathrm{u}}$, the connected components outside $R_0^{(1)}$ and $R_0^{(2)}$ being exactly the unmatched ones. Finally, the exponent appearing above is nothing but $b_{R^{(1)},R^{(2)}}$, by the definition of $|E^*|$, $|V^*|$, $|W^*|$ and since $\cc(G^*_0)= \cc(G^{R^{(1)};R^{(2)}}_{\oplus})$. This concludes the proof.

\end{proof}

\begin{proof}[Proof of Lemma~\ref{lem:Ll}]
Let us justify this counting bound (recall that $l\geq 1$, as otherwise no partition $\mathcal{L}\neq \hat{0}$ exists). Write $\hat{0}=\{\{0\},\{1\},\ldots,\{l\}\}$ for the finest partition of $[0:l]$ and order the partitions of $[0:l]$ by refinement: $\pi\preceq \pi'$ if every group of $\pi$ is contained in a group of $\pi'$, and $\pi\prec \pi'$ if moreover $\pi\neq \pi'$. Since every $\mathcal{L}^{(a)}$ satisfies $\hat{0}\prec\mathcal{L}^{(1)}\prec \cdots\prec \mathcal{L}^{(o)}=\mathcal{L}$, it suffices to bound, for an arbitrary partition $\mathcal{L}\neq \hat{0}$ of $[0:l]$, the number of such strictly increasing sequences of arbitrary length $o\geq 1$ ending at $\mathcal{L}$; forgetting the additional constraint that $\mathcal{L}^{(1)},\ldots,\mathcal{L}^{(o-1)}$ contain a singleton distinct from $\{0\}$ can only increase this number. Set $\nu=l+1=|[0:l]|$, let $b_1,\ldots,b_k$ be the sizes of the $k\geq 1$ groups of $\mathcal{L}$, and let
\[
\chi:= \nu-k= \sum_{i=1}^k (b_i-1)\geq 1\enspace .
\]
Endow the partitions with the rank function $\pi\mapsto \nu-|\pi|$; then $\pi\prec \pi'$ is a covering relation --- that is, $\pi'$ is obtained from $\pi$ by merging exactly two of its groups --- if and only if the rank increases by exactly one, and the partitions $\pi$ with $\hat{0}\preceq \pi\preceq \mathcal{L}$ are exactly those obtained by partitioning each group of $\mathcal{L}$ separately, their rank ranging over $\{0,1,\ldots,\chi\}$.

Call \emph{saturated} a chain $\hat{0}=\tau_0\prec \tau_1\prec \cdots\prec \tau_\chi=\mathcal{L}$ all of whose steps are covering relations. Inside a fixed group of $\mathcal{L}$ of size $b$, one goes from the $b$ singletons to the whole group in $b-1$ merges and, as long as $j$ groups remain, $\binom{j}{2}$ pairs may be merged; there are therefore $\prod_{j=2}^{b}\binom{j}{2}= b!(b-1)!/2^{b-1}$ such saturated chains. As merges occurring in different groups of $\mathcal{L}$ can be freely interleaved, and as the $k$ resulting sequences, of respective lengths $b_1-1,\ldots,b_k-1$, can be interleaved into a single sequence of $\chi$ merges in $\chi!/\prod_{i}(b_i-1)!$ ways, the number of saturated chains from $\hat{0}$ to $\mathcal{L}$ equals
\begin{equation}\label{eq:saturated:chains}
\frac{\chi!}{\prod_{i=1}^k(b_i-1)!}\prod_{i=1}^k \frac{b_i!(b_i-1)!}{2^{b_i-1}}= \frac{\chi!}{2^\chi}\prod_{i=1}^k b_i! \enspace .
\end{equation}
Consider now any chain $\hat{0}=\mathcal{L}^{(0)}\prec \mathcal{L}^{(1)}\prec \cdots\prec \mathcal{L}^{(o)}=\mathcal{L}$. Since two comparable partitions are always related by successive merges of two groups, each step $\mathcal{L}^{(a-1)}\prec \mathcal{L}^{(a)}$ can be refined into a chain of covering relations, so that the whole chain admits at least one refinement into a saturated chain $(\tau_0,\ldots,\tau_\chi)$ from $\hat{0}$ to $\mathcal{L}$; fix one such refinement for each chain. The ranks $R:= \{\nu-|\mathcal{L}^{(a)}|,\ a=1,\ldots,o-1\}$ of the intermediate partitions form a subset of $\{1,\ldots,\chi-1\}$, and the pair $((\tau_0,\ldots,\tau_\chi),R)$ recovers the original chain, which is nothing but $(\tau_r)_{r\in R}$ followed by $\tau_\chi=\mathcal{L}$. This map being injective and there being $2^{\chi-1}$ possible subsets $R$, we deduce from~\eqref{eq:saturated:chains} that the number of chains ending at $\mathcal{L}$ is at most
\[
\frac{\chi!}{2^\chi}\left(\prod_{i=1}^k b_i!\right)2^{\chi-1}= \frac{\chi!}{2}\prod_{i=1}^k b_i! \leq \frac{l!\,(l+1)!}{2}\enspace ,
\]
where we used $\chi= \nu-k\leq \nu-1= l$ together with $\prod_{i=1}^k b_i!\leq \nu!=(l+1)!$, the latter because $\nu!/\prod_i b_i!=\binom{\nu}{b_1,\ldots,b_k}$ is a positive integer. Finally, $j!\leq j^j$ for any integer $j\geq 1$, and $l+1\leq 2^l$ for any integer $l\geq 1$, so that
\[
\frac{l!\,(l+1)!}{2}= \frac{(l+1)(l!)^2}{2}\leq \frac{l+1}{2}l^{2l}\leq 2^{l-1}l^{2l}\leq l^{2l}2^l\enspace ,
\]
which is the announced bound.
\end{proof}

\begin{proof}[Proof of Lemma~\ref{lem:kappa_T_S}]
Fix any $T\subset [0:l]$ with $0\in T$.
Define $r(\mathcal{S},T)= |W^*| - |W^*_{T,\mathcal{S}}| - \sum_{i\notin T}|W_i^*|$. 
Define 
\[
A_T:= r(\mathcal{S}) + \frac{- r(\mathcal{S},T)+ \cc(G^*_{0}) -(l+1)+ |T|-  \cc(G^*_{T,\mathcal{S}})}{2}
\]
It follows from the definition~\eqref{eq:definition:kappa} of $\kappa_{T,\mathcal{S}}$ that 
\[
\kappa_{T,\mathcal{S}}+  r(\mathcal{S})= -\frac{[|E^*|-(2m+1)|V^*|]}{4} - \frac{\cc(G^*_{0})}{2} + \frac{|W^*|}{2}  + A_T\ , 
\]
and we only have to establish that $A_T\geq (r(\mathcal{S}) - l/2)_+/2\geq0$ for all such $T$.
First, we establish that  $A_T\geq (r(\mathcal{S}) - l/2)_+/2\geq0$ for $T=[0:l]$. Indeed, for the latter we have 
\[
2A_{[0:l]} = r(\mathcal{S}) + \cc(G^*_{0}) - \cc(G^*_{[0:l],\mathcal{S}})\enspace . 
\]
Any partition $\mathcal{S}\in \boldsymbol{\mathcal{S}}^*$ satisfies the following property:
\begin{equation}\tag{$\mathcal{P}_{\mathcal{S}}$}\label{eq:property:S*}
\text{for any } i\in [l],\ \text{at least one node of } W_i^* \text{ belongs to the same group of } \mathcal{S} \text{ as a node of } W^*\setminus W_i^*\enspace .
\end{equation}
Indeed, let $\mathcal{L}\in \boldsymbol{\mathcal{L}}^*$ be such that $\mathcal{S}\in \boldsymbol{\mathcal{S}}[\mathcal{L}]$, and let $i\in [l]$. Since $\mathcal{L}$ contains no singleton distinct from $\{0\}$, the group $L\in \mathcal{L}$ containing $i$ contains some $a'\neq i$. On the event $\mathcal{B}_L$, the chain connecting $i$ to $a'$ has to leave $i$ at some step, so that $\sigma(\pi_2(W^*_i))\cap \sigma(\pi_2(W^*_a))\neq \emptyset$ for some $a\in L\setminus\{i\}$; by compatibility with $\mathcal{C}_{\mathcal{S}}$, the two corresponding nodes, which lie respectively in $W^*_i$ and in $W^*_a\subset W^*\setminus W^*_i$, belong to the same group of $\mathcal{S}$. As a consequence of~\eqref{eq:property:S*}, at least $l$ distinct nodes of $W^*$ --~one in each $W_i^*$, $i\in [l]$~-- belong to groups of $\mathcal{S}$ of size at least $2$. Since $|S|-1\geq |S|/2$ for any such group $S$, we get $r(\mathcal{S})=\sum_{S\in\mathcal{S}}(|S|-1)\geq l/2$.

Let us also check that $\cc(G^*_{[0:l],\mathcal{S}})\leq l/2+ \cc(G^*_0)$. 
 Call \emph{atoms} the connected components of the disjoint union $G^*_0\sqcup G^*_1\sqcup \cdots\sqcup G^*_l$: since each $G^*_i$ with $i\in [l]$ is connected, the atoms are the $\cc(G^*_0)$ connected components of $G^*_0$ together with the $l$ graphs $G^*_i$, $i\in [l]$. Let $\Gamma$ be the graph whose nodes are the atoms and where two distinct atoms are joined by an edge whenever some group of $\mathcal{S}$ contains a node of the first atom and a node of the second one. As $G^*_{[0:l],\mathcal{S}}$ is obtained from the above disjoint union by identifying the nodes lying in a common group of $\mathcal{S}$, two atoms belong to the same connected component of $G^*_{[0:l],\mathcal{S}}$ if and only if they belong to the same connected component of $\Gamma$. Since every atom carries at least one edge, $\cc(G^*_{[0:l],\mathcal{S}})$ is therefore exactly the number $p$ of connected components of $\Gamma$ --isolated nodes of $\Gamma$ being counted here. Besides, by Property~\eqref{eq:property:S*}, every atom $G^*_i$ with $i\in [l]$ has degree at least one in $\Gamma$, since a node of $W^*_i$ belongs to the same group of $\mathcal{S}$ as a node of another atom. Now, given a connected component $C$ of $\Gamma$, write $g_C$ for the number of its nodes that are connected components of $G^*_0$ and $l_C$ for the number of its nodes of the form $G^*_i$ with $i\in [l]$, so that $\sum_{C}g_C= \cc(G^*_0)$ and $\sum_C l_C= l$, the sums running over the $p$ connected components of $\Gamma$. If $g_C\geq 1$, then $1\leq g_C\leq g_C+ l_C/2$. If $g_C= 0$, then all the nodes of $C$ are of the form $G^*_i$ with $i\in [l]$; any of them has a neighbor in $\Gamma$, which necessarily lies in $C$, so that $l_C\geq 2$ and again $1\leq l_C/2= g_C+l_C/2$. Summing these inequalities over the connected components of $\Gamma$, we get
\[
\cc(G^*_{[0:l],\mathcal{S}})= p\leq \sum_C \Big(g_C+ \frac{l_C}{2}\Big)= \cc(G^*_0)+ \frac{l}{2}\enspace .
\]
This leads to  $A_{[0:l]}\geq (r(\mathcal{S}) - l/2)_+/2\geq0$. Finally, we show that $A_{T}\geq A_{[0:l]} $ for any $T\subset [0:l]$ with $0\in T$. To see this, observe that $G^*_{[0:l],\mathcal{S}}$ is built from $G^*_{T,\mathcal{S}}$ and from the connected components $G^*_i$ with $i\notin T$ by identifying nodes. Since each node identification among the $r(\mathcal{S})- r(\mathcal{S},T)$ ones merges at most two connected components, we arrive at 
\begin{align} \label{eq:upper:r}
r(\mathcal{S})- r(\mathcal{S},T)\geq  +\cc(G^*_{T,\mathcal{S}})+ (l+1 - |T|)-\cc(G^*_{[0:l],\mathcal{S}}) 
\end{align}
In turn, this implies that  
\[
2(A_{T}- A_{[0:l]}) =  r(\mathcal{S})- r(\mathcal{S},T)   + \cc(G^*_{[0:l],\mathcal{S}}) -\cc(G^*_{T,\mathcal{S}})- (l+1 - |T|) \enspace   
\]
is non-negative.
\end{proof}

\section{Proofs for the number of solutions of~\eqref{eq:startsys}} \label{sec:number_solution}

\subsection{Proof of Lemma~\ref{lem:number_solutions}}
 Assume that there exists $V' \subset V$ such that $|\oplus_{v\in V'} \cN(v)|\in [1:2m]$ - indeed, since we assumed that the degree of all nodes is even, the symmetric differences of neighborhoods have even cardinality, so that we have excluded the case $|\oplus_{v\in V'} \cN(v)| = 2m+1$. \begin{align*}
\mathbb P_{\sigma}[\cA_{G}]
&\leq \mathbb P_{\sigma}\Bigg[\sum_{v\in V'}\sum_{w: w\sim v} \phi^{(2s-1)}(\sigma(\pi_2(w))) = 0, \forall 1\leq s \leq m\Bigg],
\end{align*}
since if the system of equations~\eqref{eq:startsys} is satisfied, then any sum over $v$ in a subset of $V$ of the associated Equation~\eqref{eq:startsysune} needs to be equal to $0,$ for any $s\leq m$.

Note that
$$\sum_{v\in V'}\sum_{w: w\sim v} \phi^{(2s-1)}(\sigma(\pi_2(w))) = \sum_{w \in \oplus_{v\in V'} \cN(v)} \phi^{(2s-1)}(\sigma(\pi_2(w))),$$
by definition of the sum operation in $\mathbb F_2^u$ - as being the coordinatewise sum in $\mathbb N$, modulo $2$. Hence, since $| \oplus_{v\in V'} \cN(v)| \leq 2m$ and since the $\sigma(\pi_2(w))$ are all distinct for $w\in W$, we deduce from Lemma~\ref{lem:fundBCH}
that the system of equations
$$\sum_{v\in V'}\sum_{w: w\sim v} \phi^{(2s-1)}(\sigma(\pi_2(w))) = 0, \quad \quad \forall 1\leq s \leq m,$$
has no solution whatever the injection $\sigma$ is. This completes the proof.

\subsection{Proof of Lemma~\ref{lem:eqcounting}}

This proof is based on standard arguments in algebraic geometry over finite fields. Recall that $d+1=2^u$ is a power of 2, and that, throughout, a column $w\in W$ is identified with the value $\sigma(\pi_2(w))\in\{1,\ldots,d\}$, itself identified with a nonzero element of $\mathbb F_{d+1}=\mathbb F_{2^u}$ as in Section~\ref{sec:bch:code}; we also use that $G$ has no isolated node by assumption. Since the Zariski topology, dimension, and degree of an algebraic set are only meaningfully defined over an algebraically closed field, we view all the algebraic sets introduced below ($\mathbb X$, $\mathbb U$, $\overline{\mathbb U}$, and their irreducible components) as varieties over the algebraic closure $\overline{\mathbb F}_{d+1}$ of $\mathbb F_{d+1}$. For such a variety $\mathbb Y\subset \overline{\mathbb F}_{d+1}^{W}$, we write $|\mathbb Y|$ for its number of $\mathbb F_{d+1}$-\emph{rational} points, that is $|\mathbb Y\cap \mathbb F_{d+1}^{W}|$ -- this is the only quantity we are ultimately interested in, while $\dim$ and $\deg$ below refer to the geometric invariants of $\mathbb Y$ as a variety over $\overline{\mathbb F}_{d+1}$.

Define the affine algebraic set $\mathbb{X}\subset \overline{\mathbb{F}}_{d+1}^{W}$ defined by the homogeneous equation system
\begin{equation}\label{eq:system}
\sum_{w: w\sim_G v} x_w^{2l-1}= 0\ ,
\end{equation}
for $l=1,\ldots, m$ and for $v\in V$. Note that this system is exactly the system~\eqref{eq:startsys}, read in $\mathbb F_{d+1}$ rather than in $\mathbb F_2^u$: since $\phi^{(2l-1)}(v)=\psi(v^{2l-1})$ and since $\psi$ is an isomorphism of additive groups from $\mathbb F_{2^u}$ onto $\mathbb F_2^u$, we have, for any $v\in V$ and any $l\leq m$,
$$\sum_{w:w\sim_Gv}\phi^{(2l-1)}(\sigma(\pi_2(w))) = \psi\Big(\sum_{w:w\sim_Gv}\sigma(\pi_2(w))^{2l-1}\Big)\enspace ,$$
which vanishes in $\mathbb F_2^u$ if and only if $\sum_{w:w\sim_Gv}\sigma(\pi_2(w))^{2l-1}=0$ in $\mathbb F_{d+1}$.
We define $\mathbb{U} \subset \mathbb{X}$ as the Zariski open subset such that the $x_w$ are distinct and nonzero -- matching the fact that $\sigma(\pi_2(w))\in\{1,\ldots,d\}$ is identified with a nonzero element of $\mathbb F_{d+1}$. That is, $\mathbb{U} = \mathbb{X} \setminus \mathbb{Z}$, where $\mathbb{Z}$ is the union of hyperplanes $x_w - x_{w'} = 0$ for all $w\neq w'$, and of hyperplanes $x_w=0$ for all $w$.
By definition of~\eqref{eq:startsys} and of the event $\mathcal{A}[G]$, we have
\begin{equation}\label{eq:proba:A[G]}
\mathbb{P}_{\sigma}[\cA_{G}]=  \frac{(d-|W|)!}{d!} |\mathbb{U}|\ ,
\end{equation}
where $|\mathbb U|$ is, as agreed above, the number of $\mathbb F_{d+1}$-rational points of $\mathbb U$ -- namely, the number of injective tuples $(x_w)_{w\in W}$ with values in the $d$ nonzero elements of $\mathbb F_{d+1}$ satisfying the system~\eqref{eq:system}. Thus, we look for an upper bound on this number of rational points of the set $\mathbb{U}$. The proof proceeds in two steps. First, we bound the dimension of the geometric affine variety and then, we apply the Bézout inequality.

The set $\mathbb{U}$ is a Zariski-open subset of $\mathbb{X}$, which is in general reducible and not closed. For this reason, we introduce  $\overline{\mathbb{U}}$ as the Zariski closure of $\mathbb{U}$ in $\overline{\mathbb{F}}_{d+1}^{W}$. Then $\overline{\mathbb{U}}$ is an affine algebraic set.
Let $\mathbb{V}$ denote an irreducible component of $\bar{\mathbb{U}}$. Since $\mathbb{U}$ is dense in $\overline{\mathbb{U}}$, $\mathbb{V} \cap \mathbb{U}$ is a non-empty open subset of $\mathbb{V}$. For any point $x=(x_w)_{w\in W} \in \mathbb{V} \cap \mathbb{U}$, we consider the Jacobian matrix $\mathcal{J}_{x}\in \overline{\mathbb{F}}_{d+1}^{m\cdot |V|\times  |W|}$ associated to the system~\eqref{eq:system} obtained by stacking vertically the blocks $\cJ_{x,1},\ldots,\cJ_{x,m}$, so that its rows are indexed by the pairs $(i,v)\in[m]\times V$, where
\begin{align}\label{eq:definition:Jacobian}
(\cJ_{x,i})_{v,w}= x_w^{2i-2} \mathbf{1}_{w\in \cN(v)}\ ,
\end{align}
for all $v\in V$ and $w\in W$. Here, we used that the derivative of $x_w^{2i-1}$ with respect to $x_w$ is $(2i-1)x_w^{2i-2}=x_w^{2i-2}$, since $2i-1$ is odd and $\mathbb F_{d+1}$ has characteristic $2$.

\begin{lemma}\label{lem:rank_Jacobian}
    For any such $x\in \mathbb{V} \cap \mathbb{U}$, we have 
    \[
    \text{rank}(\mathcal{J}_x) \geq b(G)
    \]
\end{lemma}

Then, since the equations~\eqref{eq:system} vanish on $\mathbb{V}$, the tangent space of $\mathbb{V}$ at $x$ is contained in the kernel of the Jacobian matrix $\cJ_x$, and its dimension is at least $\dim(\mathbb{V})$ --see, e.g.,~\cite[Chapter~2, Section~1]{shafarevich2013basic}. Hence, by Lemma~\ref{lem:rank_Jacobian}, and since $\mathrm{rank}(\cJ_x)$ is an integer,
\[ \dim(\mathbb{V}) \leq |W| - \mathrm{rank}(\cJ_x) \leq  |W| -  \lceil b(G)\rceil. \]
Since the dimension of an algebraic set is the maximum of the dimensions of its components,  we conclude
\[ \dim(\overline{\mathbb{U}}) \leq |W| -  \lceil b(G)\rceil . \]

In what follows, we use the cumulative degree of~\cite[Section~1]{lachaud2015number}, which we simply denote by $\deg$: the degree $\deg(\mathbb{Y})$ of a possibly reducible affine algebraic set $\mathbb{Y}$ is the sum of the degrees of \emph{all} the irreducible components of $\mathbb{Y}$ --~and not only of those of maximal dimension~--.
We apply the Bézout bound for the number of $\mathbb F_{d+1}$-rational points on an affine algebraic set --see Theorem 2.1 in~\cite{lachaud2015number}-- to $\overline{\mathbb{U}}$:
\[ |\mathbb{U}| \leq |\overline{\mathbb{U}}| \leq \deg(\overline{\mathbb{U}}) \cdot (d+1)^{\dim(\overline{\mathbb{U}})}. \]
 Furthermore, since $\overline{\mathbb{U}}$ is a union of irreducible components of the algebraic set $\mathbb{X}$, its degree is bounded by the degree of $\mathbb{X}$, and more specifically by the product of the degrees of the equations~\eqref{eq:system} --see e.g.~Theorem~1.1 and Equation~(3) in~\cite{lachaud2015number}.
\[ \deg(\overline{\mathbb{U}}) \leq \deg(\mathbb{X}) \leq \prod_{j=1}^m (2j-1)^{|V|}. \]
Since all nodes $v\in V$ have a degree larger than or equal to $2m+2$ --~this follows from Condition~$(\cW_G)$ applied with $V_0=\{v\}$, which gives $\mathrm{deg}_G(v)=|\cN(v)|\notin [1:2m+1]$, together with $\mathrm{deg}_G(v)\geq 1$, since $G$ contains no isolated node~--, we have $|E|=\sum_{v\in V}\mathrm{deg}_G(v)\geq (2m+2)|V|$, and, in particular, $m|V|\leq |E|/2$.
We conclude --~using that $2j-1\leq 2m$ for any $j\leq m$, so that $\prod_{j=1}^m (2j-1)^{|V|}\leq (2m)^{m|V|}\leq (2m)^{|E|/2}$, where the last inequality relies on $m|V|\leq |E|/2$ and $2m\geq 1$~-- that
\[ |\overline{\mathbb{U}}| \leq (2m)^{ |E|/2} (d+1)^{|W|- \lceil b(G)\rceil}\enspace.  \]
Coming back to~\eqref{eq:proba:A[G]}, we now convert this bound in terms of $d+1$ into a bound in terms of $d$. Write $(d+1)^{|W|-\lceil b(G)\rceil}= d^{|W|-\lceil b(G)\rceil}\left(1+\frac1d\right)^{|W|-\lceil b(G)\rceil}$. Since each node of $W$ (resp. each of the $\cc(G)$ non-trivial connected components) is incident to at least one edge, we have $|W|\leq |E|$ and $\cc(G)\leq|W|\leq|E|$; together with $|E|\leq 2D$, this gives $|W|\leq 2D$ and $\cc(G)\leq 2D$. By definition~\eqref{eq:definition:b(G)} of $b(G)$ and since $m\geq 0$ and $\lceil b(G)\rceil\geq b(G)$,
\[
|W|-\lceil b(G)\rceil\leq\ |W|- b(G)= \frac{|W|}{2}+\frac{|E|}{4}-\frac{|V|}{2}\Big(m+\frac12\Big)+\frac{\cc(G)}{2}\leq \frac{|W|}{2}+\frac{|E|}{4}+\frac{\cc(G)}{2}\leq \frac{5|E|}{4}\leq \frac{5D}{2}\ .
\]
Since $D\geq 2$ and $d\geq 2^8D^4$, we have $\frac{5D}{2d}\leq \frac{5D}{2\cdot 2^8D^4}=\frac{5}{2^9D^3}\leq 2^{-8}$, so that $\left(1+\frac1d\right)^{|W|-\lceil b(G)\rceil}\leq \left(1+\frac1d\right)^{5D/2}\leq e^{5D/(2d)}\leq e^{2^{-8}}$. On the other hand, using $-\ln(1-x)\leq 2x$ for $x\leq 1/2$ together with $|W|\leq 2D$ and $|W|/d\leq 1/2$ (which holds since $d\geq2^8D^4\geq 4D\geq2|W|$), we get $\left(1-\frac{|W|}{d}\right)^{-|W|}\leq e^{2|W|^2/d}\leq e^{8D^2/d}\leq e^{8/(2^8D^2)}\leq e^{2^{-7}}$, again using $D\geq2$. Multiplying the two bounds, $\left(1+\frac1d\right)^{|W|-\lceil b(G)\rceil}\left(1-\frac{|W|}{d}\right)^{-|W|}\leq e^{2^{-8}+2^{-7}}=e^{3/256}\leq 2$. We conclude that
\begin{align*}
\mathbb{P}_{\sigma}[\cA_{G}]&=  \frac{(d-|W|)!}{d!} |\mathbb{U}|
\leq \left(1+\frac1d\right)^{|W|-\lceil b(G)\rceil}\left(1- \frac{|W|}{d}\right)^{-|W|} (2m)^{ |E|/2}\left(\frac{1}{d}\right)^{\lceil b(G)\rceil}\leq 2(2m)^{|E|/2}\left(\frac1d\right)^{\lceil b(G)\rceil}\enspace ,
\end{align*}
which establishes~\eqref{eq:bound:proba:solutation} and concludes the proof of Lemma~\ref{lem:eqcounting}.

\begin{proof}[Proof of 
Lemma~\ref{lem:rank_Jacobian}]

Given a graph $G=(V,W,E)$, a subset $\overline{W}\subset W$ satisfies the $G$-{\bf enumeration property} if there exists an ordering $(v^{(1)},\ldots, v^{(|V|)})$ of the nodes in $V$ such that, upon recursively defining $\overline{W}^{(0)}=\overline{W}$ and $\overline{W}^{(t)}= \overline{W}^{(t-1)}\setminus \cN(v^{(t)})$, we have
\begin{align}\label{eq:definition:enumeration:property}
 |\overline{W}^{(t-1)}\cap \cN(v^{(t)})|\leq m  \quad \quad \forall t=1,\ldots, |V|\enspace .
\end{align}

\begin{lemma}\label{lem:vandermonde}
Consider  $\overline{W}\subset W$ that satisfies the $G$-enumeration property and such that every $w\in \overline{W}$ has at least one neighbor in $V$ --~which is in particular the case as soon as $G$ does not contain any isolated node~--. Then,
we have 
\[
   \text{rank}(\mathcal{J}_x)\geq |\overline{W}|
\]
\end{lemma}

Hence, the key technical step of the proof amounts to finding $\overline{W}$ with $|\overline{W}|\geq b(G)$ that satisfies the $G$-enumeration property. We prove this in several steps.

\begin{lemma}\label{lem:deg2}
Consider any simple connected bipartite graph $G=(V,W,E)$ such that:
\begin{itemize}
    \item[(i)] Each node $w\in W$ satisfies $deg_G(w)=2$
    \item[(ii)] Each node $v\in V$ satisfies $deg_G(v)\geq m+1$. 
\end{itemize}
Then, there exists $\overline{W}\subset W$ that satisfies the $G$-enumeration property and such that 
\[
|\overline{W}| \geq \frac{|V|}{2}(m+1/2)- \frac{1}{2}\enspace . 
\]
\end{lemma}

Consider a connected graph $G=(V,W,E)$. We say that a group $V^*\subset V$ is of {\bf type $A$} if there exists  a node $w^*\in W$ with  $\mathrm{deg}_G(w^*)\geq 4$ such that
\begin{itemize}
    \item[(i)] Aside from $w^*$, $V^*$ is only adjacent to nodes $w$ of degree $2$
    \item[(ii)] $w^*$ is connected to exactly two nodes in $V^*$
    \item[(iii)] If we remove $w^*$ from $G$, then $V^*$ remains connected but is disconnected from $V\setminus V^*$.
\end{itemize}

\begin{lemma}\label{lem:deg_general}
Consider a connected graph $G=(V,W,E)$ that satisfies Condition \ref{condition_W_G} ($\cW_G$) and that does not contain any isolated node.  If $G$ does not contain any group of type $A$, then there exists $\overline{W}\subset W$ that satisfies the $G$-enumeration property and such that 
\begin{align}\label{eq:ojective_W_bar}
|\overline{W}|\geq  \frac{|W|}{2} - \frac{|E|}{4} + \frac{|V|}{2} (m+1/2) - \frac{1}{2}\enspace .
\end{align}
\end{lemma}

\begin{lemma}\label{lem:deg:generat:type_A}
Consider a graph $G=(V,W,E)$ that satisfies Condition \ref{condition_W_G} ($\cW_G$) and that does not contain any isolated node. There exists $\overline{W}\subset W$ that satisfies the $G$-enumeration property and such that 
\begin{align}\label{eq:ojective_W_bar2}
|\overline{W}|\geq  \frac{|W|}{2} - \frac{|E|}{4} + \frac{|V|}{2} (m+1/2) - \frac{\cc(G)}{2}\enspace .
\end{align}

\end{lemma}

The result readily follows by combining Lemma~\ref{lem:vandermonde} with the last lemma, the additional assumption of Lemma~\ref{lem:vandermonde} being satisfied since the graph $G$ of Lemma~\ref{lem:eqcounting} does not contain any isolated node.
\end{proof}

\begin{proof}[Proof of Lemma~\ref{lem:vandermonde}]
Consider a $\overline{W}\subset W$ that satisfies the $G$-enumeration property and let $(v^{(1)},\ldots, v^{(|V|)})$ be an ordering of the nodes that satisfies~\eqref{eq:definition:enumeration:property}. 

We shall prove that the columns of the matrix $\mathcal{J}_x$ indexed by $\overline{W}$ are linearly independent. Given $w\in W$, we write $(\mathcal{J}_{x})_{\cdot,w}$ for the $w$-th column of $\mathcal{J}_{x}$. Consider any $(\lambda_w)_{w\in \overline{W}}\in \overline{\mathbb{F}}_{d+1}^{\overline{W}}$ such that
\[
\sum_{w\in \overline{W}} \lambda_w (\mathcal{J}_{x})_{\cdot,w} = 0 \enspace. 
\]

We shall prove by induction on $t$ that $\lambda_w=0$ for all $w\in \overline{W}\setminus \overline{W}^{(t)}$. Since we have $\overline{W}^{(|V|)}=\emptyset$ --~indeed, $\overline{W}^{(|V|)}= \overline{W}\setminus \cup_{t=1}^{|V|}\cN(v^{(t)})$ is the set of the nodes $w\in \overline{W}$ that have no neighbor in $V$, and there is no such node by assumption~--, this will imply that $(\lambda_w)=0$ and therefore that  $\text{rank}(\mathcal{J}_x)\geq |\overline{W}|$.

The property is obviously true for $t=0$. Let us assume that it holds at some $t\geq 0$. This implies that
\begin{align}\label{eq:condition_lambda_w}
\sum_{w\in \overline{W}^{(t)}} \lambda_w (\mathcal{J}_{x})_{\cdot,w} = 0 \enspace. 
\end{align}
Define $r= |\cN(v^{(t+1)})\cap \overline{W}^{(t)} |$. By Condition~\eqref{eq:definition:enumeration:property}, we have $r\in [0:m]$. If $r=0$, we do not have anything to prove and we assume henceforth $r\in [1:m]$. Let us focus on the rows indexed
$(\mathcal{J}_{x,i})_{v^{(t+1)},\cdot}$ for $i=1,\ldots, r$. The expression~\eqref{eq:definition:Jacobian} of the Jacobian matrix and Condition~\eqref{eq:condition_lambda_w} imply that, for all $i=1,\ldots, r$, we have 
\[
\sum_{w\in \cN(v^{(t+1)})\cap \overline{W}^{(t)} } \lambda_w (x_{w}^2)^{i-1}=0\enspace .
\]
Since $x\in \mathbb{V}\cap \mathbb{U}$, the $x_w$ are distinct, and since squaring is injective in characteristic $2$ (if $x_w^2=x_{w'}^2$ then $(x_w+x_{w'})^2=x_w^2+x_{w'}^2=0$, so $x_w=x_{w'}$), the $x_{w}^2$ are distinct too; the corresponding Vandermonde matrix is thus invertible. This implies that $\lambda_w=0$ for all $w\in \cN(v^{(t+1)})\cap \overline{W}^{(t)}$. Hence, we have $\lambda_w=0$ for all $w\in \overline{W}\setminus \overline{W}^{(t+1)}$.
\end{proof}
\subsection{Proof of technical lemmas}

\begin{proof}[Proof of Lemma~\ref{lem:deg2}]

Note first that $|V|\geq 2$: if $V=\{v\}$, no node of $W$ could have two distinct neighbors, so that $W=\emptyset$, which contradicts $\mathrm{deg}_G(v)\geq m+1$. We prove the lemma by strong induction on $|V|$.

\emph{Step 0: pruning.} We iteratively remove nodes of $W$ (together with their two edges) as long as the resulting graph remains connected and all the nodes of $V$ keep a degree at least $m+1$. Write $G_{\mathrm p}=(V,W_{\mathrm p},E_{\mathrm p})$ for the pruned graph. If $\overline{W}\subset W_{\mathrm p}$ satisfies the $G_{\mathrm p}$-enumeration property, then it also satisfies the $G$-enumeration property, since $\cN_G(v)\cap A=\cN_{G_{\mathrm p}}(v)\cap A$ for all $v\in V$ and all $A\subset W_{\mathrm p}$. As the claimed lower bound on $|\overline{W}|$ does not depend on $W$, it is therefore enough to prove the lemma for $G_{\mathrm p}$, which we still denote $G=(V,W,E)$. This graph is simple, connected, bipartite, satisfies (i) and (ii), and, by maximality of the pruning, it also satisfies
\begin{itemize}
    \item[(iii)] for any $w\in W$ such that the graph $G-w$ obtained by removing $w$ is connected, one of the two neighbors of $w$ has degree exactly $m+1$.
\end{itemize}

\emph{Step 1: a node of degree $m+1$ on a cycle.} All the nodes of $G$ have degree at least $2$, so that $G$ contains a cycle, and, $G$ being bipartite, this cycle contains a node $w\in W$. As $\mathrm{deg}_G(w)=2$, both neighbors of $w$ lie on the cycle, so that they remain connected after the removal of $w$; hence $G-w$ is connected and, by (iii), $w$ has a neighbor $v$ such that $\mathrm{deg}_G(v)=m+1$. The cycle goes through $v$, which it leaves through $w$ and through another node $w'\in\cN_G(v)\setminus\{w\}$.

\emph{Step 2: peeling and induction.} Set $v^{(1)}:=v$. Then, as long as there exists a node $u\in V\setminus\{v^{(1)},\ldots,v^{(t)}\}$ such that $|\cN_G(u)\setminus \cN_G(\{v^{(1)},\ldots,v^{(t)}\})|\leq m$, where $v^{(1)},\ldots,v^{(t)}$ are the nodes selected so far, we select such a node as $v^{(t+1)}$. Let $s\geq 1$ be the number of selected nodes, and write $V^{(0)}=\{v^{(1)},\ldots,v^{(s)}\}$, $W^{(0)}=\cN_G(V^{(0)})$, and $\overline{W}^{(0)}=W^{(0)}\setminus\{w\}$. For $t=1$, we have $|\overline{W}^{(0)}\cap\cN_G(v^{(1)})|=|\cN_G(v)\setminus\{w\}|=m$ and, for $t\geq 2$, $\overline{W}^{(0)}\cap \cN_G(v^{(t)})\setminus \cN_G(\{v^{(1)},\ldots,v^{(t-1)}\})$ has at most $m$ elements by the selection rule.

Let $G'$ be the graph obtained from $G$ by removing $V^{(0)}$ and $W^{(0)}$, and let $C_1,\ldots,C_r$ be its connected components ($r=0$ if $G'$ is empty). A node of $W\setminus W^{(0)}$ is not adjacent to $V^{(0)}$, so that it keeps its two neighbors in $G'$, and, by the stopping rule, every node $u$ of $V\setminus V^{(0)}$ satisfies $\mathrm{deg}_{G'}(u)=|\cN_G(u)\setminus W^{(0)}|\geq m+1$. Hence, each $C_k$ is a simple connected bipartite graph that satisfies (i) and (ii) and has fewer nodes in $V$ than $G$. By the induction hypothesis, there exists $\overline{W}_k\subset W(C_k)$ that satisfies the $C_k$-enumeration property and such that $|\overline{W}_k|\geq \frac{|V(C_k)|}{2}(m+1/2)-\frac12$.

Define $\overline{W}=\overline{W}^{(0)}\cup\overline{W}_1\cup\cdots\cup\overline{W}_r$, and order the nodes of $V$ by first taking $v^{(1)},\ldots,v^{(s)}$, and then the nodes of $C_1,\ldots,C_r$, each $C_k$ according to an ordering witnessing the $C_k$-enumeration property of $\overline{W}_k$. Then $\overline{W}$ satisfies the $G$-enumeration property. Indeed, the nodes of $\overline{W}_1,\ldots,\overline{W}_r$ are not adjacent to $V^{(0)}$, so that~\eqref{eq:definition:enumeration:property} holds for $v^{(1)},\ldots,v^{(s)}$ by the above. When a node $u$ of $C_k$ is reached, all the nodes of $\overline{W}^{(0)}$, which are adjacent to $V^{(0)}$, have already been removed, the nodes of $\overline{W}_j$ with $j\neq k$ are not adjacent to $u$, and $\cN_G(u)\cap\overline{W}_k=\cN_{C_k}(u)\cap\overline{W}_k$, so that~\eqref{eq:definition:enumeration:property} holds as in $C_k$.

\emph{Step 3: counting.} Let $\Sigma$ be the number of edges between $W^{(0)}$ and $V\setminus V^{(0)}$. Since every node of $W^{(0)}$ has degree $2$ and all the neighbors of the nodes of $V^{(0)}$ belong to $W^{(0)}$, we have
\[
2|W^{(0)}|=\sum_{u\in V^{(0)}}\mathrm{deg}_G(u)+\Sigma\geq (m+1)s+\Sigma\enspace .
\]
As $|\overline{W}^{(0)}|=|W^{(0)}|-1$ and $\sum_{k\leq r}|V(C_k)|=|V|-s$, we deduce that
\[
|\overline{W}|\geq \frac{(m+1)s+\Sigma}{2}-1+\frac{|V|-s}{2}(m+1/2)-\frac r2= \frac{|V|}{2}(m+1/2)-\frac12+x\ ,\qquad\text{where } x:=\frac s4+\frac{\Sigma-r}{2}-\frac12\enspace .
\]
It remains to prove that $x\geq 0$. Since $G$ is connected and the nodes of $W\setminus W^{(0)}$ have all their neighbors in $G'$, each component $C_k$ is joined to the rest of $G$ by at least one edge between $W^{(0)}$ and $V(C_k)$; hence $\Sigma\geq r$, and $x\geq 0$ as soon as $s\geq 2$. If $s=1$, that is $V^{(0)}=\{v\}$, then $W^{(0)}=\cN_G(v)$ and $\Sigma=m+1$, as each of the $m+1$ neighbors of $v$ has exactly one other neighbor. Besides, the part of the cycle of Step 1 that joins the other neighbor of $w$ to the other neighbor of $w'$ avoids $v$, hence also $\cN_G(v)$, since a node of $\cN_G(v)$ has $v$ as one of its only two neighbors; it is therefore a path in $G'$. Thus, two of the $m+1$ edges counted in $\Sigma$ lead to the same component of $G'$ and, as each component receives at least one of them, $r\leq m$. This gives $x\geq \frac14+\frac12-\frac12>0$, which concludes the induction.

\end{proof}

\begin{proof}[Proof of Lemma~\ref{lem:deg_general}]
The case where each node $w\in W$ satisfies $\mathrm{deg}_G(w)=2$ has been dealt with in Lemma~\ref{lem:deg2}. Hence, we assume henceforth that at least one node $w\in W$ satisfies $\mathrm{deg}_G(w)\geq 4$.

We shall build the set $\overline{W}$ iteratively.  We define $W^{(0)}=W$, $V^{(0)}=V$, $\overline{W}^{(0)}=\emptyset$, and $G^{(0)}= G$.

First, since all nodes $v\in V$ have degree at least $2m+2$ --~this follows from Condition ($\cW_G$) applied with $V_0=\{v\}$, which gives $\mathrm{deg}_G(v)=|\cN_G(v)|\notin [1:2m+1]$, together with $\mathrm{deg}_G(v)\geq 1$ since $G$ contains no isolated node~--, we remove the node $w^*$ with the highest degree{, which does not create any node $v\in V$ of degree at most $m$}. In other words, we define $W^{(1)}= W^{(0)}\setminus \{w^*\}$, $\overline{W}^{(1)}=\emptyset$, and we fix $G^{(1)}$ to be the graph obtained from $G$ by pruning $w^*$. More generally, at time $t$, we check whether $W^{(t-1)}$ still contains a node $w$ with degree $\mathrm{deg}_G(w)\geq 4$. If no such node exists, we stop and we define $t^*=t-1$. Otherwise, we take $w$ to be a node in $W^{(t-1)}$ with the largest degree. We define $W^{(t)}= W^{(t-1)}\setminus \{w\}$ and $V^{(t)}= V^{(t-1)}$ and we define $G^{(t)}$ as the graph induced by $V^{(t)}$ and $W^{(t)}$, and we set $\overline{W}^{(t)}= \overline{W}^{(t-1)}$. If all the nodes $v\in V^{(t)}$ satisfy $\mathrm{deg}_{G^{(t)}}(v)\geq m+1$, then we go to $t+1$. If, instead, some node $v\in V^{(t)}$ satisfies $\mathrm{deg}_{G^{(t)}}(v)\leq  m$, then we update $\overline{W}^{(t)}= \overline{W}^{(t)}\cup [W^{(t)}\cap \cN_{G^{(t)}}(v)]$, we update 
$W^{(t)}= W^{(t)}\setminus  [W^{(t)}\cap \cN_{G^{(t)}}(v)]$, $V^{(t)}=V^{(t)}\setminus \{v\}$, $G^{(t)}$ as the graph induced by $V^{(t)}$ and $W^{(t)}$. We continue updating $(\overline{W}^{(t)}, V^{(t)},W^{(t)},G^{(t)})$ until the graph $G^{(t)}$ does not contain any node $v\in V^{(t)}$ with $\mathrm{deg}_{G^{(t)}}(v)\leq m$.

It readily follows from these definitions that the graph $G^{(t^*)}$ is either empty or is such that all $w\in W^{(t^*)}$ satisfy $\mathrm{deg}_G(w)=2$ and all $v\in V^{(t^*)}$ satisfy $\mathrm{deg}_{G^{(t^*)}}(v)\geq m+1$.  Denote by $c$ the number of connected components of $G^{(t^*)}$. Then, we apply  Lemma~\ref{lem:deg2} to construct $c$ sets $\overline{W}^{*(1)},\ldots, \overline{W}^{*(c)}$ such that $\cup_{i=1}^c\overline{W}^{*(i)}$ satisfies the $G^{(t^*)}$-enumeration property. Besides, we have 
\begin{align}\label{eq:lower_W^*i}
   \sum_{i=1}^c  |\overline{W}^{*(i)}| \geq \frac{|V^{(t^*)}|}{2}(m+1/2) - \frac{c}{2}
\end{align}

One readily checks from the iterative construction that  $\overline{W}^{(t^*)}\cup_{i=1}^c \overline{W}^{*(i)}$ satisfies the $G$-enumeration property.

\medskip 

Let us bound $|\overline{W}^{(t^*)}|$. 
For that purpose, we shall first look at the form of $\cN_G(v)\cap \overline{W}^{(t^*)}$ for each $v\in V\setminus V^{(t^*)}$. 
\begin{lemma}\label{claim:degree}
Consider any node $v\in V\setminus V^{(t^*)}$. We have    $|\cN_G(v)\cap \overline{W}^{(t^*)}|\geq m$ and it is possible to order the neighbors $w\in \cN_G(v)$ by non-decreasing degree $\mathrm{deg}_G(w)$ in such a way that the first $m$ nodes of this ordering all belong to $\overline{W}^{(t^*)}$.
\end{lemma}

This lemma is straightforward in light of the recursive definitions of the sets and we do not provide any proof. 
Let us define $V_{\mathrm a}\subset V\setminus V^{(t^*)}$ as the subset of nodes $v$ such that at least $m+1$ nodes in $\cN_G(v)$ are of degree $2$.

\begin{lemma}\label{claim:degree_2}
If $v\in V_{\mathrm a}$, then $\overline{W}^{(t^*)}$ contains at least $m+1$ neighbors of $v$ of degree $2$.
\end{lemma}
\begin{proof}
First, observe that $\overline{W}^{(t^*)}\cup W^{(t^*)}$ contains all the degree-$2$ nodes of $W$. Since $v\in V_{\mathrm a}$ has been pruned at some $t$, this means that all its neighbors have either been added to $\overline{W}^{(t)}$ or been pruned. Since no degree-$2$ node has been pruned, the result follows. 
\end{proof}

Define $V_{\mathrm b}= V\setminus (V_{\mathrm a}\cup V^{(t^*)})$.
In order to bound $|\overline{W}^{(t^*)}|$, we shall control the edges between $V$ and $\overline{W}^{(t^*)}$ and rescale this by the degree of the $w\in \overline{W}^{(t^*)}$.
For each node $v\in V_{\mathrm b}$, we denote by $d_{v,i}$ the $i$-th smallest degree among $\cN_G(v)$. Then, we have 
\begin{align*}
|\overline{W}^{(t^*)}|\geq |V_{\mathrm a}|\frac{m+1}{2}
+ \sum_{v\in V_{\mathrm b}} \sum_{i=1}^m \frac{1}{d_{v,i}}     + \sum_{v\in V^{(t^*)}, w\in \overline{W}^{(t^*)}}  \frac{\mathbf{1}\{w\sim_G v\}}{\mathrm{deg}_{G}(w)}
\end{align*}

Then, it follows from~\eqref{eq:lower_W^*i} that $\overline{W}= \overline{W}^{(t^*)}\cup_{i=1}^{c}\overline{W}^{*(i)}$
satisfies 

\begin{align}\label{eq:lower_W_bis}
|\overline{W}|\geq \sum_{v\in V_{\mathrm a}} \frac{m+1}{2} + \sum_{v\in V_{\mathrm b}} \sum_{i=1}^m  \frac{1}{d_{v,i}} + \sum_{v\in V^{(t^*)}} \left(\frac{m+1/2}{2}+  \sum_{w\in \overline{W}^{(t^*)}}  \frac{\mathbf{1}\{w\sim_G v\}}{\mathrm{deg}_{G}(w)} \right) - \frac{c}{2}
\end{align}

\begin{lemma}\label{lem:degree_w_large}
Consider a node $v$ such that $\mathrm{deg}_G(v)\geq 2m+2$. If $d_{v,m+1}\geq 4$, then we have
\begin{align*}
\sum_{i=1}^m  \frac{1}{d_{v,i}} \geq \frac{1}{2}(m+1/2) + \sum_{w\in \cN_{G}(v)}\left(\frac{1}{2\mathrm{deg}_G(w)}- \frac{1}{4} \right)
\end{align*}
\end{lemma}

\begin{proof}[Proof of Lemma~\ref{lem:degree_w_large}]
Since the $d_{v,i}$'s are increasing in $i$, we have 
\[
\sum_{i=1}^m \frac{1}{d_{v,i}}\geq \sum_{i=1}^{2m}\frac{1}{2d_{v,i}}= \frac{m}{2} + \sum_{i=1}^{2m}\left(\frac{1}{2d_{v,i}}- \frac{1}{4}\right)\enspace .
\]
Besides,  we have $d_{v,2m+2}\geq d_{v,2m+1}\geq 4$, which implies that $\frac{1}{2d_{v,2m+1}}+\frac{1}{2d_{v,2m+2}}\leq 1/4$. This allows us to conclude that
\[
\sum_{i=1}^m \frac{1}{d_{v,i}}\geq \frac{m+1/2}{2} + \sum_{i=1}^{2m+2}\left(\frac{1}{2d_{v,i}}- \frac{1}{4}\right)\geq \frac{1}{2}(m+1/2) + \sum_{w\in \cN_{G}(v)}\left(\frac{1}{2\mathrm{deg}_G(w)}- \frac{1}{4} \right)\enspace , 
\]
where we used, for the last inequality, that $d_{v,i}\geq 4$ for $i\geq m+1$.

\end{proof}

Let us define $\eta$ by 
\begin{align}\label{eq:definition_eta}
\eta:= \sum_{v\in V^{(t^*)}}
 \sum_{w\in \cN_{G}(v)}\left(\frac{1}{4}-  \frac{1}{2\mathrm{deg}_G(w)} + \frac{\mathbf{1}\{w\in \overline{W}^{(t^*)}\}}{\mathrm{deg}_{G}(w)} \right)
 - \frac{c}{2}\enspace . 
\end{align}
Let us apply Lemma~\ref{lem:degree_w_large} to any $v\in V_{\mathrm b}$. 
Since $1/(2\mathrm{deg}_G(w))\leq 1/4$, it follows from~\eqref{eq:lower_W_bis} that
\begin{align*}
    |\overline{W}|\geq \eta + \sum_{v\in V} \left(\frac{1}{2}(m+1/2) + \sum_{w\in \cN_{G}(v)}\left(\frac{1}{2\mathrm{deg}_G(w)}- \frac{1}{4} \right)\right)= \eta + \frac{|V|}{2}(m+1/2)+ \frac{|W|}{2} - \frac{|E|}{4}
\end{align*}
In light of~\eqref{eq:ojective_W_bar}, we only have to establish $\eta\geq -1/2$ to conclude the proof. 

Let us partition $V^{(t^*)}$ into its connected components $V^{(t^*)}_i$ for $i=1,\ldots, c$. Define $\eta_i$ by 
\[\eta_i:= \sum_{v\in V_i^{(t^*)}}
 \sum_{w\in \cN_{G}(v)}\left(\frac{1}{4}-  \frac{1}{2\mathrm{deg}_G(w)} + \frac{\mathbf{1}\{w\in \overline{W}^{(t^*)}\}}{\mathrm{deg}_{G}(w)} \right)
 - \frac{1}{2}\enspace ,
 \]
so that $\eta= \sum_{i=1}^c \eta_i$. We shall prove that each $\eta_i$ is non-negative, which in turn will imply that $\eta\geq 0$.   
We consider two cases.~\\ 
{\bf Case (i)}: There is a degree-$2$ node $w\in W$ that connects $V_i^{(t^*)}$ and $V_{\mathrm a}\cup V_{\mathrm b}$. By construction, this node belongs to $\overline{W}^{(t^*)}$ and we get $\eta_i\geq 1/2 -1/2 = 0$.

\medskip 

\noindent 
{\bf Case (ii)}: No degree-$2$ node $w\in W$ is simultaneously connected to $V_i^{(t^*)}$ and $V_{\mathrm a}\cup V_{\mathrm b}$. By condition ($\cW_G$), we know that 
\[
|\oplus_{v\in V_i^{(t^*)}}\cN_G(v)|\notin [1:2m+1]\ . 
\]
If $|\oplus_{v\in V_i^{(t^*)}}\cN_G(v)|\geq 2m+2$, this implies that there are at least $2m+2$ edges between a node in $V_i^{(t^*)}$ and a node $w\in \oplus_{v\in V_i^{(t^*)}}\cN_G(v)$. All these nodes $w$ are of degree at least $4$ 
 and we get
\[
\eta_i \geq (2m+2) \frac{1}{8} - 1/2\geq 0 \  , 
\]
since $m\geq 1$. Finally, we consider the case where $\oplus_{v\in V_i^{(t^*)}}\cN_G(v)=\emptyset$. Since $G$ is connected, this implies that there is a positive and even number $l$ of edges between $V_i^{(t^*)}$ and nodes $w\in W$ of degree at least $4$. If $l\geq 4$, we conclude as before that 
\[
\eta_i \geq l/8 - 1/2 \geq 0\enspace . 
\]
Finally, we observe that the case $l=2$ is not possible, as this would imply that $V_i^{(t^*)}$ is a group of type A, while we assumed at the beginning that $G$ does not contain any group of type A.

\end{proof}

\begin{proof}[Proof of Lemma~\ref{lem:deg:generat:type_A}]
We only prove the result for connected graphs $G$. We do it by induction on the size $|V|$. First, the property holds for any such graph $G=(V,W,E)$ with $|V|=2$. Indeed, the degree $\mathrm{deg}_G(w)$ of any $w\in W$ is equal to $2$, and the result follows from Lemma~\ref{lem:deg2}.

Now, assume that the property holds for any graph $G=(V,W,E)$ satisfying ($\cW_G$), containing no isolated node, and such that $|V|\leq r$. Consider a graph $G=(V,W,E)$ satisfying ($\cW_G$), containing no isolated node, and such that $|V|=r+1$. If $G$ does not contain any group of type A, the result follows from Lemma~\ref{lem:deg_general}. Otherwise, let $V^*$ be a group of type $A$ and let $w^*\in W$ be the unique node that connects $V^*$ to $V^{\circ}:=V\setminus V^*$. 

Let us remove the two edges $e^*_1$ and $e^*_2$ between $V^*$ and $w^*$. The graph $G'=(V,W,E\setminus \{e^*_1,e^*_2\})$ is therefore made of two connected components that respectively contain $V^{\circ}$ and $V^*$. Let us denote by $G^{\circ}=(V^{\circ},W^{\circ},E^{\circ})$ and $G^*=(V^*,W^*, E^*)$ the two connected components of $G'$ containing $V^{\circ}$ and $V^*$ respectively; in particular, $w^*$, which keeps at least two neighbors in $V^{\circ}$, belongs to $W^{\circ}$ and not to $W^*$. One readily checks that $G^{\circ}$ is connected and satisfies ($\cW_G$), is such that $|V^{\circ}|\leq r$ and does not contain any isolated node. By the induction hypothesis, there exists $\overline{W}^{\circ}\subset W^{\circ}$ that satisfies the $G^{\circ}$-enumeration property and such that 
\[
|\overline{W}^{\circ}|\geq \frac{|W^{\circ}|}{2} - \frac{|E^{\circ}|}{4} + \frac{|V^{\circ}|}{2} (m+1/2) - \frac{1}{2}\enspace .
\]
Similarly, $G^*$ satisfies the conditions of Lemma~\ref{lem:deg2}. Hence, 
there exists $\overline{W}^*\subset W^*$ that satisfies the $G^*$-enumeration property and such that 
\[
|\overline{W}^*|\geq  \frac{|V^*|}{2} (m+1/2) - \frac{1}{2}= \frac{|W^*|}{2} - \frac{|E^*|}{4} + \frac{|V^*|}{2} (m+1/2) - \frac{1}{2}\enspace ,
\]
since all the nodes $w\in W^*$ are of degree $2$.

Since no node $v\in V^{\circ}$ is adjacent to $w\in W^*$, $\overline{W}=\overline{W}^{\circ} \cup \overline{W}^*$ satisfies the $G$-enumeration property. More precisely, let us order the nodes of $V$ by first enumerating those of $V^{\circ}$ according to an ordering witnessing the $G^{\circ}$-enumeration property of $\overline{W}^{\circ}$, and then those of $V^*$ according to an ordering witnessing the $G^*$-enumeration property of $\overline{W}^*$. During the first phase, the nodes of $\overline{W}^*$ are never counted in the left-hand side of~\eqref{eq:definition:enumeration:property}, as they are not adjacent to $V^{\circ}$, and $\cN_G(v)=\cN_{G^{\circ}}(v)$ for any $v\in V^{\circ}$, so that~\eqref{eq:definition:enumeration:property} holds along this phase as it does in $G^{\circ}$. Conversely, the only node of $\overline{W}^{\circ}$ that may be adjacent to $V^*$ is $w^*$: any other node $w\in W^{\circ}$ is adjacent to $V^{\circ}$, so that, if it were also adjacent to $V^*$, it would still connect $V^*$ to $V\setminus V^*$ after the removal of $w^*$, contradicting the property (iii) of a group of type $A$. Since $w^*$ is adjacent to at least one node of $V^{\circ}$ --~its degree in $G^{\circ}$ is at least $2$, as recalled above~--, it has been removed from the sets $\overline{W}^{(t)}$ before the end of the first phase. Hence, along the second phase, only the nodes of $\overline{W}^*$ are counted in the left-hand side of~\eqref{eq:definition:enumeration:property}, which therefore holds as it does in $G^*$. We conclude that
\[
|\overline{W}|\geq \frac{|W^*|+|W^{\circ}|}{2} - \frac{|E^*|+|E^{\circ}|}{4}+ \frac{|V^*|+|V^{\circ}|}{2}(m+1/2)-1 =  \frac{|W|}{2} - \frac{|E|}{4} + \frac{|V|}{2} (m+1/2) - \frac{1}{2}\enspace ,
\]
since $|E|= 2 + |E^*|+ |E^{\circ}|$. We have proved the induction property and the result follows.

\medskip

The extension to graphs with multiple connected components is straightforward as we only need to consider each connected component independently.

\end{proof}

\subsection{Optimality of the exponent in Lemma~\ref{lem:eqcounting}}\label{sec:optimality:b}

We prove the claim of the remark following Lemma~\ref{lem:eqcounting}. Fix $r\geq 1$, let $G^{\circ}$ be the cycle of $2r$ pairs of twin nodes defined there, and let $d\to\infty$ with $m$ and $r$ fixed. The upper bound follows from Lemma~\ref{lem:eqcounting} applied with $D=4r(m+1)$, as soon as $d\geq 2^8D^4$. For the lower bound, we use a direct counting argument, which does not rely on Lang--Weil type estimates.

\medskip
\noindent
\emph{Syndromes.} Write $\mathbb F_{d+1}^*:=\mathbb F_{d+1}\setminus\{0\}$ and, for a finite subset $A\subset \mathbb F_{d+1}^*$, define its syndrome $s(A):=\big(\sum_{a\in A}a^{2l-1}\big)_{l\leq m}\in\mathbb F_{d+1}^m$. Since $\mathbb F_{d+1}$ has characteristic $2$, we have $s(A\oplus B)=s(A)+s(B)$, where $A\oplus B$ is the symmetric difference of $A$ and $B$. Besides, if $1\leq |A|\leq 2m$, then $s(A)\neq 0$: otherwise, Lemma~\ref{lem:fundBCH}, read in $\mathbb F_{d+1}$ through $\psi$ as in the proof of Lemma~\ref{lem:eqcounting}, would imply that the vector made of the pairwise distinct elements of $A$ is paired, which is impossible. As a consequence:
\begin{itemize}
\item[(i)] if $|A|=|B|=m$ and $s(A)=s(B)$, then $A=B$, since $|A\oplus B|\leq 2m$ and $s(A\oplus B)=0$;
\item[(ii)] if $|A|=|B|=m+1$, $A\neq B$ and $s(A)=s(B)$, then $A\cap B=\emptyset$, since otherwise $1\leq |A\oplus B|\leq 2m$. In particular, $A\cup B$ is then a subset of size $2m+2$ with $s(A\cup B)=0$.
\end{itemize}

\medskip
\noindent
\emph{Counting.} Since $\cN(v_t)=\cN(v'_t)$, a tuple $(x_w)_{w\in W^{\circ}}$ whose values on each neighborhood $\cN(v_t)$ are pairwise distinct satisfies the system~\eqref{eq:system} for $G^{\circ}$ if and only if it satisfies the $2rm$ equations
\begin{equation}\label{eq:optimality:blocks}
s\big(\{x_{w_{t,1}},\ldots,x_{w_{t,2m}}\}\big)=s\big(\{x_{u_{t-1}},x_{u_t}\}\big)\enspace ,\qquad t\in\mathbb Z/2r\mathbb Z\enspace .
\end{equation}
For $a,a'\in\mathbb F_{d+1}^*$ with $a\neq a'$, let $M_{a,a'}$ be the number of injective tuples $(y_1,\ldots,y_{2m})$ with values in $\mathbb F_{d+1}^*\setminus\{a,a'\}$ such that $s(\{y_1,\ldots,y_{2m}\})=s(\{a,a'\})$, and set $M_{a,a}=0$. The $d\times d$ matrix $M$ is symmetric with nonnegative entries. Expanding the trace,
$$\mathrm{Tr}(M^{2r})=\sum_{(a_t)_{t\in\mathbb Z/2r\mathbb Z}}\ \prod_{t\in\mathbb Z/2r\mathbb Z}M_{a_{t-1},a_t}$$
is the number of tuples $(x_w)_{w\in W^{\circ}}$ with values in $\mathbb F_{d+1}^*$ satisfying~\eqref{eq:optimality:blocks} and such that, for every $t$, the $2m+2$ values $x_w$, $w\in\cN(v_t)$, are pairwise distinct. Such a tuple belongs to $\mathbb U$, unless two nodes $w,w'\in W^{\circ}$ without common neighbor receive the same value. For a given pair $\{w,w'\}$ of such nodes, we bound the number of these tuples with $x_w=x_{w'}$ as follows. We first choose the values of the nodes $u_t$, and then, for each $t$, the values of $m$ nodes among $w_{t,1},\ldots,w_{t,2m}$, including $w$ or $w'$ if one of them belongs to this block. By~(i) and~\eqref{eq:optimality:blocks}, the set of values of the $m$ remaining nodes of each block is then uniquely determined, so that there are at most $m!$ choices for these values. As $x_w=x_{w'}$ removes one of the $2r(m+1)$ free choices, the number of such tuples is at most $(m!)^{2r}d^{2r(m+1)-1}$. Summing over the at most $|W^{\circ}|^2$ pairs $\{w,w'\}$, we get
$$|\mathbb U|\geq \mathrm{Tr}(M^{2r})-|W^{\circ}|^2(m!)^{2r}d^{2r(m+1)-1}\enspace .$$
Since $M$ is symmetric, its eigenvalues are real and, as $2r$ is even, $\mathrm{Tr}(M^{2r})\geq\lambda_{\max}(M)^{2r}\geq (\mathbf 1^\top M\mathbf 1/d)^{2r}$, where $\mathbf 1$ is the all-ones vector. The quantity $\mathbf 1^\top M\mathbf 1$ is the number of injective $(2m+2)$-tuples with values in $\mathbb F_{d+1}^*$ whose set of values has a zero syndrome, that is $(2m+2)!\,Z$, where $Z$ is the number of subsets of $\mathbb F_{d+1}^*$ of size $2m+2$ with a zero syndrome. For $\varsigma\in\mathbb F_{d+1}^m$, let $n_{\varsigma}$ be the number of subsets of $\mathbb F_{d+1}^*$ of size $m+1$ with syndrome $\varsigma$. By~(ii), any ordered pair $(A,B)$ of distinct such subsets with the same syndrome gives rise to the subset $A\cup B$, which is counted in $Z$, and each subset counted in $Z$ arises from at most $\binom{2m+2}{m+1}$ such pairs. Hence, by the Cauchy--Schwarz inequality and since there are $(d+1)^m$ syndromes,
$$\binom{2m+2}{m+1}Z\geq \sum_{\varsigma}n_{\varsigma}(n_{\varsigma}-1)\geq \frac{1}{(d+1)^m}\binom{d}{m+1}^2-\binom{d}{m+1}\enspace .$$
Since $(2m+2)!/\binom{2m+2}{m+1}=((m+1)!)^2$, we obtain $\mathbf 1^\top M\mathbf 1\geq (1-o(1))d^{m+2}$, so that $\mathrm{Tr}(M^{2r})\geq (1-o(1))d^{2r(m+1)}$ and $|\mathbb U|\geq (1-o(1))d^{2r(m+1)}$. Finally, by~\eqref{eq:proba:A[G]} and since $d!/(d-|W^{\circ}|)!\leq d^{|W^{\circ}|}$,
$$\mathbb{P}_{\sigma}[\cA_{G^{\circ}}]\geq \frac{|\mathbb U|}{d^{|W^{\circ}|}}\geq (1-o(1))\,d^{2r(m+1)-2r(2m+1)}=(1-o(1))\Big(\frac{1}{d^{m}}\Big)^{2r}\enspace .$$

\medskip
\noindent
\emph{Rank of the Jacobian.} For any $x\in\mathbb U$, the rows of the Jacobian matrix $\cJ_x$ defined in~\eqref{eq:definition:Jacobian} that are indexed by $(i,v_t)$ and $(i,v'_t)$ coincide, so that $\mathrm{rank}(\cJ_x)\leq 2rm$. Together with Lemma~\ref{lem:rank_Jacobian}, which gives $\mathrm{rank}(\cJ_x)\geq b(G^{\circ})=2rm-1/2$, we get $\mathrm{rank}(\cJ_x)=\lceil b(G^{\circ})\rceil$, so that the lower bound on the rank used in the proof of Lemma~\ref{lem:eqcounting} is attained for $G^{\circ}$.

\section{Proof of Theorem~\ref{thm:isorefo2}}\label{sec:proof:thm:isorefo2}

We define the Gram matrix $\Gamma$ associated to the family $(1, (\tilde{\Psi}_G)_{G\in \cG^{\mathrm{even}}_{\leq D}})$
by $\Gamma_{G^{(1)},G^{(2)}}=\mathbb E\left[\tilde \Psi_{G^{(1)}} \tilde \Psi_{G^{(2)}}\right]$ for any $G^{(1)}$ and $G^{(2)}$ in $\cG^{\mathrm{even}}_{\leq D}$, $\Gamma_{\emptyset,G}= \Gamma_{G, \emptyset}= \mathbb E[\tilde \Psi_G]$ for any $G\in \cG^{\mathrm{even}}_{\leq D}$, and $\Gamma_{\emptyset,\emptyset}= 1$.
Let us write $\mathrm{Id}$ for the identity matrix with the same dimension as $\Gamma$.
Then, Theorem~\ref{thm:isorefo2} is a straightforward corollary of the following result. 

\begin{proposition}\label{prop:ortho}
Assume that Condition~\eqref{eq:signalcond} holds for some  $c_{\mathrm{s}} \geq 10$, that  $2^8D^4\leq d$, that $D\geq2$, and that $n\geq d$. Then, we have
    \[
\|\Gamma - \mathrm{Id}\|_{op}\leq 4 D^{-c_{\mathrm{s}}/2}+  \frac{4D^2}{d} \enspace ,
\]
where $\|.\|_{op}$ stands for the operator norm. 
\end{proposition}

In order to establish Proposition~\ref{prop:ortho}, we shall simply bound the operator norm of a symmetric matrix by the largest $l_1$ norm of its rows. For that purpose, we need to evaluate $|\Gamma_{G^{(1)},G^{(2)}}- \mathbf{1}\{G^{(1)}=G^{(2)}\}|$. This is done in Proposition~\ref{prop:scalprod} below. This requires controlling the sum of moments of the form $\mathbb{E}[\Psi_{G^{(1)},\pi^{(1)}}\Psi_{G^{(2)},\pi^{(2)}}]$ over all possible labelings $\pi^{(1)}$ and $\pi^{(2)}$. For that purpose, we need to introduce additional graph notation.

\subsection{Additional graph notation}\label{sec:additional:graph:notations}

\medskip 

\noindent 
\paragraph{Node matching.} Consider two graphs $G^{(1)}=(V^{(1)},W^{(1)},E^{(1)})$, $G^{(2)}=(V^{(2)},W^{(2)},E^{(2)})$. 
Given labelings $\pi^{(1)}=(\pi^{(1)}_1,\pi^{(1)}_2)$ and $\pi^{(2)}=(\pi^{(2)}_1,\pi^{(2)}_2)$, recall that we say that two nodes $v^{(1)}$ and $v^{(2)}$ (resp. $w^{(1)}$ and $w^{(2)}$) are matched if $\pi_1^{(1)}(v^{(1)})=  \pi_1^{(2)}(v^{(2)})$ (resp. $\pi_2^{(1)}(w^{(1)})=  \pi_2^{(2)}(w^{(2)})$). 
More generally,  a matching $\mathbf M_1$ stands for a set of pairs of nodes $(v^{(1)},v^{(2)})\in V^{(1)}\times V^{(2)}$ where no node in $V^{(1)}$ or $V^{(2)}$ appears twice. Similarly, we define matchings $\mathbf M_2$ for $W^{(1)}$ and $W^{(2)}$.  We denote by $\mathcal M_1$ and  $\mathcal{M}_2$ the collections of all possible node matchings, where, for $\mathcal M_1$, we have the additional constraint that $(v_1^{(1)},v_1^{(2)})$ and $(v_2^{(1)},v_2^{(2)})$ belong to the matching - and we write $\mathcal M = \mathcal M_1\times \mathcal M_2$. 
Given $\mathbf M_1 \in \mathcal M_1$ and $\mathbf M_2 \in \mathcal{M}_2$, we define the set of labelings compatible with the matchings $\mathbf M_1$ and $\mathbf M_2$ by 
\begin{align*}
&\bar \Pi_1(\mathbf M_1)\\ 
&= \left\{(\pi^{(1)}_1,\pi^{(2)}_1)\in \bar \Pi_{V^{(1)}}\times \bar \Pi_{V^{(2)}}: \forall (v^{(1)},v^{(2)}) \in V^{(1)}\times V^{(2)}, \{\pi^{(1)}_1(v^{(1)}) = \pi^{(2)}_1(v^{(2)})\}\Longleftrightarrow \{(v^{(1)},v^{(2)}) \in \mathbf M_1\}\right\}\enspace ;  \\
&\Pi_2(\mathbf M_2)\\
&= \left\{(\pi^{(1)}_2,\pi^{(2)}_2)\in \Pi_{W^{(1)}}\times\Pi_{W^{(2)}}: \forall (w^{(1)},w^{(2)}) \in W^{(1)}\times W^{(2)}, \{\pi^{(1)}_2(w^{(1)}) = \pi^{(2)}_2(w^{(2)})\}\Longleftrightarrow \{(w^{(1)},w^{(2)}) \in \mathbf M_2\}\right\}\enspace .
\end{align*}
For short, we write $\mathbf M = (\mathbf{M}_1,\mathbf M_2) \in \mathcal M$ and $\Pi(\mathbf M)$ for the collection of labelings $(\pi^{(1)}_1,\pi^{(1)}_2,\pi^{(2)}_1,\pi^{(2)}_2)$ compatible with $\mathbf M$.
Given a matching $\mathbf M$, we say that two edges $e\in E^{(1)}$ and $e'\in E^{(2)}$ are matched if the corresponding
incident nodes are matched. Importantly, as $\mathbb{P}$ is permutation invariant, $\mathbb{E}[P_{G^{(1)},\pi_1^{(1)},\pi_2^{(1)}}P_{G^{(2)},\pi_1^{(2)},\pi_2^{(2)}}]$ is the same for all $(\pi^{(1)},\pi^{(2)})$ in $\Pi(\mathbf{M})$.

Note that, for a fixed matching $\mathbf{M}$, all labeled  graphs $G_{\cup}$ (resp. $G_{\cap}$, $G_{\oplus}$) are isomorphic for $(\pi_1^{(1)},\pi_1^{(2)})\in \bar \Pi_1(\mathbf M_1)$ and $(\pi_2^{(1)},\pi_2^{(2)})\in \Pi_2(\mathbf M_2)$. Hence, we shall refer to quantities such as $|E_{\oplus}|$, $|V_{\oplus}|$, $|W_{\oplus}|$\ldots associated to a matching $\mathbf{M}$.\\
We also have $|E_{\cup}| = | E^{(1)}| +|E^{(2)}| - |E_\cap|$ and $|V_{\cup}| = |V^{(1)}|+|V^{(2)}|-|\mathbf{M}_1|$ for $(\pi_1^{(1)},\pi_1^{(2)})\in \bar \Pi_1(\mathbf M_1)$.

\medskip

\noindent
\paragraph{Sets of unmatched nodes and of semi-matched nodes.} 
 Write $U^{(1)}=(U^{(1)}_1,U^{(1)}_2)$, resp.~$U^{(2)}=(U^{(2)}_1,U^{(2)}_2)$ for the set of nodes in $(\pi^{(1)}_1,\pi^{(1)}_2)[G^{(1)}]$
 resp. $(\pi^{(2)}_1,\pi^{(2)}_2)[G^{(2)}]$  that are not matched, namely the {\bf unmatched nodes}, that is 
\begin{align*}
U^{(1)}_1 =\pi_1^{(1)}(V^{(1)})\setminus \pi_1^{(2)}(V^{(2)})\ ; \quad \quad\quad 
U^{(2)}_1 =\pi_1^{(2)}(V^{(2)})\setminus \pi_1^{(1)}(V^{(1)})\  ; \\ 
U^{(1)}_2 =\pi_2^{(1)}(W^{(1)})\setminus \pi_2^{(2)}(W^{(2)})\ ; \quad \quad\quad 
U^{(2)}_2 =\pi_2^{(2)}(W^{(2)})\setminus \pi_2^{(1)}(W^{(1)})\  ; 
\end{align*}
Here, $V^{(1)}$ and $V^{(2)}$ include the possibly isolated nodes $v^{(i)}_1$ and $v^{(i)}_2$, which are always matched since $\mathbf M_1\in\mathcal M_1$. As the other nodes of $G^{(1)}$ and $G^{(2)}$ are not isolated, $U^{(1)}_1$ and $U^{(2)}_1$ only contain non-isolated nodes. Again, $|U^{(1)}_1|$, $|U^{(2)}_1|$, $|U^{(1)}_2|$, and $|U^{(2)}_2|$ only  depend on $(\pi^{(1)},\pi^{(2)})$ through the matching $\mathbf{M}$. 
We have, for $i\in \{1,2\}$,
\begin{align}\label{eq:unmatched}
    |V^{(i)}| = |\mathbf M_1| + |U_1^{(i)}|\, \quad \quad |W^{(i)}| = |\mathbf M_2| + |U_2^{(i)}|\enspace . 
\end{align}
Write also
$$\mathbf M_{\mathrm{SM}}(\mathbf M_1) \subset \mathbf M_1,~~~~~~\mathbf M_{\mathrm{SM}}(\mathbf M_2) \subset \mathbf M_2,~~~~~~\mathrm{and}~~~~~\mathbf M_{\mathrm{SM}} := \mathbf M_{\mathrm{SM}}(\mathbf M) = (\mathbf M_{\mathrm{SM}}(\mathbf M_1),\mathbf M_{\mathrm{SM}}(\mathbf M_2)) $$
for the set of node matches of $(G^{(1)},G^{(2)})$ that are matched, and yet that are not pruned when creating the symmetric difference graph $G_{\oplus}$.
This is the set of {\bf semi-matched nodes} - i.e.~at least one of their incident edges is not matched.  The remaining pairs of nodes $\mathbf M \setminus \mathbf M_{\mathrm{SM}}$ are said to be {\bf perfectly matched} as all the edges incident to them are matched. We write  $\mathbf M_{\mathrm{PM}} = \mathbf M_{\mathrm{PM}}(\mathbf M)$ for the set of perfect matches in $\mathbf M$. Note that 
\begin{equation}\label{eq:PMSM}
     |V^{(1)}| +|V^{(2)}| =   |V_{\oplus}| + |(\mathbf M_{\mathrm{SM}})_1| + 2|(\mathbf M_{\mathrm{PM}})_1|\enspace  .
\end{equation}

\medskip

\paragraph{Edit Distance between graphs.} For any two graphs $G^{(1)}$ and $G^{(2)}$, we define  the so-called edit-distance.
\begin{equation}\label{eq:definition:edit:distance}
d(G^{(1)}, G^{(2)}) := \min_{\mathbf M \in \mathcal M} |E_{\oplus}| \enspace .
\end{equation}
Note that $d(G^{(1)}, G^{(2)})=0$ if and only if $G^{(1)}$ and $G^{(2)}$ are equivalent. As a consequence, if $G^{(1)}$ and $G^{(2)}$ are in $\mathcal{G}_{\leq D}$, the edit distance is equal to $0$ if and only if $G^{(1)}=G^{(2)}$.

\subsection{Entry-wise control of $\Gamma$}

\begin{proposition}\label{prop:scalprod}
Assume that Condition~\eqref{eq:signalcond} holds for some $c_{\mathrm{s}} \geq 10$, that $2^8D^4\leq d$, that $D\geq 2$, and that $n\geq d$.
    Consider two graphs $G^{(1)},G^{(2)} \in \mathcal{G}^{\mathrm{even}}_{\leq D}$. We have% are two graphs:
    \begin{equation*}
     |\Gamma_{G^{(1)}, G^{(2)}} - \mathbf 1\{G^{(1)} = G^{(2)}\}|=    \left|\mathbb E\left[\tilde \Psi_{G^{(1)}} \tilde \Psi_{G^{(2)}}\right] - \mathbf 1\{G^{(1)} = G^{(2)}\}\right|\leq 2D^{-c_{\mathrm{s}} (d (G^{(1)}, G^{(2)})\lor 1)}+  \frac{4D^2}{d} \mathbf 1\{G^{(1)} = G^{(2)}\}\enspace , 
    \end{equation*}
    where we recall that $d(.,.)$ is the edit distance. 
    The same bound holds for $G^{(2)}=\emptyset$ with the convention  $d (G^{(1)}, G^{(2)})= |E^{(1)}|$.  
\end{proposition}

In order to establish Proposition~\ref{prop:ortho} in a self-contained way, we first bound, for a fixed graph, the number of graphs lying at a given edit distance from it.

\begin{lemma}\label{lem:count:distance}
Fix $G^{(1)}\in\mathcal{G}_{\leq D}$ and a positive integer $\zeta$. The number of graphs $G^{(2)}\in\mathcal{G}_{\leq D}$ such that $d(G^{(1)},G^{(2)})=\zeta$ is at most $4\cdot 36^\zeta D^{3\zeta+1}$.
\end{lemma}

\begin{proof}[Proof of Lemma~\ref{lem:count:distance}]
Since $G^{(1)}$ has no isolated node outside $\{v_1,v_2\}$, we have $|E^{(1)}|\leq D$, $|V^{(1)}|\leq D+2$, and $|W^{(1)}|\leq D$. Comparing $G^{(1)}$ and $G^{(2)}$ through the fully unmatched matching shows $d(G^{(1)},G^{(2)})\leq |E^{(1)}|+|E^{(2)}|\leq 2D$, so we only need to consider $\zeta\leq 2D$.

Fix a matching $\mathbf{M}$ achieving $d(G^{(1)},G^{(2)})=|E_{\oplus}|=\zeta$, and write $a:=|E^{(1)}\cap E_{\oplus}|\leq \zeta$ for the number of edges of $G^{(1)}$ that do not survive into $G^{(2)}$, and $b:=\zeta-a=|E^{(2)}\cap E_{\oplus}|$ for the number of edges of $G^{(2)}$ that are not already in $G^{(1)}$. The graph $G^{(2)}$ (up to the isomorphism defining $\mathcal{G}_{\leq D}$) is entirely determined by (I) which $a$ edges of $E^{(1)}$ are dropped, and (II) how the $b$ remaining edges of $E_{\oplus}$ are attached.

(I) There are at most $\binom{|E^{(1)}|}{a}\leq D^{a}\leq D^{\zeta}$ choices for a given $a\leq \zeta$, hence at most $(\zeta+1)D^{\zeta}$ choices summed over $a=0,\ldots,\zeta$. Since $\zeta\leq 2D$, we have $\zeta+1\leq 3D$, so this contributes a factor at most $3D^{\zeta+1}$.

(II) By the no-isolated-node convention on $G_{\oplus}$, the $b\leq \zeta$ new edges touch at most $2b\leq 2\zeta$ nodes that are not among the $\leq D+2$ (resp.~$\leq D$) original nodes of $V^{(1)}$ (resp.~$W^{(1)}$). Hence each new edge has at most $(D+2)+2\zeta\leq 6D$ possible choices for its endpoint in $V$, and at most $D+2\zeta\leq 6D$ possible choices for its endpoint in $W$ (using $\zeta\leq 2D$). Bounding the number of possible edge \emph{sets} by the (larger) number of ordered edge \emph{sequences}, there are at most $(6D\times 6D)^{b}=(36D^2)^{b}\leq (36D^2)^{\zeta}$ ways to place these $b$ edges.

Multiplying (I) and (II), the number of possible $G^{(2)}$ is at most $3D^{\zeta+1}\cdot (36D^2)^{\zeta}=3\cdot 36^{\zeta}D^{3\zeta+1}\leq 4\cdot 36^{\zeta}D^{3\zeta+1}$.
\end{proof}

\begin{proof}[Proof of Proposition~\ref{prop:ortho}]
As explained earlier, we simply use
\begin{equation}\label{eq:upper_op_l1}
\|\Gamma - \mathrm{Id}\|_{op}\leq \max\Bigg[\max_{G^{(1)}\in \mathcal{G}^{\mathrm{even}}_{\leq D}}\Big[|\Gamma_{G^{(1)},\emptyset}| +  \sum_{G^{(2)}\in \mathcal{G}^{\mathrm{even}}_{\leq D}}     |\Gamma_{G^{(1)}, G^{(2)}} - \mathbf 1\{G^{(1)} = G^{(2)}\}|\Big],  \sum_{G^{(2)}\in \mathcal{G}^{\mathrm{even}}_{\leq D}} |\Gamma_{G^{(2)},\emptyset}| \Bigg]
\end{equation}

\paragraph{Off-diagonal terms.} Fix $G^{(1)}\in\mathcal{G}_{\leq D}^{\mathrm{even}}$. Since both $G^{(1)}$ and $G^{(2)}$ have only even-degree nodes, so does their symmetric difference graph $G_{\oplus}$ associated to any matching $\mathbf{M}$ -- indeed, for any node $v$, $\deg_{G_{\oplus}}(v)=\deg_{G^{(1)}}(v)+\deg_{G^{(2)}}(v)-2\deg_{\mathrm{shared}}(v)$ (or simply $\deg_{G^{(1)}}(v)$, resp.~$\deg_{G^{(2)}}(v)$, if $v$ is unmatched), which is even whenever $\deg_{G^{(1)}}(v)$ and $\deg_{G^{(2)}}(v)$ are. Since $G_{\oplus}$ is bipartite, $|E_{\oplus}|$ equals the sum of the degrees on its $V_{\oplus}$-side, hence is itself even. Moreover, if $G^{(1)}\neq G^{(2)}$, then for the matching achieving $d(G^{(1)},G^{(2)})=|E_{\oplus}|$, the graph $G_{\oplus}$ is non-empty, and (being simple, bipartite, with no isolated node, and all degrees even, hence $\geq 2$) it has at least $2$ nodes of degree $\geq 2$ on each side, so $|E_{\oplus}|\geq 4$. We conclude that $d(G^{(1)},G^{(2)})$, whenever nonzero, is an \emph{even} integer $\geq 4$.

Writing $N(\zeta):=\#\{G^{(2)}\in\mathcal{G}_{\leq D}: d(G^{(1)},G^{(2)})=\zeta\}$, Lemma~\ref{lem:count:distance} gives $N(\zeta)\leq 4\cdot 36^{\zeta}D^{3\zeta+1}$ (this bound applies a fortiori to $\mathcal{G}^{\mathrm{even}}_{\leq D}\subset \mathcal{G}_{\leq D}$), so that, using Proposition~\ref{prop:scalprod} and the previous paragraph,
\[
\sum_{G^{(2)}\in \mathcal{G}^{\mathrm{even}}_{\leq D}, G^{(2)}\neq G^{(1)}} |\Gamma_{G^{(1)},G^{(2)}}| \leq \sum_{\substack{\zeta\geq 4\\ \zeta \text{ even}}} N(\zeta)\cdot 2D^{-c_{\mathrm{s}}\zeta} \leq  \sum_{\substack{\zeta\geq 4\\ \zeta \text{ even}}} 8\cdot 36^{\zeta}D^{3\zeta+1-c_{\mathrm{s}}\zeta}\enspace .
\]
Writing $\theta:=36D^{3-c_{\mathrm{s}}}$, the right-hand side equals $8D\sum_{k\geq 2}\theta^{2k}=8D\,\theta^{4}/(1-\theta^{2})$ as soon as $\theta<1$. 

\paragraph{Diagonal and $\emptyset$-terms.} By Proposition~\ref{prop:scalprod} applied with $G^{(2)}=G^{(1)}$, for which $d(G^{(1)},G^{(1)})=0$ so that $d(G^{(1)},G^{(1)})\vee 1=1$, we get $|\Gamma_{G^{(1)},G^{(1)}}-1|\leq 2D^{-c_{\mathrm{s}}} + \frac{4D^2}{d}$. Finally, using~\eqref{eq:definition:PsiG} together with the bound $\big|\mathbb{E}[\tilde\Psi_{G}]\big|\leq 2D^{-c_{\mathrm{s}}|E|}$ established at the end of the proof of Proposition~\ref{prop:scalprod} below, and the same counting of graphs by number of edges as in the proof of Theorem~\ref{thm:corr} (at most $(D+2)^{2e}$ graphs in $\mathcal{G}^{\mathrm{even}}_{\leq D}$ have $e$ edges), we get, for $c_{\mathrm{s}}\geq 10$ and $D\geq 2$,
\[
\sum_{G^{(2)}\in \mathcal{G}^{\mathrm{even}}_{\leq D}} |\Gamma_{G^{(2)},\emptyset}| \leq \sum_{\substack{e\geq 2\\ e\text{ even}}} 2(D+2)^{2e}D^{-c_{\mathrm{s}}e} \leq 4D^{-c_{\mathrm{s}}/2}\enspace ,
\]
where the last inequality holds because $(D+2)^2\leq D^4$ for $D\geq 2$: writing $y:=D^{8-2c_{\mathrm{s}}}\leq 2^{8-2c_{\mathrm{s}}}\leq 2^{-12}$, the middle sum is at most $2\sum_{k\geq 1}y^{k}=2y/(1-y)\leq 4y\leq 4D^{-c_{\mathrm{s}}/2}$, since $8-2c_{\mathrm{s}}\leq -c_{\mathrm{s}}/2$ for $c_{\mathrm{s}}\geq 10$. Since $|E^{(1)}|\geq 1$, the same bound $\big|\mathbb{E}[\tilde\Psi_{G}]\big|\leq 2D^{-c_{\mathrm{s}}|E|}$ also gives the single-term bound $|\Gamma_{G^{(1)},\emptyset}|\leq 2D^{-c_{\mathrm{s}}}$ needed in the first branch of the max above.

\paragraph{Combining the three displays.} For a fixed $G^{(1)}$, the first part of the max in~\eqref{eq:upper_op_l1} is
\[
|\Gamma_{G^{(1)},\emptyset}| + |\Gamma_{G^{(1)},G^{(1)}}-1| + \sum_{G^{(2)}\neq G^{(1)}}|\Gamma_{G^{(1)},G^{(2)}}| \leq 2D^{-c_{\mathrm{s}}} + \Big(2D^{-c_{\mathrm{s}}}+\frac{4D^2}{d}\Big) + 8D\,\frac{\theta^4}{1-\theta^2} = 4D^{-c_{\mathrm{s}}}+8D\,\frac{\theta^4}{1-\theta^2}+\frac{4D^2}{d}\enspace .
\]
It therefore suffices to check that $4D^{-c_{\mathrm{s}}}+8D\,\theta^4/(1-\theta^2)\leq 4D^{-c_{\mathrm{s}}/2}$. Since $D\geq 2$ and $c_{\mathrm{s}}\geq 10$, we have $\theta\leq 36\cdot 2^{-7}<0.29$, so that $8/(1-\theta^2)\leq 9$ and
\[
8D\,\frac{\theta^4}{1-\theta^2}\leq 9\cdot 36^4\, D^{13-4c_{\mathrm{s}}}\leq 9\cdot 36^4\cdot 2^{-22}\, D^{-c_{\mathrm{s}}/2}\leq 3.7\, D^{-c_{\mathrm{s}}/2}\enspace ,
\]
where we used $D^{13-7c_{\mathrm{s}}/2}\leq 2^{13-35}$. Besides, $4D^{-c_{\mathrm{s}}}\leq 4\cdot 2^{-5}D^{-c_{\mathrm{s}}/2}=D^{-c_{\mathrm{s}}/2}/8$, and the claim follows by summing these two bounds.  The second part, $\sum_{G^{(2)}}|\Gamma_{G^{(2)},\emptyset}|\leq 4D^{-c_{\mathrm{s}}/2}$, is bounded by the same quantity. This concludes the proof.
\end{proof}

\subsection{Proof of Proposition~\ref{prop:scalprod}}

The proof of this proposition follows the same steps as the proof of Proposition A.2 in~\cite{carpentier2025rank}, with some small changes. We start from the decomposition
\begin{align*}
    &\mathbb{E}\left[\tilde{\Psi}_{G^{(1)}}\tilde{\Psi}_{G^{(2)}}\right]\\
    &= \frac{1}{\sqrt{|\mathrm{Aut}(G^{(1)})||\mathrm{Aut}(G^{(2)})|}}\sum_{\mathbf M \in \mathcal M} n^{2-(|V^{(1)}|+|V^{(2)}|)/2}d^{-(|W^{(1)}|+|W^{(2)}|)/2} \sum_{(\pi^{(1)}, \pi^{(2)}) \in \Pi(\mathbf M)} \mathbb E[\Psi_{G^{(1)}, \pi^{(1)}}\Psi_{G^{(2)}, \pi^{(2)}}] \enspace . 
\end{align*}
We introduce  $\mathcal M_{\mathrm{PM}}\subset \mathcal{M}$ as 
  the collection of perfect matchings, that is matchings $\mathbf M$ such that all the nodes in $(V^{(1)},W^{(1)})$ and $(V^{(2)},W^{(2)})$ are {\bf perfectly matched}. Note that, if $\mathbf M\in \mathcal M_{\mathrm{PM}}$, then $G_{\oplus}$ is the empty graph with $E_{\oplus} = \emptyset$. Besides, $\mathcal{M}_{\mathrm{PM}} \neq \emptyset$ if and only if $G^{(1)}$ and $G^{(2)}$ are equivalent, which amounts to $G^{(1)}= G^{(2)}$ for $G^{(1)}, G^{(2)} \in \mathcal G_{\leq D}$. Let us call $\mathcal{M}^\star$ the collection of matchings $\mathbf{M}$ such that all nodes $v\in V^{(1)}$ (resp. in $V^{(2)}$) whose degree is smaller than or equal to $2m+1$ arise in $\mathbf{M}_1$.

Then, we shall bound the cross moment between $\tilde{\Psi}_{G^{(1)}}$ and $\tilde{\Psi}_{G^{(2)}}$ by applying Proposition~\ref{lem:E_pi1pi2}. Write $\Pi_{\mathrm{(PM)}} = \cup_{\mathbf M \in  \mathcal M_{\mathrm{PM}}} \Pi(\mathbf M)$.
Let us first argue that only the matchings $\mathbf{M}\in \mathcal{M}^{\star}$ contribute to the above sum over $\mathcal{M}$. Consider $\mathbf{M}\in \mathcal{M}\setminus \mathcal{M}^{\star}$ and, by symmetry, assume that some node $v\in V^{(1)}$ with $\mathrm{deg}_{G^{(1)}}(v)\leq 2m+1$ does not arise in $\mathbf{M}_1$. Since $\mathcal{M}_1$ enforces that $(v_1^{(1)},v_1^{(2)})$ and $(v_2^{(1)},v_2^{(2)})$ belong to the matching, we have $v\notin \{v^{(1)}_1,v^{(1)}_2\}$, so that $v$ is non-isolated by the definition of $\mathcal{G}_{\leq D}$ and $\mathrm{deg}_{G^{(1)}}(v)\in [1:2m+1]$. Besides, for any $(\pi^{(1)},\pi^{(2)})\in \Pi(\mathbf{M})$, the equivalence in the definition of $\bar\Pi_1(\mathbf{M}_1)$ ensures that $\pi_1^{(1)}(v)\notin \pi_1^{(2)}(V^{(2)})$, as $v$ is matched to no node of $V^{(2)}$. Hence, Proposition~\ref{lem:E_pi1pi2}(a) yields $\mathbb E[\Psi_{G^{(1)}, \pi^{(1)}}\Psi_{G^{(2)}, \pi^{(2)}}]=0$ for every $(\pi^{(1)},\pi^{(2)})\in \Pi(\mathbf{M})$, so that the term indexed by $\mathbf{M}$ vanishes. Note also that $\mathcal{M}_{\mathrm{PM}}\subset \mathcal{M}^{\star}$, all the nodes being matched for a perfect matching.
Since $\mathbb E[{\Psi}_{G^{(1)}, \pi^{(1)}}{\Psi}_{G^{(2)}, \pi^{(2)}}]$ only depends on $G^{(1)}$, $G^{(2)}$, and $\mathbf{M}$, we deduce from this proposition that
\begin{align}\nonumber
        &\left|\mathbb{E}\left[\tilde{\Psi}_{G^{(1)}}\tilde{\Psi}_{G^{(2)}}\right] - \mathbf 1\{G^{(1)} = G^{(2)}\}|\Pi_{\mathrm{(PM)}}| n^{2 -|V^{(1)}|}d^{-|W^{(1)}|} \frac{1}{|\mathrm{Aut}(G^{(1)})|}\right|\\
        &\leq 2  \frac{1}{\sqrt{|\mathrm{Aut}(G^{(1)})||\mathrm{Aut}(G^{(2)})|}}\sum_{\mathbf M \in \mathcal M^{\star}\setminus  \mathcal M_{\mathrm{PM}}} 
        (32 D^3)^{\cc_{\mathrm{u}}} n^{(|V^{(1)}|+|V^{(2)}|)/2 - |\mathbf{M}_1|}d^{(|W^{(1)}|+|W^{(2)}|)/2- |\mathbf{M}_2|} \left(\frac{4m\Delta}{d^{1/4}}\right)^{|E_{\oplus}|}\nonumber\\ & \,\quad  \quad \quad \quad \quad  \cdot \left(\frac{1}{d}\right)^{(2m+1)|V_{\oplus}|/4 + |W_{\oplus}|/2 - \cc(G_{\oplus})/2 +\cc_{\mathrm{u}}/2} + \frac{8m D \Delta }{d^{1/4}} \mathbf 1\{G^{(1)} = G^{(2)}\}\enspace ,\label{eq:upper_correlation}
\end{align}

Write $C_{\mathbf M}$ for the expression in the rhs of~\eqref{eq:upper_correlation} in the sum over the matchings. We recall that $U^{(1)}_1$ and $U^{(2)}_1$ are the sets of unmatched nodes in resp.~$V^{(1)}$ and $V^{(2)}$, and we write $U_1 = U^{(1)}_1\cup U^{(2)}_1$ for the set of all unmatched nodes on the $V$-side, so that, by~\eqref{eq:unmatched}, $|U_1| = |U^{(1)}_1|+|U^{(2)}_1|$. Also, we have 
\[
|V_{\oplus}|= |V^{(1)}|+ |V^{(2)}|- |\mathbf{M}_1|- |(\mathbf{M}_{\mathrm{PM}})_1|= |U_1| + |\mathbf{M}_1|- |(\mathbf{M}_{\mathrm{PM}})_1|
\]
 Similarly, $|W_{\oplus}|= |W^{(1)}|+ |W^{(2)}|- |\mathbf{M}_2|- |(\mathbf{M}_{\mathrm{PM}})_2|$. We deduce from~\eqref{eq:upper_correlation} that
\begin{align*}
C_{\mathbf M} &\leq  (32 D^3)^{\cc_{\mathrm{u}}} \Big[\sqrt{n} \Big(\frac{4m\Delta}{\sqrt{d}}\Big)^{2m+2} d^{1/4}\Big]^{|U_1|} \Big(\frac{4m\Delta}{d^{1/4}}\Big)^{|E_{\oplus}| - (2m+2)|U_1|} d^{b_{\mathbf{M}}}\ , 
\end{align*}
where $b_{\mathbf{M}}:= 
-(2m+1)[|\mathbf{M}_1|- |(\mathbf{M}_{\mathrm{PM}})_1|]/4 - [|\mathbf{M}_2|- |(\mathbf{M}_{\mathrm{PM}})_2|]/2+ \cc(G_{\oplus})/2-\cc_{\mathrm{u}}/2$. 
Since $\cc(G_{\oplus})- \cc_{\mathrm{u}}$ connected components of $G_{\oplus}$ contain a matched node, we have $\cc(G_{\oplus})-\cc_{\mathrm{u}} \leq |\mathbf M_1|- |(\mathbf{M}_{\mathrm{PM}})_1|+ |\mathbf M_2|- |(\mathbf{M}_{\mathrm{PM}})_2|$. Combined with $m\geq1$, this yields $b_{\mathbf M}\leq [|\mathbf M_1|-|(\mathbf M_{\mathrm{PM}})_1|](1-2m)/4\leq0$, so that $d^{b_{\mathbf M}}\leq 1$.
Since we restrict our attention to $\mathbf{M}\in \mathcal M^{\star}$, each node in $U_1$ must have a degree at least equal to $2m+2$ in $G_{\oplus}$. This implies that $|E_{\oplus}|\geq (2m+2)|U_1|$.
We claim that, moreover,
\begin{equation}\label{eq:ccu:U1}
\cc_{\mathrm{u}}\leq |U_1| + \big[|E_{\oplus}|-(2m+2)|U_1|\big]\enspace .
\end{equation}
Call an unmatched connected component \emph{exceptional} if it contains no node of $U_1$, and write $\cc^{\mathrm{exc}}_{\mathrm{u}}$ for the number of such components, so that 
\begin{equation}\label{eq:ccu:exc}
\cc_{\mathrm{u}}\leq |U_1| + \cc^{\mathrm{exc}}_{\mathrm{u}}\enspace .
\end{equation}
Let us prove that $\cc^{\mathrm{exc}}_{\mathrm{u}}\leq 1$ and that $\cc^{\mathrm{exc}}_{\mathrm{u}}=1$ enforces $|E_{\oplus}|\geq (2m+2)|U_1|+4$, which, together with~\eqref{eq:ccu:exc}, gives~\eqref{eq:ccu:U1}. Consider any exceptional connected component. By symmetry, we assume that it corresponds to a non-trivial connected component $G^{(1)}_r$ of $G^{(1)}$. Since the nodes $V^{(1)}_r$ of $G^{(1)}_r$ are matched, this implies that 
$V^{(1)}_r\subset \{v^{(1)}_1,v^{(1)}_2\}$ and that $v^{(2)}_1$ and $v^{(2)}_2$ are isolated in $G^{(2)}$. Since $G^{(1)}$ has even degrees, we conclude  $|V^{(1)}_r|=2$. Thus,  $\cc^{\mathrm{exc}}_{\mathrm{u}}\leq 1$. If $\cc^{\mathrm{exc}}_{\mathrm{u}}= 1$, then, since the degrees of $G^{(1)}$ and $G^{(2)}$ are even, $|E_{\oplus}|- (2m+2)|U_1|$ is larger than or equal to the number of edges in the unmatched exceptional connected component, which, in turn, is at least 4. We have proved~\eqref{eq:ccu:U1}.

Write $\rho_{\mathrm a}$ and $\rho_{\mathrm b}$ for the first and second terms in the maximum defining $\rho$ in Condition~\eqref{eq:signalcond}, so that $\rho= \rho_{\mathrm a}\vee \rho_{\mathrm b}\leq 1$. Since $\sqrt n \big(4m\Delta/\sqrt d\big)^{2m+2}d^{1/4}= \rho^{2m+2}_{\mathrm a}/(32D^3)$ and $4m\Delta/d^{1/4}= \rho_{\mathrm b}/(32D^3)$, we deduce that
\begin{align*}
C_{\mathbf M}\leq (32D^3)^{\cc_{\mathrm{u}}-|U_1|-[|E_{\oplus}|-(2m+2)|U_1|]}\rho^{(2m+2)|U_1|}_{\mathrm a}\rho^{|E_{\oplus}|-(2m+2)|U_1|}_{\mathrm b}d^{b_{\mathbf M}}\leq \rho^{|E_{\oplus}|}\enspace ,
\end{align*}
where we used $32D^3\geq 1$ together with~\eqref{eq:ccu:U1}, then $d^{b_{\mathbf M}}\leq 1$ and $\rho_{\mathrm a}\vee \rho_{\mathrm b}=\rho\leq 1$. We
conclude that 
\begin{align*}
&\left|\mathbb{E}\left[\tilde{\Psi}_{G^{(1)}}\tilde{\Psi}_{G^{(2)}}\right] - \mathbf 1\{G^{(1)} = G^{(2)}\}|\Pi_{\mathrm{(PM)}}| n^{2 -|V^{(1)}|}d^{-|W^{(1)}|} \frac{1}{|\mathrm{Aut}(G^{(1)})|} \right|\\ 
&\leq 
 2  \frac{1}{\sqrt{|\mathrm{Aut}(G^{(1)})||\mathrm{Aut}(G^{(2)})|}} \sum_{\mathbf M \in \mathcal M^{\star} \setminus  \mathcal M_{\mathrm{PM}}}
         \rho^{|E_{\oplus}|} + \frac{8m D \Delta }{d^{1/4}} \mathbf 1\{G^{(1)} = G^{(2)}\}.
\end{align*}
Since 
\begin{align*}|\Pi_{\mathrm{(PM)}}| = |\mathrm{Aut}(G^{(1)})| \frac{(n-2)!}{(n-|V^{(1)}|)!}\frac{d!}{(d-|W^{(1)}|)!},
\end{align*}
and $(|V^{(1)}|-2) \lor |W^{(1)}| \leq D$, we deduce that 
\begin{align*}
\left||\Pi_{\mathrm{(PM)}}| n^{2 -|V^{(1)}|}d^{-|W^{(1)}|} \frac{1}{|\mathrm{Aut}(G^{(1)})|} - 1\right| &\leq |((n-D-4)/n)^{D}((d-D)/d)^{D} - 1|\\
&\leq  1- \left( 1- \frac{D+4}{n}- \frac{D}{d}\right)^{D}\leq \frac{D^2+4D}{n}+ \frac{D^2}{d}\enspace , 
\end{align*}
since $D\leq (n \land d)/4$, which follows from $2^8D^4\leq d$, $D\geq 2$, and $n\geq d$. And so, using the assumption $n \geq d$, we conclude that
\begin{align}
\left|\mathbb{E}\left[\tilde{\Psi}_{G^{(1)}}\tilde{\Psi}_{G^{(2)}}\right] - \mathbf 1\{G^{(1)} = G^{(2)}\}\right|
&\leq 
   \frac{2}{\sqrt{|\mathrm{Aut}(G^{(1)})||\mathrm{Aut}(G^{(2)})|}} \sum_{\mathbf M \in \mathcal M^{\star}\setminus  \mathcal M_{\mathrm{PM}}}  \rho^{|E_{\oplus}|} + \left(\frac{4D^2}{d}+\frac{8m D \Delta }{d^{1/4}}\right) \mathbf 1\{G^{(1)} = G^{(2)}\}\enspace .\label{upper:correlation:rho}
\end{align}

\paragraph{Shadow matchings.} Given two sets $\bar U^{(1)} = (\bar U^{(1)}_1,\bar U^{(1)}_2)\subset V^{(1)} \times W^{(1)}$ and  $\bar U^{(2)} = (\bar U^{(2)}_1,\bar U^{(2)}_2)\subset V^{(2)} \times W^{(2)}$ and for a set of node matches $\underline{\mathbf M} = (\underline{\mathbf M}_1,\underline{\mathbf M}_2) \in \widetilde{\mathcal M}_1\times \mathcal{M}_2$, where $\widetilde{\mathcal M}_1$ stands for the collection of all matchings of $V^{(1)}$ and $V^{(2)}$ --~without the additional constraint, imposed in the definition of $\mathcal M_1$, that $(v^{(1)}_1,v^{(2)}_1)$ and $(v^{(1)}_2,v^{(2)}_2)$ belong to the matching, as these two pairs, although always present in $\mathbf M_1$, are usually \emph{not} semi-matched~--, we define $\mathcal M_{\mathrm{shadow}}(\bar U^{(1)},\bar U^{(2)}, \underline{\mathbf M})$
 as the collection of matchings $\mathbf{M}$ satisfying 
 \begin{equation}\label{eq:definition:mshadow}
 \mathbf M_{\mathrm{SM}}(\mathbf M) = \underline{\mathbf M}, \quad  (\pi^{(a)}_b)^{-1}(U_b^{(a)})=\bar U^{(a)}_b \text{ for }a=1,2 \text{ and }b=1,2 \enspace  ,
\end{equation}
 for any  $(\pi^{(1)},\pi^{(2)})\in \Pi(\mathbf{M})$. Note that, as long as~\eqref{eq:definition:mshadow} is satisfied for one labeling $(\pi^{(1)},\pi^{(2)})\in \Pi(\mathbf{M})$, it is satisfied for all such $(\pi^{(1)},\pi^{(2)})$. Equivalently, $\mathcal M_{\mathrm{shadow}}(\bar U^{(1)},\bar U^{(2)}, \underline{\mathbf M})$ is the collection of all matchings that lead to the set $\underline{\mathbf M}$ of semi-matched nodes and such that  $\bar U^{(1)},\bar U^{(2)}$ correspond  to unmatched nodes in resp.~$G^{(1)}, G^{(2)}$. We say that these matchings satisfy a given {\bf shadow} $(\bar U^{(1)},\bar U^{(2)}, \underline{\mathbf M} )$. The only thing that can vary between two elements of $\mathcal M_{\mathrm{shadow}}(\bar U^{(1)},\bar U^{(2)}, \underline{\mathbf M} )$ is the matching of the nodes that are not in $\bar U^{(1)},\bar U^{(2)}$, nor part of a pair of nodes in $\underline{\mathbf M}$. This matching must however ensure that all of these nodes are perfectly matched.

Each matching $\mathbf M\in \mathcal M$ gives rise to exactly one shadow, namely $\mathbf s(\mathbf M):= \big(\bar U^{(1)},\bar U^{(2)},\mathbf M_{\mathrm{SM}}(\mathbf M)\big)$, so that the sets $\mathcal M_{\mathrm{shadow}}(\bar U^{(1)}, \bar U^{(2)}, \underline{\mathbf M} )$, when $(\bar U^{(1)}, \bar U^{(2)}, \underline{\mathbf M} )$ ranges over all shadows, form a partition of $\mathcal M$. Let us identify those shadows whose matchings belong to $\mathcal M^{\star}\setminus \mathcal M_{\mathrm{PM}}$. Write $\mathcal S^{\star}$ for the collection of shadows $(\bar U^{(1)}, \bar U^{(2)}, \underline{\mathbf M} )$ such that all nodes $v\in \bar U^{(1)}_1$ (resp.~$v\in \bar U^{(2)}_1$) have degree at least $2m+2$ in $G^{(1)}$ (resp.~in $G^{(2)}$). Since the nodes of $V^{(1)}$ (resp.~$V^{(2)}$) that do not arise in $\mathbf M_1$ are exactly those of $\bar U^{(1)}_1$ (resp.~$\bar U^{(2)}_1$), a matching $\mathbf M$ belongs to $\mathcal M^{\star}$ if and only if $\mathbf s(\mathbf M)\in \mathcal S^{\star}$ --~membership in $\mathcal M^{\star}$ is a condition on the shadow only. Besides, a matching is perfect if and only if all its nodes are matched and none of them is semi-matched, that is if and only if its shadow is the empty shadow $\mathbf s_{\emptyset}:= (\emptyset,\emptyset,\emptyset)$, which does belong to $\mathcal S^{\star}$. Hence, $\mathcal M^{\star}\setminus \mathcal M_{\mathrm{PM}}$ is the disjoint union of the sets $\mathcal M_{\mathrm{shadow}}(\bar U^{(1)}, \bar U^{(2)}, \underline{\mathbf M} )$ over $(\bar U^{(1)}, \bar U^{(2)}, \underline{\mathbf M} )\in \mathcal S^{\star}\setminus \{\mathbf s_{\emptyset}\}$.
We then argue as in~\cite{carpentier2025rank}, proof of Proposition A.2, Steps 3 and 4. We have
\begin{align*}
\sum_{\mathbf M \in \mathcal M^{\star}\setminus  \mathcal M_{\mathrm{PM}}}  \rho^{|E_{\oplus}|} &=   \sum_{\substack{(\bar U^{(1)}, \bar U^{(2)}, \underline{\mathbf M} )\in \mathcal S^{\star}\setminus \{\mathbf s_{\emptyset}\}}}\sum_{\mathbf M \in \mathcal M_{\mathrm{shadow}}(\bar U^{(1)}, \bar U^{(2)}, \underline{\mathbf M} )}  \rho^{|E_{\oplus}|}\\
&\leq \rho^{[d(G^{(1)}, G^{(2)})\lor 1]/2}\sum_{\substack{(\bar U^{(1)}, \bar U^{(2)}, \underline{\mathbf M} )\in \mathcal S^{\star}\setminus \{\mathbf s_{\emptyset}\}}}\sum_{\mathbf M \in \mathcal M_{\mathrm{shadow}}(\bar U^{(1)}, \bar U^{(2)}, \underline{\mathbf M} )}  \rho^{|E_{\oplus}|/2},
\end{align*}
since, by definition of the edit distance, $|E_{\oplus}| \geq d(G^{(1)}, G^{(2)})\lor 1$ as long as  $\mathbf M\not\in \mathcal M_{\mathrm{PM}}$.

Arguing as in the proof of Lemma A.5 in~\cite{CGGV25} --~see also Equation~(117) in~\cite{CGGV25}, which deals with matchings fixing both $v_1$ and $v_2$, and Lemma A.5 in~\cite{carpentier2025rank}~--, we have
$$|\mathcal M_{\mathrm{shadow}}(\bar U^{(1)}, \bar U^{(2)}, \underline{\mathbf M} )| \leq \min(|\mathrm{Aut}(G^{(1)})|, |\mathrm{Aut}(G^{(2)})|)\enspace.$$
Therefore,
$$ \sum_{\mathbf M \in \mathcal M^{\star}\setminus  \mathcal M_{\mathrm{PM}}}  \rho^{|E_{\oplus}|/2} \leq  \min(|\mathrm{Aut}(G^{(1)})|, |\mathrm{Aut}(G^{(2)})|)   \sum_{\substack{(\bar U^{(1)}, \bar U^{(2)}, \underline{\mathbf M} )\in \mathcal S^{\star}\setminus \{\mathbf s_{\emptyset}\}}}  \rho^{(|\bar U^{(1)}| + |\bar U^{(2)}|+ |\underline{\mathbf M} |)/4},$$
since, for any $\mathbf M \in \mathcal M_{\mathrm{shadow}}(\bar U^{(1)}, \bar U^{(2)}, \underline{\mathbf M} )$, each unmatched node and each matched node that is not perfectly matched connects to at least one edge in $E_{\oplus}$.
This leads us to
\begin{align*}
&\left|\mathbb{E}\left[\tilde{\Psi}_{G^{(1)}}\tilde{\Psi}_{G^{(2)}}\right] - \mathbf 1\{G^{(1)} = G^{(2)}\}\right| \nonumber\\
&\leq 
 2 \rho^{[d(G^{(1)}, G^{(2)})\lor 1]/2} \sum_{\substack{(\bar U^{(1)}, \bar U^{(2)}, \underline{\mathbf M} )\in \mathcal S^{\star}\setminus \{\mathbf s_{\emptyset}\}}}  \rho^{(|\bar U^{(1)}| + |\bar U^{(2)}|+ |\underline{\mathbf M} |)/4} + \left[\frac{4D^2}{d}+ \frac{8m D \Delta }{d^{1/4}}\right] \mathbf 1\{G^{(1)} = G^{(2)}\}.
\end{align*}
We then conclude as in the proof of Proposition A.2 (Step 5) in~\cite{carpentier2025rank}: assuming that $\rho \leq D^{-16c_{\mathrm{s}}}$ with $c_{\mathrm{s}} \geq 10$ and $D \geq 2$, we have by summing over $\bar U^{(1)}, \bar U^{(2)},\underline{\mathbf M} $ and grouping them by cardinality --~the last sum being bounded by the sum of $\rho^{(|\bar U^{(1)}| + |\bar U^{(2)}|+ |\underline{\mathbf M} |)/4}$ over \emph{all} non-empty shadows, that is by dropping the constraint $(\bar U^{(1)}, \bar U^{(2)}, \underline{\mathbf M} )\in \mathcal S^{\star}$~--
\begin{align*}
\left|\mathbb{E}\left[\tilde{\Psi}_{G^{(1)}}\tilde{\Psi}_{G^{(2)}}\right] - \mathbf 1\{G^{(1)} = G^{(2)}\}\right| & \leq 
 D^{-c_{\mathrm{s}}[d(G^{(1)}, G^{(2)})\lor 1]}  +   \left(\frac{4D^2}{d}+\frac{8m D \Delta }{d^{1/4}}\right) \mathbf 1\{G^{(1)} = G^{(2)}\}\\  & \leq  2D^{-c_s[d(G^{(1)}, G^{(2)})\lor 1]}+ \frac{4D^2}{d} \mathbf 1\{G^{(1)} = G^{(2)}\} ,
\end{align*}
by definition of $\rho$ and Condition~\eqref{eq:signalcond}.

\medskip 

Finally, it remains to bound the term $\mathbb{E}[\tilde{\Psi}_{G^{(1)}}]$ which corresponds to the cross moment of $1$ and $\tilde{\Psi}_{G^{(1)}}$. Note that the previous arguments and  derivations still hold in the degenerate situation where $G^{(2)}$ is the empty graph, so that $\Psi_{G^{(2)}}=1$, $G_{\oplus}=G^{(1)}$ and $|E_{\oplus}|=|E^{(1)}|$. There is then no matched node, $\cc_{\mathrm{u}}=\cc(G^{(1)})$, and the set $U_1$ of unmatched nodes on the $V$-side is the set of non-isolated nodes of $V^{(1)}$; as it contains $V^{(1)}\setminus\{v_1,v_2\}$, we have $n^{(|V^{(1)}|-2)/2}\leq n^{|U_1|/2}$. In that degenerate situation, 
 every non-isolated node of $V^{(1)}$ has degree at least $2m+2$ --otherwise $\mathbb{E}[\tilde\Psi_{G^{(1)}}]=0$ by Proposition~\ref{lem:E_pi1pi2}(a)-- and we again have $\cc_{\mathrm{u}}\leq  |U_1|+ |E^{(1)}|- (2m+2)|U_1|$ so that~\eqref{eq:ccu:U1} remains valid. In particular, we readily get
\[
\big|\mathbb{E}[\tilde{\Psi}_{G^{(1)}}]\big|\leq \frac{2}{\sqrt{|\mathrm{Aut}(G^{(1)})|}} \rho^{|E^{(1)}|}\leq 2D^{-c_s |E^{(1)}|}\enspace .
\]

\section{Proof of Lemma~\ref{lem:scalprod}}\label{sec:proof:lem:scalprod}

Fix a graph $G$ and a labeling $\pi$ of $G$.

\medskip

\noindent 
{\bf Case 1}.
First, we deal with the case where either  $v_1$ or $v_2$ is isolated in $G$. Then,  we readily check that $\mathbb{E}[x\Psi_{G,\pi}]=\frac{1}{K}\mathbb{E}[\Psi_{G,\pi}]$. Besides, if $G$ contains only one non-trivial connected component, then it follows from the definition of $\Psi_{G,\pi}$ that $\mathbb{E}[\Psi_{G,\pi}]= 0$, which leads to $\mathbb{E}[x\tilde{\Psi}_{G}]=0$. If $G$ is made of several connected components, then we can decompose $G$ into two disjoint graphs $G_1$ and $G_2$ and we are in a position to apply Proposition~\ref{lem:E_pi1pi2}. In particular, we also have $\mathbb{E}[x\tilde{\Psi}_{G}]=0$ if one node $v\in V\setminus\{v_1,v_2\}$ has a degree smaller than $2m+2$. Otherwise, we have 
\begin{align*}
 \left|\mathbb{E}\left[\Psi_{G,\pi}\right]\right|&\leq 2 (32 D^3)^{\cc(G)}  \left(4m\frac{\Delta}{d^{1/4}}\right)^{|E|}\left(\frac{1}{d}\right)^{(2m+1)|V_{\oplus}|/4 + |W|/2 }\enspace ,
\end{align*}
where $|V_{\oplus}|$ is the number of non-isolated nodes of $V$ in the disjoint union $G_1\sqcup G_2$ (Proposition~\ref{lem:E_pi1pi2}(b) is stated in terms of $V_{\oplus}$, which excludes isolated nodes), and where we also used $\cc(G)\leq |V_{\oplus}|$ (each non-trivial connected component contains at least one non-isolated node of $V$). Since only $v_1$ and $v_2$ may be isolated in a graph of $\mathcal G_{\leq D}$, we have $|V_{\oplus}|\geq |V|-2$. Since $\cc(G)\leq |V_{\oplus}|$ and $d$ is supposed to be large enough compared to $D$, we have
\[(32D^3)^{\cc(G)}(1/d)^{(2m+1)|V_{\oplus}|/4}\leq (32D^3)^{|V|-2}(1/d)^{(2m+1)(|V|-2)/4} = (32D^3)^{|V|-2}(1/d)^{(2m+1)|V|/4}\cdot d^{(2m+1)/2}\enspace \]
Summing over all labelings $\pi$, we arrive at
\begin{align*}
\left|\mathbb{E}[x\tilde{\Psi}_{G}]\right|& \leq 2 \frac{d^{(2m+1)/2}}{Kd^{m+1/2}}(32 D^3)^{|V|-2}  \left[ \left(4m\frac{\Delta}{d^{1/2}}\right)^{2m+2} d^{1/4}\sqrt{n}\right]^{(|V|-2)}    \left(4m\frac{\Delta}{d^{1/4}}\right)^{|E|-(2m+2)(|V|-2)}\\
&\leq 2 \frac{1}{K}  \left[ \left(4m\frac{\Delta}{d^{1/2}}\right)^{2m+2} (32 D^3)d^{1/4}\sqrt{n}\right]^{(|V|-2)}    \left(4m\frac{\Delta}{d^{1/4}}\right)^{|E|-(2m+2)(|V|-2)} \\
&\leq \frac{2}{K} \rho^{|E|}\enspace  ,
\end{align*}
where the first line uses the bound on $(32D^3)^{\cc(G)}(1/d)^{(2m+1)|V_\oplus|/4}$ established above, the second line uses $d^{(2m+1)/2}/d^{m+1/2}=1$, and the last line uses that each node $v\in V\setminus\{v_1,v_2\}$ has a degree at least $2m+2$. We have shown the desired inequality.

\medskip 

Let us turn to the more interesting situation where neither $v_1$ nor $v_2$ is isolated in $G$.

\noindent 
{\bf Case 2}.
First, we focus on the case where {\bf $v_1$ and $v_2$ belong to the same connected component} of $G$. We decompose $G$ into $G_0$ and $G_1$ where $G_1$ stands for this connected component and $G_0$ contains all the remaining connected components. We use the convention that $\Psi_{G_0,\pi}=1$ if $G_0$ is the empty graph. Then, we can decompose
\begin{align}\label{eq:decomposition}
 \mathbb{E}\left[x\Psi_{G,\pi}\right]=   \mathbb{E}[xP_{G_1,\pi}\Psi_{G_0,\pi}]-\mathbb{E}[x\Psi_{G_0,\pi}]\mathbb{E}[P_{G_1,\pi}]= \mathbb{E}[xP_{G_1,\pi}\Psi_{G_0,\pi}] - \frac{1}{K}\mathbb{E}[P_{G_1,\pi}]\mathbb{E}[\Psi_{G_0,\pi}]\enspace . 
\end{align} 
Define the graph $\bar{G}_1=(\bar{V}_1,\bar{W}_1,\bar{E}_1)$ by identifying $v_1$ and $v_2$ into $v_1$ and removing all edges that appear twice, that is all edges that connect $v_1$ and $v_2$ to a common neighbor $w\in W$.  One  checks that 
\begin{align*}
\mathbb{E}[xP_{G_1,\pi}\Psi_{G_0,\pi}]= \frac{1}{K}\left(\frac{\Delta}{\sqrt{d}}\right)^{|E_1|-|\bar{E}_1|}\mathbb{E}[P_{\bar{G}_1,\pi}\Psi_{G_0,\pi}]\enspace .
%= \frac{1}{K}\left(\frac{\Delta}{\sqrt{d}}\right)^{|E_1|-|\bar{E}_1|} \left[\mathbb{E}\left[P_{\bar{G}_1,\pi}\right]\mathbb{E}[\Psi_{G_0,\pi}] + \mathbb{E}[\Psi_{\bar{G},\pi}] \right]\ , 
\end{align*}
As in Case 1, we readily check that $\mathbb{E}\left[x\Psi_{G,\pi}\right]=0$ if  one node $v\in V\setminus\{v_1,v_2\}$ has a degree smaller than $2m+2$.
Define $\bar{G}=(\bar{V},\bar{W},\bar{E})$ by combining $\bar{G}_1$ and $G_0$ and removing isolated nodes.
To control the term $\mathbb{E}[P_{\bar{G}_1,\pi}\Psi_{G_0,\pi}]$, we apply Proposition~\ref{lem:E_pi1pi2} (d). Note that $\cc(G_0)\leq|V|-2$, since $G_0$ does not contain $v_1$ or $v_2$ (both being in $G_1$) and every non-trivial connected component has at least one node in $V$.
This leads us to
\begin{align}
\left|\mathbb{E}[xP_{G_1,\pi}\Psi_{G_0,\pi}]\right|\leq \frac{1}{K}(32 D^3)^{\cc(G_0)}\left(4m\frac{\Delta}{d^{1/4}}\right)^{|E|}
\left(\frac{1}{d}\right)^{(|E_1|-|\bar{E}_1|)/4+ (2m+1)|\bar{V}|/4 + |\bar{W}|/2 - \cc(\bar{G}_1)/2}\enspace . \label{eq:1:upper:G1}
\end{align}
Applying Lemma~\ref{lem:eqcounting} and Proposition~\ref{lem:E_pi1pi2}, we get 
\begin{align*}
 \frac{1}{K}\left|\mathbb{E}[P_{G_1,\pi}]\mathbb{E}[\Psi_{G_0,\pi}]\right| \leq \frac{4}{K}(32 D^3)^{\cc(G_0)}   \left(4m \frac{\Delta}{d^{1/4}}\right)^{|E|}\left(\frac{1}{d}\right)^{|W|/2 + (2m+1)|V|/4 - 1/2}\enspace .
\end{align*}
Then, we combine the two previous inequalities with the decomposition~\eqref{eq:decomposition} and we sum over all possible labelings $\pi$. This leads us to 
\begin{align}
\left|\mathbb{E}\left[x \tilde{\Psi}_G\right]\right|&\leq \frac{4}{K} (32 D^3)^{\cc(G_0)} \left[\left(4m\frac{\Delta}{d^{1/2}} \right)^{2m+2} d^{1/4}\sqrt{n} \right]^{|V|-2}\left(4m \frac{\Delta}{d^{1/4}}\right)^{|E|-(2m+2)(|V|-2)}\left[\frac{1}{d^{m}}+ \frac{1}{d^a}\right]\ ; \label{eq:x:psi} \\
a&:=  (|E_1|-|\bar{E}_1|)/4+ (2m+1)[|\bar{V}|-|V|+2]/4 + [|\bar{W}|-|W|]/2 - \cc(\bar{G}_1)/2\enspace . \nonumber
\end{align}
Let us prove that $a$ is non-negative. 
Set $r=|W|-|\bar{W}|$: these $r$ nodes are connected to $v_1$ and $v_2$ in $G$ and do not have any other neighbor. If $\cc(\bar{G}_1)=0$, we readily deduce that $a\geq 0$. If $\cc(\bar{G}_1)>0$, then each connected component of $\bar{G}_1$ that does not contain $v_1$ contains a node $w$ of degree at least $4$ in $G$ which is also connected to $v_1$ and $v_2$ in $G$. Denoting by $\cc_-(\bar{G}_1)$ the number of connected components of $\bar{G}_1$ that do not contain $v_1$, we deduce that 
\[
(|E_1|-|\bar{E}_1|)\geq 2 [|W|- |\bar{W}|]+  2\cc_-(\bar{G}_1)\ . 
\]
We conclude that 
\[
a\geq (2m+1)[|\bar{V}|-|V|+2]/4  - (\cc(\bar{G}_1)- \cc_-(\bar{G}_1))/2\geq 0\enspace . 
\]
Hence, we deduce from~\eqref{eq:x:psi}
and the definition of $\rho$ that 
\[
\left|\mathbb{E}\left[x \tilde{\Psi}_G\right]\right|\leq \frac{8}{K}\rho^{|E|}
\]

\bigskip 

\noindent 
{\bf Case 3}.
Finally, we consider the case where $v_1$ and $v_2$ belong to distinct connected components of $G$. 
As in the previous cases, we readily check that $\mathbb{E}\left[x\Psi_{G,\pi}\right]=0$ if  one node $v\in V\setminus\{v_1,v_2\}$ has a degree smaller than $2m+2$.
We decompose $G$ into $G_0$, $G_1$, and $G_2$  where $G_1$ (resp. $G_2$) stands for the connected component of $v_1$ (resp. $v_2$) and $G_0$ contains all the remaining connected components. We use the convention that $\Psi_{G_0,\pi}=1$ if $G_0$ is the empty graph. Define the graph $\bar{G}_1=(\bar{V}_1,\bar{W}_1,\bar{E}_1)$ by identifying $v_1$ and $v_2$ into $v_1$ in the disjoint union $G_1\sqcup G_2$. Note that $\bar{G}_1$ is connected.
Then, we can decompose
\begin{align} \nonumber
 \mathbb{E}\left[x\Psi_{G,\pi}\right]&=   \mathbb{E}[xP_{G_1,\pi}P_{G_2,\pi}\Psi_{G_0,\pi}]+ \mathbb{E}[x\Psi_{G_0,\pi}]\mathbb{E}[P_{G_1,\pi}]\mathbb{E}[P_{G_2,\pi}]\\ &\quad  - \mathbb{E}[x\Psi_{G_0,\pi} \nonumber P_{G_1,\pi}]\mathbb{E}[P_{G_2,\pi}]- \mathbb{E}[x\Psi_{G_0,\pi}P_{G_2,\pi}]\mathbb{E}[P_{G_1,\pi}]\\ 
 & =\frac{1}{K}\mathbb{E}[P_{\bar{G}_1,\pi}\Psi_{G_0,\pi}]+\frac{1}{K}\mathbb{E}[\Psi_{G_0,\pi}]\mathbb{E}[P_{G_1,\pi}]\mathbb{E}[P_{G_2,\pi}]\nonumber \\& \quad  -\frac{1}{K} \mathbb{E}[\Psi_{G_0,\pi}P_{G_1,\pi}]\mathbb{E}[P_{G_2,\pi}]- \frac{1}{K}\mathbb{E}[\Psi_{G_0,\pi}P_{G_2,\pi}]\mathbb{E}[P_{G_1,\pi}] \label{eq:decomposition2}
 \enspace . 
\end{align}
Each of these expressions is bounded using either Proposition~\ref{lem:E_pi1pi2} or Lemma~\ref{lem:eqcounting}, using as in Case 2 that $\cc(G_0)\leq|V|-2$. This leads us to
\begin{align*}
\left|\mathbb{E}\left[x\Psi_{G,\pi}\right]\right|
\leq \frac{16}{K}(32 D^3)^{\cc(G_0)}   \left(4m \frac{\Delta}{d^{1/4}}\right)^{|E|}\left(\frac{1}{d}\right)^{|W|/2 + (2m+1)(|V|-1)/4 - 1/2}\enspace .
\end{align*}
Summing over all possible labelings $\pi$ yields 
\begin{align}
\left|\mathbb{E}\left[x \tilde{\Psi}_G\right]\right|&\leq \frac{16}{K} (32 D^3)^{\cc(G_0)} \left[\left(4m\frac{\Delta}{d^{1/2}} \right)^{2m+2} d^{1/4}\sqrt{n} \right]^{|V|-2}\left(4m \frac{\Delta}{d^{1/4}}\right)^{|E|-(2m+2)(|V|-2)}d^{1/2-(2m+1)/4}\enspace .
\end{align}
Since $m\geq 1$, we conclude that
\[
\left|\mathbb{E}\left[x \tilde{\Psi}_G\right]\right|\leq \frac{16}{Kd^{1/4}}\rho^{|E|}\enspace .
\]

This concludes the proof.

\section{Proof of reduction lemmas}

\begin{proof}[Proof of Lemma~\ref{lem:reduction:p}]
The distribution of $(x,Y)$ is invariant under any permutation of the rows in $[3:n]$ and any permutation of $[1:d]$. As a consequence, the supremum in~\eqref{eq:def:corr} may be restricted to polynomials $P$ that satisfy these invariance properties (indeed, for any $P$, its symmetrization $\bar P:=\mathbb E_g[P\circ g]$ over the invariance group satisfies $\mathbb E[x\bar P]=\mathbb E[xP]$ by invariance of the law of $(x,Y)$, and $\mathbb E[\bar P^2]\leq\mathbb E[P^2]$ by Jensen's inequality, so that $\bar P$ is admissible in~\eqref{eq:def:corr} as soon as $P$ is, achieves the same value, and is itself invariant). Since $\Big(1, (P_{G})_{G\in \mathcal{G}_{\leq D}}\Big)$ is a basis of the invariant polynomials with degrees at most equal to $D$ (see Section~\ref{sec:graph}), this concludes the proof.
\end{proof}

\begin{proof}[Proof of Lemma~\ref{lem:first_moment_bch}]
Fix $\pi= (\pi_1,\pi_2)$. We have
\begin{align*}
\mathbb E[P_{G,\pi_1,\pi_2}(Y)]&= \mathbb E_{a,z,\kg,\sigma}\Bigg[\prod_{e\in E} z_{\pi_1(l(e))}  X_{\kg(\pi_1(l(e))),\pi_2(r(e))}\Bigg]\\
&= \left(\frac{\Delta}{\sqrt{d}}\right)^{|E|}\mathbb E_{a,z,\kg,\sigma}\Bigg[\prod_{e\in E} z_{\pi_1(l(e))} a_{\pi_2(r(e))} \mu_{\kg(\pi_1(l(e))),\sigma(\pi_2(r(e)))}\Bigg]\enspace . 
\end{align*}
Similarly, we have 
\begin{align*}
\mathbb E[xP_{G,\pi_1,\pi_2}(Y)]&= \left(\frac{\Delta}{\sqrt{d}}\right)^{|E|}\mathbb E_{a,z,\kg,\sigma}\Bigg[x \prod_{e\in E} z_{\pi_1(l(e))} a_{\pi_2(r(e))} \mu_{\kg(\pi_1(l(e))),\sigma(\pi_2(r(e)))}\Bigg]\enspace .     
\end{align*}

\noindent
\underline{Proof of a).} Write $w^*\in W$ for (one of the) nodes of odd degree, and write $\alpha$ for its odd degree. We have
\begin{align*}
 \mathbb E[P_{G,\pi_1,\pi_2}(Y)]&= \left(\frac{\Delta}{\sqrt{d}}\right)^{|E|} \mathbb E_{a,z,\kg,\sigma}\Bigg[\prod_{e\in E \setminus \{e\in E: r(e)=w^*\}} z_{\pi_1(l(e))}  a_{\pi_2(r(e))} \mu_{\kg(\pi_1(l(e))),\sigma(\pi_2(r(e)))} \\
&\quad \quad  \cdot  \prod_{e\in E: r(e)=w^*} z_{\pi_1(l(e))}  a_{\pi_2(w^*)} \mu_{\kg(\pi_1(l(e))),\sigma(\pi_2(w^*))} \Bigg]\\
&= \left(\frac{\Delta}{\sqrt{d}}\right)^{|E|}\mathbb E_{(a_{\pi_2(w)})_{w\neq w^*},z,\kg,\sigma}\Bigg[\prod_{e\in E \setminus \{e\in E: r(e)=w^*\}} z_{\pi_1(l(e))}  a_{\pi_2(r(e))} \mu_{\kg(\pi_1(l(e))),\sigma(\pi_2(r(e)))} \\
&\quad \quad \cdot  \quad \quad \prod_{e\in E: r(e)=w^*} z_{\pi_1(l(e))}  \mu_{\kg(\pi_1(l(e))),\sigma(\pi_2(w^*))} \Bigg] \mathbb E_{a_{\pi_2(w^*)}}[a_{\pi_2(w^*)}^\alpha]\\
&=0\enspace ,
\end{align*}
because $\alpha$ is odd and the entries of $a$ are independent Rademacher random variables. Similarly, $a_{\pi_2(w^*)}$ is independent of the functional $x$ and we obtain $\mathbb E[xP_{G,\pi_1,\pi_2}(Y)]=0$.

\medskip
\noindent
\underline{Proof of b).} Write $v^*\in V$ for (one of the) nodes of odd degree, and write $\alpha$ for its odd degree. We have
\begin{align*}
\mathbb E[P_{G,\pi_1,\pi_2}(Y)]&= \left(\frac{\Delta}{\sqrt{d}}\right)^{|E|} \mathbb E_{a,z,\kg,\sigma}\Bigg[\prod_{e\in E \setminus \{e\in E: l(e)=v^*\}} z_{\pi_1(l(e))}  a_{\pi_2(r(e))} \mu_{\kg(\pi_1(l(e))),\sigma(\pi_2(r(e)))} \\
&\quad \cdot  \prod_{e\in E: l(e)=v^*} z_{\pi_1(v^*)} a_{\pi_2(r(e))} \mu_{\kg(\pi_1(v^*)),\sigma(\pi_2(r(e)))} \Bigg]\\
&=\left(\frac{\Delta}{\sqrt{d}}\right)^{|E|} \mathbb E_{a,(z_{\pi_1(v)})_{v\neq v^*},\kg,\sigma}\Bigg[\prod_{e\in E \setminus \{e\in E: l(e)=v^*\}} z_{\pi_1(l(e))}  a_{\pi_2(r(e))} \mu_{\kg(\pi_1(l(e))),\sigma(\pi_2(r(e)))} \\
&\quad \cdot  \prod_{e\in E: l(e)=v^*}  a_{\pi_2(r(e))} \mu_{\kg(\pi_1(v^*)),\sigma(\pi_2(r(e)))} \Bigg] \mathbb E_{z_{\pi_1(v^*)}}[z_{\pi_1(v^*)}^\alpha]\\
&=0,
\end{align*}
because $\alpha$ is odd and the entries of $z$ follow independent Rademacher distributions. As the entries of $z$ are also independent of $x$, we also have $\mathbb E[xP_{G,\pi_1,\pi_2}(Y)]=0$.
\end{proof}

\begin{proof}[Proof of Lemma~\ref{lem:moment:ordre2:odd:degree}]

Consider the labeled graph $G_{\oplus}$ formed by the symmetric difference of $(\pi_1^{(1)},\pi_2^{(1)})[G^{(1)}]$ and $(\pi_1^{(2)},\pi_2^{(2)})[G^{(2)}]$. We have

\begin{align}\label{eq:moment_ordre_2}
\mathbb{E}\left[P_{G^{(1)},\pi^{(1)}}P_{G^{(2)},\pi^{(2)}}\right] & = \left(\frac{\Delta}{\sqrt{d}}\right)^{|E_{\oplus}|}\mathbb{E}_{a,z,\kg,\sigma}\left[\prod_{e\in E_{\oplus}} z_{l(e)} a_{r(e)} \mu_{\kg(l(e)),\sigma(r(e))}\right]\enspace .
\end{align}

Suppose that there exists $v_0\in V^{(2)}$ whose degree is odd. Then, in $E_{\oplus}$, the degree of $\pi^{(2)}_1(v_0)$ is also odd. Hence, $z_{\pi^{(2)}_1(v_0)}$ arises an odd number of times in~\eqref{eq:moment_ordre_2}. Since the $z_i$'s are independent and follow independent Rademacher distributions, we deduce that $\mathbb{E}\left[P_{G^{(1)},\pi^{(1)}}P_{G^{(2)},\pi^{(2)}}\right]=0$.

Now assume that there exists $w_0\in W^{(2)}$ whose degree is odd. Then, in $E_{\oplus}$, the degree of $\pi^{(2)}_2(w_0)$ is also odd. Hence, $a_{\pi^{(2)}_2(w_0)}$ arises an odd number of times in~\eqref{eq:moment_ordre_2}. Since the $a_j$'s are independent and follow independent Rademacher distributions, we deduce that $\mathbb{E}\left[P_{G^{(1)},\pi^{(1)}}P_{G^{(2)},\pi^{(2)}}\right]=0$.

\end{proof}

\begin{proof}[Proof of Lemma~\ref{lem:reduction:psi_even_bis}]

By Lemma~\ref{lem:reduction:p}, we have $\mathrm{Corr}_{\leq D}= \sup\big\{\mathbb E[xP]:\ P\in \mathcal{S},\ \mathbb E[P^2]\leq 1\big\}$, where we write $\mathcal{S}:= \mathrm{span}\big(1,(P_G)_{G\in \mathcal{G}_{\leq D}}\big)$, so that we may restrict our attention to polynomials lying in $\mathcal{S}$. Consider any polynomial $P\in \mathcal{S}$ such that $\mathbb E[P^2]\leq 1$.

Decompose $P=P_1+ P_2$ where $P_1$ lies in the span of $1$ and the $P_G$ with $G\in \mathcal{G}^{\mathrm{even}}_{\leq D}$ and $P_2$ lies in the span of $P_G$ with $G\in \mathcal{G}_{\leq D}\setminus \mathcal{G}^{\mathrm{even}}_{\leq D}$. By Lemmas~\ref{lem:first_moment_bch}
and~\ref{lem:moment:ordre2:odd:degree}, we have $\mathbb{E}[xP_2]=0$ and $\mathbb{E}[P_1P_2]=0$, so that $\mathbb{E}[P_1^2]\leq \mathbb{E}[P_1^2]+\mathbb{E}[P_2^2]=\mathbb{E}[P^2]\leq 1$ and $\mathbb{E}[xP]=\mathbb{E}[xP_1]$, that is, $P_1$ is admissible and achieves the same value as $P$. Since, conversely, any polynomial in the span of $1$ and of the $P_G$ with $G\in \mathcal{G}^{\mathrm{even}}_{\leq D}$ belongs to $\mathcal S$, we have
\[
\mathrm{Corr}_{\leq D}= \sup_{P\in \mathrm{span}\big(1, (P_{G})_{G\in \mathcal{G}^{\mathrm{even}}_{\leq D}}\big):\ \mathbb{E}[P^2]\leq 1} \mathbb{E}[xP]\enspace . 
\]
To conclude, it remains to check that 
$\mathrm{span}\Big(1, (P_{G})_{G\in \mathcal{G}^{\mathrm{even}}_{\leq D}}\Big)=\mathrm{span}\Big(1, (\Psi_{G})_{G\in \mathcal{G}^{\mathrm{even}}_{\leq D}}\Big)$. Fix any $G\in \mathcal{G}^{\mathrm{even}}_{\leq D}$. We easily check by expanding 
$\Psi_{G}$ that it lies in the span of $1$ and of the $P_{G'}$'s. Conversely, $P_G$ is a linear combination of $1$ and all $\Psi_{G'}$ where the $G'$ are all subgraphs of $G$ (up to equivalence between graphs) where some of the connected components of $G$ have been removed.

\end{proof}

\section{Proofs of the properties of the BCH construction}\label{sec:proof:bch}

In this section, we prove Lemmas~\ref{lem:BCHeq} and~\ref{lem:BCHdesign}, stated in Section~\ref{sec:bch:code}, as well as Lemmas~\ref{lem:moments:XY} and~\ref{lem:BCH:separation} on the moments and the separation of the centers.

\begin{proof}[Proof of Lemma~\ref{lem:BCHeq}]
    
   We have 
      \begin{align*}
       \Big(\frac{1}{K}\sum_{k \leq K} \mu_k^{\otimes l}\Big)_{j_1,\ldots, j_l} &= \frac{1}{K}\sum_{k \leq K} \prod_{i \leq l}\mu_{k,j_i}\\
       &= \frac{1}{K}\sum_{k \leq K} \prod_{i \leq l} H_{j_i}\Bigg(\Big(2(\bar \psi(k-1))_{a,b}-1\Big)_{a\leq m, b\leq u}\Bigg)\\
       &= \mathbb E_{R \sim \mathcal R^{m\times u}} \Big[\prod_{i \leq l} H_{j_i}(R)\Big],
   \end{align*}
where $\mathcal R$ is the Rademacher distribution. This holds since $2\bar\psi(k-1)-1$ is uniformly distributed on $\{-1,1\}^{m\times u}$ when $k$ is uniformly distributed on $\{1,\ldots,K\}$ -- recall $K=2^{um}$.
Hence, we have
    \begin{align*}
        \mathbb E_{R \sim \mathcal R^{m\times u}} \Big[\prod_{i \leq l} H_{j_i}(R)\Big] &= \mathbb E_{R \sim \mathcal R^{m\times u}} \Big[\prod_{i=1}^l \prod_{a=1}^m \prod_{b=1}^u (R_b^{(a)})^{[\phi^{(2a-1)}(j_i)]_b}\Big]\\
        &= \mathbb E_{R \sim \mathcal R^{m\times u}} \Big[\prod_{a=1}^m \prod_{b=1}^u (R_b^{(a)})^{\sum_{i=1}^l [\phi^{(2a-1)}(j_i)]_b}\Big]\\
        &= \mathbf 1\Big\{\forall a \leq m, \forall b \leq u : \sum_{i=1}^l [\phi^{(2a-1)}(j_i)]_b~\mathrm{even}\Big\}.
    \end{align*}
    This concludes the proof by remarking that the sum in $\mathbb F_2^u$ has a given coordinate equal to $0$ if and only if the standard sum is even.
\end{proof}

The following lemma is the key property of the BCH construction, see~\cite{macwilliams1977theory}. Here, we provide a self-contained proof without reference to coding theory.

\begin{lemma}\label{lem:fundBCH}
Consider a positive integer $l \leq 2m$ and consider $l$ elements, not necessarily distinct, $v_1,\ldots, v_l \in \mathbb F_{2^u}\setminus\{0\}$$=\{1,\ldots,d\}$. Assume that for any $a \leq m $
$$\sum_{i=1}^l \phi^{(2a-1)}(v_i) = 0,$$
where the sum here is taken in $\mathbb F_{2}^u$ - and therefore corresponds to a coordinate-wise sum of the $u$-dimensional vectors, modulo $2$. Then $l$ is even and the vector $(v_1,\ldots,v_l)$ is paired, that is its entries can be grouped two by two into $l/2$ pairs of equal elements.
\end{lemma}

\begin{proof}[Proof of Lemma~\ref{lem:fundBCH}]
For any integer $k\geq 1$, write
$$p_k := \sum_{i=1}^l v_i^{k}\ \in \mathbb F_{2^u}$$
for the $k$-th power sum of $v_1,\ldots,v_l$, all the operations being those of the field $\mathbb F_{2^u}$, which has characteristic $2$. Since $\phi^{(k)}(v)=\psi(v^k)$ by~\eqref{eq:definition_phi} and since $\psi$ is an isomorphism of additive groups from $\mathbb F_{2^u}$ onto $\mathbb F_2^u$, we have, for every $a\leq m$,
$$\sum_{i=1}^l \phi^{(2a-1)}(v_i) = \sum_{i=1}^l \psi\big(v_i^{2a-1}\big) = \psi\Big(\sum_{i=1}^l v_i^{2a-1}\Big) = \psi\big(p_{2a-1}\big)\enspace ,$$
where the first two sums are taken in $\mathbb F_2^u$ and the third one in $\mathbb F_{2^u}$. As $\psi(x)=0$ if and only if $x=0$, the assumption of the lemma is therefore equivalent to
\begin{equation}\label{eq:fundBCH:hyp}
p_{2a-1}=0\enspace , \qquad \text{for every } a=1,\ldots,m\enspace .
\end{equation}

\emph{Step 1: all the power sums $p_1,\ldots,p_{2m}$ vanish.} As $\mathbb F_{2^u}$ has characteristic $2$, we have $(x+y)^2=x^2+2xy+y^2=x^2+y^2$ for all $x,y$, so that $x\mapsto x^2$ is additive (this is the Frobenius endomorphism). Consequently, for every integer $k\geq 1$,
\begin{equation}\label{eq:fundBCH:frob}
p_{2k}=\sum_{i=1}^l v_i^{2k} = \sum_{i=1}^l \big(v_i^{k}\big)^2 = \Big(\sum_{i=1}^l v_i^{k}\Big)^2 = p_k^2\enspace .
\end{equation}
We prove by strong induction on $k$ that $p_k=0$ for every $k=1,\ldots,2m$. Let $1\leq k \leq 2m$ and assume that $p_{k'}=0$ for every $1\leq k'<k$. If $k$ is odd, then $k\leq 2m-1$, so that $k=2a-1$ with $a=(k+1)/2\leq m$, and $p_k=0$ by~\eqref{eq:fundBCH:hyp}. If $k$ is even, write $k=2k'$ with $k'=k/2$; then $1\leq k'<k$, so that $p_{k'}=0$ by the induction hypothesis, and $p_k=p_{k'}^2=0$ by~\eqref{eq:fundBCH:frob}. This proves the claim, namely
\begin{equation}\label{eq:fundBCH:allp}
p_k = 0\enspace ,\qquad \text{for every } k=1,\ldots,2m\enspace .
\end{equation}

\emph{Step 2: reduction to pairwise distinct values.} For $v\in \mathbb F_{2^u}\setminus\{0\}$, let $n_v := |\{i\leq l:\ v_i=v\}|$ denote the multiplicity of $v$ in the list $(v_1,\ldots,v_l)$, so that $\sum_{v} n_v = l$. Since $\mathbb F_{2^u}$ has characteristic $2$, we have $n_v v^k = (n_v \bmod 2)\,v^k$ for every $k$, so that, letting $w_1,\ldots,w_s$ be the (pairwise distinct, nonzero) elements $v$ whose multiplicity $n_v$ is \emph{odd},
\begin{equation}\label{eq:fundBCH:oddpart}
p_k = \sum_{v} n_v v^{k} = \sum_{j=1}^s w_j^{k}\enspace , \qquad \text{for every } k\geq 1\enspace .
\end{equation}
Note that $s$ is at most the number of distinct values among $v_1,\ldots,v_l$, so that $s\leq l \leq 2m$.

We easily check that $(v_1,\ldots,v_l)$ is paired if and only if $s=0$. It therefore only remains to prove that $s=0$.

\emph{Step 3: $s=0$.} Assume, for the sake of contradiction, that $s\geq 1$. Note that $s=1$ is not possible since $0=p_1=w_1$ contradicts the fact that the {$w_j$}'s are nonzero. Thus, we assume that $s\geq 2$ and   
 we introduce the two polynomials
$$Q(T) := \prod_{j=1}^s (1-w_j{T})\enspace , \qquad N(T):= \sum_{j=1}^s w_j{T}\prod_{i\neq j}(1-w_i{T})\enspace ,$$
with coefficients in $\mathbb F_{2^u}$, and where the second product ranges over $i\in\{1,\ldots,s\}\setminus\{j\}$. We work in the ring $\mathbb F_{2^u}[[T]]$ of formal power series over $\mathbb F_{2^u}$. Since $Q(0)=1$, the series $Q$ is invertible in $\mathbb F_{2^u}[[T]]$, and, since $\sum_{k\geq 1}(w_j{T})^k = w_j{T}/(1-w_j{T})$ in $\mathbb F_{2^u}[[T]]$ for every $j$, we deduce from~\eqref{eq:fundBCH:oddpart} that
$$\sum_{k\geq 1} p_k {T}^k = \sum_{j=1}^s \sum_{k\geq 1} (w_j{T})^k = \sum_{j=1}^s \frac{w_j{T}}{1-w_j{T}} = \frac{N(T)}{{Q}(T)}\enspace ,$$
the last equality being obtained by putting the $s$ fractions over the common denominator $Q(T)$. By~\eqref{eq:fundBCH:allp}, the coefficients of ${T}^0,{T}^1,\ldots,{T}^{2m}$ in the power series $N(T)/{Q}(T)=\sum_{k\geq 1}p_k{T}^k$ all vanish, that is $N(T)/{Q}(T)\in {T}^{2m+1}\mathbb F_{2^u}[[T]]$. Multiplying by $Q(T)$, we get $N(T)\in {T}^{2m+1}\mathbb F_{2^u}[[T]]$. But $N$ is a polynomial of degree at most $1+(s-1)=s\leq 2m < 2m+1$, so that this forces
\begin{equation}\label{eq:fundBCH:Nzero}
N = 0\enspace .
\end{equation}
We now contradict~\eqref{eq:fundBCH:Nzero} by evaluating $N$ at $T=1/w_1$, which is legitimate since $w_1\neq 0$. For any $j\neq 1$, the product $\prod_{i\neq j}(1-w_i{T})$ contains the factor $(1-w_1{T})$, which vanishes at $T=1/w_1$. Hence, only the term $j=1$ survives in the sum defining $N$, and
$$N(1/w_1) = w_1\cdot \frac{1}{w_1}\cdot \prod_{i\neq 1}\Big(1-\frac{w_i}{w_1}\Big) = \prod_{i=2}^{s}\Big(1-\frac{w_i}{w_1}\Big)\enspace .$$
Since $w_1,\ldots,w_s$ are pairwise distinct, each factor $1-w_i/w_1$ with $i\neq 1$ is a nonzero element of the field $\mathbb F_{2^u}$, so that the above product of nonzero elements is nonzero. This contradicts~\eqref{eq:fundBCH:Nzero}. Therefore $s=0$ and, by Step 2, $l$ is even and $(v_1,\ldots,v_l)$ is paired.
\end{proof}

\begin{proof}[Proof of Lemma~\ref{lem:BCHdesign}]
   First note that for any $j_1,\ldots, j_l \in \{1,\ldots,d\}$ 
   \begin{align*}
       \mathcal U^{(l)}_{j_1,\ldots, j_l} = \mathbb E_{U\sim \mathcal U\Big(\{-1,1\}^d\Big)} \Big[\prod_{i\leq l}U_{j_i}\Big] = \mathbf 1\Big\{\mathrm{there~exists~a~pairing~for~}(j_1,\ldots, j_l)\Big\}.
   \end{align*}

We now check the converse  of Lemma~\ref{lem:fundBCH}: if $(v_i)_{i\leq l}$ is paired, then $\sum_{i=1}^l\phi^{(2a-1)}(v_i)=0$ for every $a\leq m$. Indeed, pairing $v_1,\ldots,v_l$ into $l/2$ pairs of equal elements, each pair $(v_{J_i},v_{J_i'})$ with $v_{J_i}=v_{J_i'}$ contributes $\phi^{(2a-1)}(v_{J_i})+\phi^{(2a-1)}(v_{J_i'})=2\phi^{(2a-1)}(v_{J_i})=0$ to the sum, since the sum is taken in $\mathbb F_2^u$ (where $2x=0$ for every $x$); summing over all $l/2$ pairs gives the identity. Together with Lemma~\ref{lem:fundBCH}, this shows that, for $l\leq 2m$ and $v_1,\ldots,v_l\in\mathbb F_{2^u}\setminus\{0\}$, we have $\sum_{i=1}^l\phi^{(2a-1)}(v_i)=0$ for all $a\leq m$ if and only if $(v_i)_{i\leq l}$ is paired. In particular, if $l$ is odd, no pairing of $(v_1,\ldots,v_l)$ is possible, so by the contrapositive of Lemma~\ref{lem:fundBCH}, the system $\sum_{i=1}^l\phi^{(2a-1)}(v_i)=0,\ \forall a\leq m$, has no solution among $v_1,\ldots,v_l\in\mathbb F_{2^u}\setminus\{0\}$ -- so the equivalence above holds vacuously (both sides are always false) for odd $l$ as well.

Combining this equivalence, applied with $v_i=j_i$ for $i\leq l$ --~recall that $j_1,\ldots,j_l\in\{1,\ldots,d\}$ are identified with nonzero elements of $\mathbb F_{2^u}$~--, with Lemma~\ref{lem:BCHeq}, we obtain
$$\Big(\frac{1}{K}\sum_{k\leq K}\mu_k^{\otimes l}\Big)_{j_1,\ldots,j_l} = \mathbf 1\{\forall a\leq m: \textstyle\sum_{i=1}^l\phi^{(2a-1)}(j_i)=0\} = \mathbf 1\{(j_1,\ldots,j_l)\text{ is paired}\} = \mathcal U^{(l)}_{j_1,\ldots,j_l}\enspace ,$$
which concludes the proof.

\end{proof}

We now prove Lemma~\ref{lem:moments:XY}, stated in Section~\ref{sec:main_results}, which transfers the moment property of Lemma~\ref{lem:BCHdesign} to the centers $\pm X_k$ and to the observations $Y_i$.
\begin{proof}[Proof of Lemma~\ref{lem:moments:XY}]
Throughout the proof, we work conditionally on $\sigma$ and $(a_j)$, and we fix $i\in[n]$. Fix a positive integer $l\leq 2m+1$ and $j_1,\ldots,j_l\in[d]$. For $j\in[d]$, let $n_j:=|\{s\leq l:\ j_s=j\}|$ be the multiplicity of $j$ in $(j_1,\ldots,j_l)$, and let $S:=\{j\in[d]:\ n_j\text{ is odd}\}$. As shown in Step~2 of the proof of Lemma~\ref{lem:fundBCH}, $(j_1,\ldots,j_l)$ is paired if and only if $S=\emptyset$, so that $\bbE[U^{\otimes l}]_{j_1,\ldots,j_l}=\mathbf 1\{S=\emptyset\}$. Note also that $|S|\leq l\leq 2m+1$.

\emph{Proof of (i).} Since $\kg(i)$ and $z_i$ are independent, independent of $\sigma$ and $(a_j)$, and respectively uniform on $[K]$ and on $\{-1,1\}$, the first equality in~\eqref{eq:moments:X} is the definition of the conditional expectation. By~\eqref{eq:definition:Xlj}, the entry at $(j_1,\ldots,j_l)$ of the left-hand side of~\eqref{eq:moments:X} is
$$\bbE\big[z_i^l\big]\Big(\frac{\Delta}{\sqrt d}\Big)^l\prod_{s\leq l}a_{j_s}\ \frac{1}{K}\sum_{k\leq K}\prod_{s\leq l}\mu_{k,\sigma(j_s)}\enspace .$$
If $l$ is odd, then $\bbE[z_i^l]=0$, and $(j_1,\ldots,j_l)$ cannot be paired, so that both sides of~\eqref{eq:moments:X} vanish. If $l$ is even, then $l\leq 2m$ and $\bbE[z_i^l]=1$. Applying Lemma~\ref{lem:BCHdesign} at the entry $(\sigma(j_1),\ldots,\sigma(j_l))$, we get that $\frac1K\sum_{k\leq K}\prod_{s\leq l}\mu_{k,\sigma(j_s)}$ is equal to $1$ if $(\sigma(j_1),\ldots,\sigma(j_l))$ is paired and to $0$ otherwise. Since $\sigma$ is a bijection, $(\sigma(j_1),\ldots,\sigma(j_l))$ is paired if and only if $(j_1,\ldots,j_l)$ is paired, that is if and only if $S=\emptyset$. Finally, if $S=\emptyset$, then $\prod_{s\leq l}a_{j_s}=\prod_{j\in[d]}a_j^{n_j}=1$ since all the $n_j$ are even and $a_j^2=1$. Hence, the above entry is equal to $(\Delta/\sqrt d)^l\,\mathbf 1\{S=\emptyset\}=(\Delta/\sqrt d)^l\,\bbE[U^{\otimes l}]_{j_1,\ldots,j_l}$, which proves~\eqref{eq:moments:X}.

\emph{Proof of (ii).} Since $Y_{ij}\in\{-1,1\}$, we have $Y_{ij}^2=1$, so that $\prod_{s\leq l}Y_{ij_s}=\prod_{j\in S}Y_{ij}$. Conditionally on $\sigma$, $(a_j)$, $\kg(i)$ and $z_i$, the coordinates of $Y_i$ are independent and $\bbE[Y_{ij}\,|\,\sigma,(a_j),\kg(i),z_i]=2\cdot\frac{1+z_iX_{\kg(i),j}}{2}-1=z_iX_{\kg(i),j}$. As the elements of $S$ are pairwise distinct, we deduce that
$$\bbE\Big[\prod_{s\leq l}Y_{ij_s}\,\Big|\,\sigma,(a_j),\kg(i),z_i\Big]=\prod_{j\in S}z_iX_{\kg(i),j}\enspace .$$
If $S=\emptyset$, this product is empty and equal to $1$. If $S\neq\emptyset$, write $S=\{j'_1,\ldots,j'_{|S|}\}$. Integrating with respect to $\kg(i)$ and $z_i$, and applying~\eqref{eq:moments:X} at the order $|S|\leq 2m+1$ and at the entry $(j'_1,\ldots,j'_{|S|})$, we obtain $\bbE[\prod_{s\leq l}Y_{ij_s}\,|\,\sigma,(a_j)]=(\Delta/\sqrt d)^{|S|}\,\bbE[U^{\otimes |S|}]_{j'_1,\ldots,j'_{|S|}}=0$, because the $j'_r$ are pairwise distinct, so that $(j'_1,\ldots,j'_{|S|})$ is not paired. In both cases, $\bbE[\prod_{s\leq l}Y_{ij_s}\,|\,\sigma,(a_j)]=\mathbf 1\{S=\emptyset\}=\bbE[U^{\otimes l}]_{j_1,\ldots,j_l}$, which proves~\eqref{eq:moments:Y}.

Finally, the unconditional versions of (i) and (ii) follow by integrating with respect to $\sigma$ and $(a_j)$, since the conditional quantities do not depend on them.
\end{proof}

\begin{proof}[Proof of Lemma~\ref{lem:BCH:separation}]
For $x,y\in\mathbb F_2^u$, write $\langle x,y\rangle_{\mathbb F_2}:=\sum_{b\leq u}x_by_b\in\mathbb F_2$. As in Section~\ref{sec:bch:code}, we identify $[d]$ with $\mathbb F_{2^u}\setminus\{0\}$, and all the products of elements of $[d]$ below are taken in the field $\mathbb F_{2^u}$.

\emph{Step 1: reduction to the sums $S_k$.} For $k\in[K]$, let $\vartheta(k)\in\mathbb F_2^{m\times u}$ be defined by $\vartheta(k)_{a,b}:=1-(\bar\psi(k-1))_{a,b}$, so that $2(\bar\psi(k-1))_{a,b}-1=(-1)^{{\vartheta}(k)_{a,b}}$, and write ${\vartheta}^{(a)}(k)\in\mathbb F_2^u$ for its $a$-th row. Since $\bar\psi$ is a bijection, so is $\vartheta:[K]\to\mathbb F_2^{m\times u}$. By~\eqref{eq:BCH} and~\eqref{eq:definition_mu_k}, we have, for all $k\in[K]$ and $v\in[d]$,
\begin{equation}\label{eq:sep:param}
\mu_{k,v}=\prod_{a\leq m}\prod_{b\leq u}(-1)^{{\vartheta}(k)_{a,b}[\phi^{(2a-1)}(v)]_b}=(-1)^{\sum_{a\leq m}\langle{\vartheta}^{(a)}(k),\phi^{(2a-1)}(v)\rangle_{\mathbb F_2}}\enspace .
\end{equation}
In particular, for $k,k'\in[K]$, the entrywise product of $\mu_k$ and $\mu_{k'}$ is equal to $\mu_{k''}$, where $k''$ is defined by $\vartheta(k'')=\vartheta(k)+\vartheta(k')$. Let $k_0$ be the index such that $\vartheta(k_0)=0$, so that all the entries of $\mu_{k_0}$ are equal to $1$. If $k\neq k'$, then $\vartheta(k)\neq{\vartheta}(k')$, so that $\vartheta(k'')\neq 0$, that is $k''\neq k_0$. Writing $S_k:=\sum_{v\leq d}\mu_{k,v}$, we get $\langle\mu_k,\mu_{k'}\rangle=S_{k''}$. Hence, it suffices to prove that
\begin{equation}\label{eq:sep:goal}
s_*:=\max_{k\neq k_0}|S_k|\leq \eta_m d\enspace .
\end{equation}
The bounds on $\|\mu_k-\mu_{k'}\|_2^2$ and $\|\mu_k+\mu_{k'}\|_2^2$ then follow from $\|\mu_k\pm\mu_{k'}\|_2^2=2d\pm2\langle\mu_k,\mu_{k'}\rangle$, since $\|\mu_k\|_2^2=d$.

\emph{Step 2: a moment bound.} By Lemma~\ref{lem:BCHdesign} with $l=2m$,
$$\sum_{k\leq K}S_k^{2m}=\sum_{v_1,\ldots,v_{2m}\in[d]}\ \sum_{k\leq K}\ \prod_{s\leq 2m}\mu_{k,v_s}=K\,\Big|\big\{(v_1,\ldots,v_{2m})\in[d]^{2m}:\ (v_1,\ldots,v_{2m})\text{ is paired}\big\}\Big|\enspace .$$
Any paired $(v_1,\ldots,v_{2m})$ is obtained by choosing a partition of $\{1,\ldots,2m\}$ into $m$ pairs --~there are $(2m-1)!!=(2m-1)(2m-3)\cdots 1$ of them~-- and a common value in $[d]$ for each pair. Hence there are at most $(2m-1)!!\,d^m$ paired vectors and, since $K=(d+1)^m$,
\begin{equation}\label{eq:sep:moment}
\sum_{k\neq k_0}S_k^{2m}\leq \sum_{k\leq K}S_k^{2m}\leq (2m-1)!!\,d^m(d+1)^m\enspace .
\end{equation}

\emph{Step 3: invariance under dilations.} The sums $S_k$, $k\neq k_0$, are in general not all equal, but we show in this step that $S_k$ is invariant under the dilations $v\mapsto\lambda v$ of the coordinates. More precisely, we define an action $k\mapsto\lambda\cdot k$ of the nonzero elements $\lambda$ of $\mathbb F_{2^u}$ on $[K]$, and we prove that $S_{\lambda\cdot k}=S_k$ for all $k\in[K]$ and all $\lambda\neq 0$. Together with Step~4, which shows that the orbits $\{\lambda\cdot k:\ \lambda\neq 0\}$, $k\neq k_0$, contain at least $d/(2m-1)$ indices, this will allow us to turn the moment bound~\eqref{eq:sep:moment} into the uniform bound~\eqref{eq:sep:goal}. Fix $\lambda\in\mathbb F_{2^u}\setminus\{0\}$ and $\vartheta\in\mathbb F_2^{m\times u}$. For each $a\leq m$, the map $y\mapsto\langle{\vartheta}^{(a)},\psi(\lambda^{2a-1}y)\rangle_{\mathbb F_2}$ is an $\mathbb F_2$-linear form on $\mathbb F_{2^u}$, since $\psi$ is additive. As $\psi$ is moreover a bijection onto $\mathbb F_2^u$, any $\mathbb F_2$-linear form on $\mathbb F_{2^u}$ is of the form $y\mapsto\langle r,\psi(y)\rangle_{\mathbb F_2}$ for a unique $r\in\mathbb F_2^u$. Hence, there exists a unique $\lambda\cdot{\vartheta}\in\mathbb F_2^{m\times u}$ such that
\begin{equation}\label{eq:sep:action}
\langle(\lambda\cdot{\vartheta})^{(a)},\psi(y)\rangle_{\mathbb F_2}=\langle{\vartheta}^{(a)},\psi(\lambda^{2a-1}y)\rangle_{\mathbb F_2}\enspace ,\qquad\text{for all } a\leq m \text{ and } y\in\mathbb F_{2^u}\enspace .
\end{equation}
By uniqueness, $1\cdot{\vartheta}=\vartheta$, $\lambda\cdot 0=0$, and $\lambda\cdot(\lambda'\cdot{\vartheta})=(\lambda\lambda')\cdot{\vartheta}$ for all nonzero $\lambda,\lambda'$, since both sides of the latter satisfy~\eqref{eq:sep:action} with $\lambda\lambda'$ in place of $\lambda$. In particular, $\vartheta\mapsto\lambda\cdot{\vartheta}$ is a bijection of $\mathbb F_2^{m\times u}$, with inverse $\vartheta\mapsto\lambda^{-1}\cdot{\vartheta}$, and it maps nonzero elements to nonzero elements. For $k\in[K]$, define $\lambda\cdot k\in[K]$ by $\vartheta(\lambda\cdot k)=\lambda\cdot{\vartheta}(k)$. Since $\phi^{(2a-1)}(\lambda v)=\psi(\lambda^{2a-1}v^{2a-1})$, applying~\eqref{eq:sep:action} with $y=v^{2a-1}$ in~\eqref{eq:sep:param} yields $\mu_{k,\lambda v}=\mu_{\lambda\cdot k,v}$ for all $v\in[d]$. As $v\mapsto\lambda v$ is a bijection of $\mathbb F_{2^u}\setminus\{0\}=[d]$, we deduce that
$$S_{\lambda\cdot k}=\sum_{v\leq d}\mu_{k,\lambda v}=S_k\enspace .$$

\emph{Step 4: the orbits are large.} Let $\vartheta\neq0$ and let $a^\star\leq m$ be such that ${\vartheta}^{(a^\star)}\neq 0$. Assume that $\lambda\cdot{\vartheta}=\vartheta$. Then, by~\eqref{eq:sep:action} with $a=a^\star$ and by additivity of $\psi$, we have $\langle{\vartheta}^{(a^\star)},\psi((\lambda^{2{a^\star}-1}-1)y)\rangle_{\mathbb F_2}=0$ for all $y\in\mathbb F_{2^u}$. If $\lambda^{2{a^\star}-1}\neq1$, then $y\mapsto(\lambda^{2{a^\star}-1}-1)y$ is a bijection of $\mathbb F_{2^u}$ and, $\psi$ being onto $\mathbb F_2^u$, we get $\langle{\vartheta}^{(a^\star)},x\rangle_{\mathbb F_2}=0$ for all $x\in\mathbb F_2^u$, that is ${\vartheta}^{(a^\star)}=0$, a contradiction. Hence $\lambda$ is a root of the polynomial $x^{2{a^\star}-1}-1$, which has at most $2{a^\star}-1\leq 2m-1$ roots in the field $\mathbb F_{2^u}$. Now, if $\lambda\cdot{\vartheta}=\lambda'\cdot{\vartheta}$, then $(\lambda'^{-1}\lambda)\cdot{\vartheta}=\vartheta$. Therefore, the map $\lambda\mapsto\lambda\cdot{\vartheta}$, defined on the $d$ nonzero elements of $\mathbb F_{2^u}$, takes each of its values at most $2m-1$ times, so that the orbit $\{\lambda\cdot{\vartheta}:\ \lambda\neq0\}$ contains at least $d/(2m-1)$ elements, all of them nonzero.

\emph{Step 5: conclusion.} Let $k_1\neq k_0$ be such that $|S_{k_1}|=s_*$. By Steps~3 and~4 applied with $\vartheta=\vartheta(k_1)\neq 0$, the set $\{\lambda\cdot k_1:\ \lambda\neq 0\}$ contains at least $d/(2m-1)$ indices, all distinct from $k_0$, and $|S_{\lambda\cdot k_1}|=s_*$ for each of them. Together with~\eqref{eq:sep:moment}, this gives
$$\frac{d}{2m-1}\,s_*^{2m}\leq(2m-1)!!\,d^m(d+1)^m\enspace ,\qquad\text{that is}\qquad \Big(\frac{s_*}{d}\Big)^{2m}\leq\frac{(2m-1)\,(2m-1)!!\,(1+1/d)^m}{d}\enspace .$$
By the arithmetic-geometric mean inequality, $(2m-1)!!=\prod_{i=1}^m(2i-1)\leq\big(\frac1m\sum_{i=1}^m(2i-1)\big)^m=m^m$. Besides, $2m-1\leq 2m$ and, since $d\geq 2m$, $(1+1/d)^m\leq e^{m/d}\leq e^{1/2}\leq 2$. Hence $(s_*/d)^{2m}\leq 4m^{m+1}/d=\eta_m^{2m}$, which is~\eqref{eq:sep:goal}.

Finally, assume that $d\geq(4m)^{m+1}$. Then $d\geq 2m$ and $\eta_m^{2m}\leq 4m^{m+1}/(4m)^{m+1}=4^{-m}$, so that $\eta_m\leq1/2$ and $\|\mu_k-\mu_{k'}\|_2^2\geq 2(1-\eta_m)d\geq d$ for all $k\neq k'$. As for the centers, let $\varepsilon,\varepsilon'\in\{-1,1\}$ and $k,k'\in[K]$ with $(\varepsilon,k)\neq(\varepsilon',k')$. By~\eqref{eq:definition:Xlj}, since $|a_j|=1$ and $\sigma$ is a permutation, we have $\|\varepsilon X_k-\varepsilon'X_{k'}\|_2=\frac{\Delta}{\sqrt d}\|\varepsilon\mu_k-\varepsilon'\mu_{k'}\|_2$. If $k\neq k'$, then $\|\varepsilon\mu_k-\varepsilon'\mu_{k'}\|_2^2=2d-2\varepsilon\varepsilon'\langle\mu_k,\mu_{k'}\rangle\geq 2(1-\eta_m)d\geq d$. If $k=k'$, then $\varepsilon'=-\varepsilon$ and $\|\varepsilon\mu_k-\varepsilon'\mu_{k'}\|_2^2=\|2\mu_k\|_2^2=4d$. In both cases, $\|\varepsilon X_k-\varepsilon'X_{k'}\|_2\geq\Delta$.
\end{proof}

\bibliographystyle{plain}
\bibliography{biblio}

\begin{thebibliography}{10}

\bibitem{aloise2009np}
Daniel Aloise, Amit Deshpande, Pierre Hansen, and Preyas Popat.
\newblock {NP}-hardness of {E}uclidean sum-of-squares clustering.
\newblock {\em Machine Learning}, 75(2):245--248, 2009.

\bibitem{awasthi2015hardness}
Pranjal Awasthi, Moses Charikar, Ravishankar Krishnaswamy, and Ali~Kemal Sinop.
\newblock {The Hardness of Approximation of Euclidean k-Means}.
\newblock In {\em 31st International Symposium on Computational Geometry (SoCG
  2015)}, volume~34, pages 754--767. Schloss Dagstuhl -- Leibniz-Zentrum
  f{\"u}r Informatik, 2015.

\bibitem{BBP05}
Jinho Baik, G{\'e}rard~Ben Arous, and Sandrine P{\'e}ch{\'e}.
\newblock {Phase transition of the largest eigenvalue for nonnull complex
  sample covariance matrices}.
\newblock {\em The Annals of Probability}, 33(5):1643 -- 1697, 2005.

\bibitem{brennan2020reducibility}
Matthew Brennan and Guy Bresler.
\newblock Reducibility and statistical-computational gaps from secret leakage.
\newblock In {\em Proceedings of Thirty Third Conference on Learning Theory},
  volume 125 of {\em Proceedings of Machine Learning Research}, pages 648--847.
  PMLR, 2020.

\bibitem{brennan2018reducibility}
Matthew Brennan, Guy Bresler, and Wasim Huleihel.
\newblock Reducibility and computational lower bounds for problems with planted
  sparse structure.
\newblock In {\em Proceedings of the 31st Conference On Learning Theory},
  volume~75 of {\em Proceedings of Machine Learning Research}, pages 48--166.
  PMLR, 2018.

\bibitem{buhai2025quasi}
Rares-Darius Buhai, Jun-Ting Hsieh, Aayush Jain, and Pravesh~K. Kothari.
\newblock The quasi-polynomial low-degree conjecture is false.
\newblock In {\em 2025 IEEE 66th Annual Symposium on Foundations of Computer
  Science (FOCS)}, pages 2577--2590. IEEE, 2025.

\bibitem{CGGV25}
Alexandra Carpentier, Simone~Maria Giancola, Christophe Giraud, and Nicolas
  Verzelen.
\newblock Low-degree lower bounds via almost orthonormal bases.
\newblock {\em arXiv preprint arXiv:2509.09353}, 2025.

\bibitem{carpentier2025phase}
Alexandra Carpentier, Christophe Giraud, and Nicolas Verzelen.
\newblock Phase transition for stochastic block model with more than $\sqrt{n}$
  communities.
\newblock In {\em Proceedings of Thirty Ninth Conference on Learning Theory},
  volume 336 of {\em Proceedings of Machine Learning Research}, pages
  938--1000. PMLR, 2026.

\bibitem{carpentier2025rank}
Alexandra Carpentier and Nicolas Verzelen.
\newblock Statistical and computational challenges in ranking.
\newblock {\em arXiv preprint arXiv:2512.21111}, 2025.

\bibitem{carpentier2026low}
Alexandra Carpentier and Nicolas Verzelen.
\newblock Low-degree lower bounds for clustering in moderate dimension.
\newblock {\em arXiv preprint arXiv:2602.23023}, 2026.

\bibitem{chen2019beyond}
Sitan Chen and Ankur Moitra.
\newblock Beyond the low-degree algorithm: mixtures of subcubes and their
  applications.
\newblock In {\em Proceedings of the 51st Annual ACM SIGACT Symposium on Theory
  of Computing}, pages 869--880, 2019.

\bibitem{Dasgupta99}
Sanjoy Dasgupta.
\newblock Learning mixtures of gaussians.
\newblock In {\em 40th Annual Symposium on Foundations of Computer Science
  (Cat. No. 99CB37039)}, pages 634--644. IEEE, 1999.

\bibitem{pmlr-v195-diakonikolas23b}
Ilias Diakonikolas, Daniel~M. Kane, Thanasis Pittas, and Nikos Zarifis.
\newblock {SQ} lower bounds for learning mixtures of separated and bounded
  covariance {G}aussians.
\newblock In Gergely Neu and Lorenzo Rosasco, editors, {\em Proceedings of
  Thirty Sixth Conference on Learning Theory}, volume 195 of {\em Proceedings
  of Machine Learning Research}, pages 2319--2349. PMLR, 2023.

\bibitem{SQclustering}
Ilias Diakonikolas, Daniel~M. Kane, and Alistair Stewart.
\newblock Statistical query lower bounds for robust estimation of
  high-dimensional {G}aussians and {G}aussian mixtures.
\newblock In {\em 2017 IEEE 58th Annual Symposium on Foundations of Computer
  Science (FOCS)}, pages 73--84. IEEE, 2017.

\bibitem{diakonikolas2018list}
Ilias Diakonikolas, Daniel~M Kane, and Alistair Stewart.
\newblock List-decodable robust mean estimation and learning mixtures of
  spherical gaussians.
\newblock In {\em Proceedings of the 50th Annual ACM SIGACT Symposium on Theory
  of Computing}, pages 1047--1060, 2018.

\bibitem{dillon2025fixed}
Travis Dillon.
\newblock Fixed-strength spherical designs.
\newblock {\em SIAM Journal on Discrete Mathematics}, 40(2):789--815, 2026.

\bibitem{Even24}
Bertrand Even, Christophe Giraud, and Nicolas Verzelen.
\newblock Computation-information gap in high-dimensional clustering.
\newblock In Shipra Agrawal and Aaron Roth, editors, {\em Proceedings of Thirty
  Seventh Conference on Learning Theory}, volume 247 of {\em Proceedings of
  Machine Learning Research}, pages 1646--1712. PMLR, 2024.

\bibitem{Even25}
Bertrand Even, Christophe Giraud, and Nicolas Verzelen.
\newblock Computational lower bounds in latent models: clustering,
  sparse-clustering, biclustering.
\newblock {\em arXiv preprint arXiv:2506.13647}, 2025.

\bibitem{fei2018hidden}
Yingjie Fei and Yudong Chen.
\newblock Hidden integrality of sdp relaxations for sub-gaussian mixture
  models.
\newblock In {\em Proceedings of the 31st Conference On Learning Theory},
  volume~75 of {\em Proceedings of Machine Learning Research}, pages
  1931--1965. PMLR, 2018.

\bibitem{feldman2008learning}
Jon Feldman, Ryan O'Donnell, and Rocco~A. Servedio.
\newblock Learning mixtures of product distributions over discrete domains.
\newblock {\em SIAM Journal on Computing}, 37(5):1536--1564, 2008.

\bibitem{giraud2019partial}
Christophe Giraud and Nicolas Verzelen.
\newblock Partial recovery bounds for clustering with the relaxed $ k $-means.
\newblock {\em Mathematical Statistics and Learning}, 1(3/4):317--374, 2018.

\bibitem{holmgren2021counterexamples}
Justin Holmgren and Alexander~S. Wein.
\newblock Counterexamples to the low-degree conjecture.
\newblock In {\em 12th Innovations in Theoretical Computer Science Conference
  (ITCS 2021)}, volume 185 of {\em Leibniz International Proceedings in
  Informatics (LIPIcs)}, pages 75:1--75:9. Schloss Dagstuhl -- Leibniz-Zentrum
  f{\"u}r Informatik, 2021.

\bibitem{hopkins2018statistical}
Samuel Hopkins.
\newblock {\em Statistical inference and the sum of squares method}.
\newblock PhD thesis, Cornell University, 2018.

\bibitem{DBLP:journals/corr/abs-1711-07454}
Samuel~B. Hopkins and Jerry Li.
\newblock Mixture models, robustness, and sum of squares proofs.
\newblock In {\em Proceedings of the 50th Annual ACM SIGACT Symposium on Theory
  of Computing}, pages 1021--1034. ACM, 2018.

\bibitem{Hopkins17}
Samuel~B. Hopkins and David Steurer.
\newblock Efficient bayesian estimation from few samples: Community detection
  and related problems.
\newblock In {\em 2017 IEEE 58th Annual Symposium on Foundations of Computer
  Science (FOCS)}, pages 379--390, 2017.

\bibitem{hsu2013learning}
Daniel Hsu and Sham~M Kakade.
\newblock Learning mixtures of spherical {G}aussians: moment methods and
  spectral decompositions.
\newblock In {\em Proceedings of the 4th conference on Innovations in
  Theoretical Computer Science}, pages 11--20, 2013.

\bibitem{kothari2018robust}
Pravesh~K Kothari, Jacob Steinhardt, and David Steurer.
\newblock Robust moment estimation and improved clustering via sum of squares.
\newblock In {\em Proceedings of the 50th Annual ACM SIGACT Symposium on Theory
  of Computing}, pages 1035--1046, 2018.

\bibitem{KuniskyWeinBandeira}
Dmitriy Kunisky, Alexander~S. Wein, and Afonso~S. Bandeira.
\newblock Notes on computational hardness of hypothesis testing: Predictions
  using the low-degree likelihood ratio.
\newblock In Paula Cerejeiras and Michael Reissig, editors, {\em Mathematical
  Analysis, its Applications and Computation (ISAAC 2019)}, volume 385 of {\em
  Springer Proceedings in Mathematics \& Statistics}, pages 1--50. Springer,
  Cham, 2022.

\bibitem{kuperberg2006numerical}
Greg Kuperberg.
\newblock Numerical cubature using error-correcting codes.
\newblock {\em SIAM Journal on Numerical Analysis}, 44(3):897--907, 2006.

\bibitem{Kwon20}
Jeongyeol Kwon and Constantine Caramanis.
\newblock The em algorithm gives sample-optimality for learning mixtures of
  well-separated gaussians.
\newblock In Jacob Abernethy and Shivani Agarwal, editors, {\em Proceedings of
  Thirty Third Conference on Learning Theory}, volume 125 of {\em Proceedings
  of Machine Learning Research}, pages 2425--2487. PMLR, 2020.

\bibitem{lachaud2015number}
Gilles Lachaud and Robert Rolland.
\newblock On the number of points of algebraic sets over finite fields.
\newblock {\em Journal of Pure and Applied Algebra}, 219(11):5117--5136, 2015.

\bibitem{lesieur2016phase}
Thibault Lesieur, Caterina De~Bacco, Jess Banks, Florent Krzakala, Cris Moore,
  and Lenka Zdeborov{\'a}.
\newblock Phase transitions and optimal algorithms in high-dimensional
  {G}aussian mixture clustering.
\newblock In {\em 2016 54th Annual Allerton Conference on Communication,
  Control, and Computing (Allerton)}, pages 601--608. IEEE, 2016.

\bibitem{li2026algorithmic}
Zhangsong Li.
\newblock Algorithmic contiguity from low-degree heuristic {II}: Predicting
  detection-recovery gaps.
\newblock {\em arXiv preprint arXiv:2604.17410}, 2026.

\bibitem{LiuLi2022}
Allen Liu and Jerry Li.
\newblock Clustering mixtures with almost optimal separation in polynomial
  time.
\newblock In {\em Proceedings of the 54th Annual ACM SIGACT Symposium on Theory
  of Computing}, STOC 2022, pages 1248--1261, New York, NY, USA, 2022.
  Association for Computing Machinery.

\bibitem{loffler2021optimality}
Matthias L{\"o}ffler, Anderson~Y Zhang, and Harrison~H Zhou.
\newblock Optimality of spectral clustering in the gaussian mixture model.
\newblock {\em The Annals of Statistics}, 49(5):2506--2530, 2021.

\bibitem{LuZhou2016}
Yu~{Lu} and Harrison~H. {Zhou}.
\newblock {Statistical and Computational Guarantees of Lloyd's Algorithm and
  its Variants}.
\newblock {\em ArXiv e-prints}, December 2016.

\bibitem{luo2023computational}
Yuetian Luo and Chao Gao.
\newblock Computational lower bounds for graphon estimation via low-degree
  polynomials.
\newblock {\em The Annals of Statistics}, 52(5):2318--2348, 2024.

\bibitem{ma2015computational}
Zongming Ma and Yihong Wu.
\newblock Computational barriers in minimax submatrix detection.
\newblock {\em The Annals of Statistics}, 43(3):1089--1116, 2015.

\bibitem{macwilliams1977theory}
F.~Jessie MacWilliams and Neil J.~A. Sloane.
\newblock {\em The Theory of Error-Correcting Codes}, volume~16 of {\em
  North-Holland Mathematical Library}.
\newblock North-Holland, Amsterdam, 1977.

\bibitem{mao2026polynomial}
Songtao Mao.
\newblock The polynomial-time low-degree conjecture is false.
\newblock {\em arXiv preprint arXiv:2607.20318}, 2026.

\bibitem{montanari2024equivalence}
Andrea Montanari and Alexander~S. Wein.
\newblock Equivalence of approximate message passing and low-degree polynomials
  in rank-one matrix estimation.
\newblock {\em Probability Theory and Related Fields}, 191(1--2):181--233,
  2025.

\bibitem{ndaoud2022sharp}
Mohamed Ndaoud.
\newblock {Sharp optimal recovery in the two component Gaussian mixture model}.
\newblock {\em The Annals of Statistics}, 50(4):2096 -- 2126, 2022.

\bibitem{Regev2017}
Oded {Regev} and Aravindan {Vijayaraghavan}.
\newblock {On Learning Mixtures of Well-Separated Gaussians}.
\newblock In {\em {2017 IEEE 58th Annual Symposium on Foundations of Computer
  Science (FOCS)}}, pages 85--96, 2017.

\bibitem{SchrammWein22}
Tselil Schramm and Alexander~S Wein.
\newblock Computational barriers to estimation from low-degree polynomials.
\newblock {\em The Annals of Statistics}, 50(3):1833--1858, 2022.

\bibitem{shafarevich2013basic}
Igor~R. Shafarevich.
\newblock {\em Basic Algebraic Geometry 1: Varieties in Projective Space}.
\newblock Springer, Berlin, Heidelberg, 3 edition, 2013.

\bibitem{SohnWein25}
Youngtak Sohn and Alexander~S. Wein.
\newblock Sharp phase transitions in estimation with low-degree polynomials.
\newblock In {\em Proceedings of the 57th Annual ACM Symposium on Theory of
  Computing}, pages 891--902. ACM, 2025.

\bibitem{stroud1971approximate}
A.~H. Stroud.
\newblock {\em Approximate Calculation of Multiple Integrals}.
\newblock Prentice-Hall Series in Automatic Computation. Prentice-Hall,
  Englewood Cliffs, NJ, 1971.

\bibitem{VEMPALA2004}
Santosh Vempala and Grant Wang.
\newblock A spectral algorithm for learning mixture models.
\newblock {\em Journal of Computer and System Sciences}, 68(4):841--860, 2004.

\bibitem{SurveyWein2025}
Alexander~S. Wein.
\newblock Computational complexity of statistics: New insights from low-degree
  polynomials, 2025.

\end{thebibliography}

\end{document}